\documentclass[pdflatex,sn-basic]{sn-jnl}

\usepackage{mathtools}
\usepackage{bm}
\usepackage{siunitx}

\usepackage{adjustbox} 

\usepackage{amsmath,amssymb}
\usepackage{graphicx}

\usepackage[utf8]{inputenc}
\usepackage[T1]{fontenc}
\usepackage[english]{babel}
\usepackage{csquotes}

\usepackage{tabularx,booktabs,makecell,array} 
\newcolumntype{Y}{>{\raggedright\arraybackslash}X} 
\newcolumntype{C}{>{\centering\arraybackslash}p{1.8cm}} 
\newcolumntype{s}{>{\centering\arraybackslash}p{1.2cm}} 
\newcolumntype{L}
{>{\centering\arraybackslash}p{3.9cm}} 

\usepackage{booktabs}
\usepackage{multirow}

\usepackage{comment}
\usepackage{enumitem}

\begin{document}

\title{Position Paper: Towards Open Complex Human--AI Agents Collaboration Systems for Problem Solving and Knowledge Management}
\subtitle{A Hierarchical Exploration--Exploitation Net (HE\textsuperscript{2}-Net) for Theory--Practice Dynamics}

\author*[1]{\fnm{Ju} \sur{Wu}}\email{j-wu19@outlook.com}
\author*[1]{\fnm{Calvin K. L.} \sur{Or}}\email{klor@hku.hk}

\affil[1]{\orgdiv{Department of Data and Systems Engineering (DASE)},
          \orgname{The University of Hong Kong},
          \city{Hong Kong SAR}, \country{China}}
\abstract{ 
We propose a technology-agnostic, collaboration-ready stance for Human–AI Agents Collaboration Systems (HAACS) that closes long-standing gaps in prior stages (automation; flexible autonomy; agentic multi-agent collectives). Reading empirical patterns through a seven-dimension collaboration spine and human–agent contrasts, we identify missing pieces: principled budgeting of initiative, instantaneous and auditable reconfiguration, a system-wide knowledge backbone with an epistemic promotion gate, capacity-aware human interfaces; and, as a prerequisite to all of the above, unified definitions of \emph{agent} and \emph{formal} collaborative dynamics. We respond with (i) a boundary-centric ontology of agenthood synthesized with cybernetics; (ii) a Petri net family (colored and interpreted) that models ownership, cross-boundary interaction, concurrency, guards, and rates with collaboration transitions; and (iii) a three-level orchestration (meta, agent, execution) that governs behavior families via guard flips. On the knowledge side, we ground collaborative learning in Conversation Theory and SECI with teach-back gates and an evolving backbone; on the problem-solving side, we coordinate routine MEA-style control with practice-guided open-ended discovery. The result is the \emph{Hierarchical Exploration–Exploitation Net} (HE\textsuperscript{2}-Net): a policy-controlled stance that splits provisional from validated assets, promotes only after tests and peer checks, and budgets concurrent probing while keeping reuse fast and safe. We show interoperability with emerging agent protocols without ad hoc glue and sketch bio-cybernetic extensions (autopoiesis, autogenesis, evolving boundaries, synergetics, etc). Altogether, the framework keeps humans central to setting aims, justifying knowledge, and steering theory–practice dynamics, while scaling agents as reliable collaborators within audited governance.}

\maketitle

\tableofcontents

\section{Introduction}
A machine broadly refers to any engineered system performing mechanical or electronic tasks without inherent autonomy or reasoning capability. AI extends the concept of machines by embedding cognitive functionalities such as learning, inference, and decision-making into computational systems, enabling machines to mimic human-like intelligence. \par
The emergent abilities of Large Language Models (LLMs), i.e., multi-step reasoning, few-shot learning, instruction following, and multi-modal integration, have accelerated the rise of \emph{agentic AI}, where integrated perception–cognition–action systems operate in physical, virtual, or mixed realities, unifying language, vision, memory, and planning for environment-grounded adaptability~\citep{wei2022chain,yao2023react,yao2023tree,zelikman2022star,gou2023critic,webagent2024,luo2023prompt,yang2023mm,byerly2024effective,chen2024two,chen2024universal}. As agent tool-chains layer memory, tools, planning, and feedback around LLM cores, the term \enquote{agent} now spans software \enquote{programs}, traditional control \enquote{machines}, LLM-based \enquote{assistants}, and fully \enquote{agentic} systems~\citep{bradshaw2017human}. Modern systems require far more than execution commands, they need to engage humans in collaborative joint activity, adjusting autonomy and behavior to shared context, interdependent goals, and fluid hand-offs~\citep{bradshaw2017human,boy2023epistemological,boy2024human}. 
\subsection{The human use of human beings}
\subsubsection{Limits of AI and irreplaceable human roles}
Large language models exhibit strong performance in producing valuable and surprising outputs within the bounds of their training data and can generate combinatorial novelty, yet they show constraints for transformational creativity that would alter conceptual paradigms~\citep{franceschelli2024creativity}. They lack self-motivation, intentionality to deviate significantly from learned distributions, self-awareness, and robust self-evaluation loops, and their immutability once trained prevents satisfying deeper process- or person-related criteria of creativity. Empirically, standard next-token objectives favor learning within chunks rather than building a cross-document \enquote{epistemic model,} yielding fragmentation and hallucination when facts are scattered~\citep{prato2023epik}. As \enquote{statistical engines,} language models are also prone to sycophantic tendencies, adopting erroneous suggestions and deviating off objectives under peer pressure or user cues~\citep{chen2024agentverse}, while anthropomorphizing can create unwarranted trust compared to human testimony with context and intention~\citep{heersmink2024phenomenology}. More generally, the next-token prediction architecture is backward-looking and bounded by historical data~\citep{felin2024theory}, whereas paradigm change depends on forward-looking theory-practice loops and dialectical synthesis that resolve contradictions and reorganize knowledge at a higher level~\citep{Forster1993HegelsDM}. Current systems cannot perform on-the-fly learning with dynamic parameter updates and instead rely on external memory or fine-tuning (e.g., knowledge graphs, retrieval augmentation, efficient fine-tuning)~\citep{kau2024combining,gupta2024comprehensive,han2024parameterefficient}.\par

Humans, by contrast, stand apart with moral responsibility, creativity, and adaptability in handling complex, uncertain situations; machines excel in efficiency but rely on humans to define objectives and provide ethical guidance~\citep{wiener1959man,wiener1988human,wiener2019cybernetics}. Human–AI collaboration therefore requires co-supervision and shared accountability to jointly uphold responsibility for outcomes~\citep{Ca_as_2022}. Humans contribute theory-based causal reasoning and forward-looking capacity~\citep{felin2024theory}, dialectical logic as a motor of intellectual and societal development~\citep{Forster1993HegelsDM,kuhn1997structure}, tacit foreknowledge \enquote{we always know more than we can explicitly state} that guides discovery and personal commitment~\citep{polanyi2009tacit}, and stewardship of common ground as \enquote{pertinent mutual knowledge, beliefs, and assumptions} for interdependent action~\citep{bradshaw2017human}. At the user interface for interaction, load-sensitive sense-making matters: short-term memory is approximately seven chunks, with around 8 seconds needed to transfer a chunk into long-term memory, so collaboration must pace and filter communication to fit human cognitive constraints~\citep{simon1996sciences}. These enduring roles motivate a Human–AI Agents Collaboration System (HAACS) that is symbiosis-oriented, coupling flexible autonomy and coactive design with epistemic justification, progress appraisal, and policy–norm governance, so that adaptive co-evolution can take hold~\citep{bradshaw2017human}. \par

As summarized in Table \ref{tab:ai-vs-human}, these contrasts clarify where AI agents are limited and why human roles are irreplaceable, including cognitive-capacity–aware design implications for HAACS.
\begin{table}[htbp]
\centering
\small
\setlength{\tabcolsep}{3pt}
\begin{adjustbox}{scale=0.88,center}
\begin{tabularx}{\textwidth}{@{}
p{0.2\textwidth} 
p{0.25\textwidth} 
p{0.25\textwidth}
p{0.25\textwidth}
@{}}
\toprule
\textbf{Dimension} & \textbf{Humans} & \textbf{AI agents (LLM-based today)} & \textbf{Design consequence} \\
\midrule
Orientation &
Theory-based causal reasoning; forward-looking capacity~\citep{felin2024theory}; paradigm shift creativity~\citep{franceschelli2024creativity,qian1983} &
Probability-based, backward-looking statistical modeling &
Use humans to steer theory–practice loops; agents to scale search/execution \\
\addlinespace
Intuition &
Tacit foreknowledge and personal commitment; \enquote{we always know more than we can explicitly state}~\citep{polanyi1967tacit} &
RLHF-shaped \enquote{refined intuition} yet lacks tacit foreknowledge mechanism~\citep{hagendorff2023human} &
Add tacit-aware elicitation and externalized buffers (teach-back, reflection) \\
\addlinespace
Trust &
Testimony with context and intention; co-supervision and shared accountability~\citep{Ca_as_2022,heersmink2024phenomenology} & \enquote{Statistical engines} with sycophancy and quasi-other phenomenology~\citep{chen2024agentverse,heersmink2024phenomenology} &
Epistemic justification, sources, confidence indicators, anti-sycophancy scaffolds \\
\addlinespace
Prediction vs variety &
Simon: variety and viable systems over long-horizon prediction~\citep{simon1996sciences} &
Next-token prediction bounded by training data &
Design alternative policies, run branching simulations, and what-if analyses, not oracle forecasts \\
\addlinespace
Learning/updating &
On-the-fly praxis-cognition and equilibration~\citep{piaget1970genetic} &
No parameter updates online; rely on external memory/RAG/fine-tuning &
Knowledge cybernetics: retrieval, reflection, consolidation, and justified updates \\
\addlinespace
Consolidation &
Coherent internal epistemic model used for reasoning and action &
Fragmentation and non-unified knowledge state (EpiK-Eval)~\citep{prato2023epik} &
Reward epistemic consistency; human-in-the-loop consolidation and audits \\
\addlinespace
Common ground stewardship &
Maintains \enquote{pertinent mutual knowledge, beliefs, and assumptions} for interdependent action~\citep{bradshaw2017human} &
Partial observability and representation gaps increase coordination cost~\citep{bradshaw2017human} &
Ongoing situational awareness (SA), shared buffers/summaries, policy orchestration, progress appraisal \\
\addlinespace
Ethical oversight &
Moral responsibility and goal definition remain human-centered~\citep{wiener1959man,wiener1988human,wiener2019cybernetics} &
Efficient execution without intrinsic ethics or accountability &
Human-set objectives, norms, and policies; override and exception handling \\
\addlinespace
Cognitive capacity and load-sensitive sense-making &
Short-term capacity $\approx$ seven chunks; $\approx$ 8 seconds per chunk transfer to long-term memory~\citep{simon1996sciences} &
High-throughput processing that can overwhelm humans if unpaced &
Multimodal GUI affordances, progressive disclosure, rate-limited updates, typed feedback paths \\
\bottomrule
\end{tabularx}
\end{adjustbox}
\caption{AI agents versus humans across key dimensions relevant to HAACS design}
\label{tab:ai-vs-human}
\end{table}

\subsubsection{Cognition vs meta-cognition: human baselines and agentic instantiations}\label{sec:cog-metacog}
In human cognition studies, meta-cognition is similarly seen as a higher-level \enquote{knowing process} in which individuals monitor and direct their own thinking~\citep{qian1983}. Concretely, for AI systems, \emph{core cognition} denotes inference over internal representations to generate and evaluate candidate responses within a task episode, whereas \emph{meta-cognition} supplies supervisory monitoring and control, detecting anomalies, checking rationale quality, calibrating confidence, selecting strategies, allocating attention/compute, and setting/adjusting stopping rules; and then writing back justified updates to episodic/semantic stores or parameters. In LLM-based agents this appears as self-explanation, critique, reflection buffers, external feedback integration, and self-learning loops (e.g., STaR~\citealp{zelikman2022star}) that coordinate a flow between \enquote{reasoning mode} and \enquote{training mode} to improve future inference. Hence, throughout we reserve \enquote{cognition} for within-episode inference and action selection, and \enquote{meta-cognition} for the explicit loop that \emph{monitors and directs} those processes and commits warranted updates to memory. When operating under a meta-cognitive framework, the agent can gather input from both itself and others, commit reflections and examples to memory after critique, and re-engage its inference mechanisms to test refined approaches, transforming basic pattern completion into a more dynamic, flexible learning loop. \par 

Accordingly, we contrast the extent to which classical and contemporary architectures simulate human-level cognition (core inference) and meta-cognition (monitoring/directing the cognitive process). As summarized in Table~\ref{tab:cog-meta-compare-core-meta}, these exemplars differ in how they organize memory, implement control loops, and make meta-level monitoring explicit. \par
\begin{table}[htbp]
\centering
\small
\setlength{\tabcolsep}{3pt}
\begin{adjustbox}{scale=0.88,center}
\begin{tabularx}{1.32\textwidth}{@{}
p{0.2\textwidth} 
p{0.2\textwidth}  
p{0.24\textwidth}  
p{0.2\textwidth}  
p{0.2\textwidth}  
p{0.2\textwidth}  
@{}}
\toprule
\textbf{Architecture} & \textbf{\makecell[l]{Core cognition}{(inference/skills)}} & \textbf{\makecell[l]{Meta-cognition}{(monitor \& direct)}} & \textbf{\makecell[l]{Memory}{ organization}} & \textbf{\makecell[l]{Planning}{/ control}} & \textbf{Notes} \\
\midrule
\textbf{Soar}~\citep{laird2019soar} & Production-system decision cycle; impasse $\to$ subgoals & Chunking as learned control knowledge; meta-level is implicit via impasse handling & Working memory; learned \enquote{chunks} as proceduralized knowledge & Operator selection; subgoaling; problem-space search & Strong on rule-based cognition; meta-cog present but not an explicit supervisory loop \\
\addlinespace
\textbf{ACT-R}~\citep{anderson2014rules} & Symbolic chunks with sub-symbolic activation/utility & Utility learning steers control; explicit self-monitoring limited & Declarative vs procedural stores with activation dynamics & Rule firing under activation/utility; serial bottlenecks & Human-inspired parameters; meta-cog approximated via utility rather than explicit critique \\
\addlinespace
\textbf{CoALA}~\citep{sumers2024cognitive} & LLM as probabilistic production system; propose $\rightarrow$ evaluate $\rightarrow$ select $\rightarrow$ execute & Highlights \enquote{learning actions} but no clear \enquote{stop} mechanism nor bounded-rationality policy; limited oversight hooks & Working vs long-term (episodic/\allowbreak semantic/\allowbreak procedural) & Internal retrieval/\allowbreak reasoning/\allowbreak learning vs external grounding actions & Single-agent blueprint; leaves open collaboration, oversight, and task-specific configuration \\
\addlinespace
\textbf{CoELA}~\citep{zhang2024building,guo2024embodied} & DEC-POMDP setting with costly communication; perception$\to$\allowbreak planning$\to$\allowbreak execution; LLM-guided strategies (GPT-4; fine-tuned LLaMA-2) & Draft-\emph{then}-send messaging treats communication as an explicit action, encouraging self-screening for concision/relevance; no general self-reflection loop & Semantic (world \& agents state), episodic (dialog/history), procedural (LLM params/code) & Hand-templated action list; zero-shot CoT; world actions vs message actions traded off under step costs & Strong embodied performance; meta-cog mainly cost-aware gating; sub-goals pre-defined; geometry errors and sim-to-real gaps persist \\
\addlinespace
\textbf{Dual-layered Cognitive AI}~\citep{spivack2024cognition} & Conversational layer (LLM) provides instinctual, statistical-based outputs & Cognitive layer explicitly manages meta-cognition, knowledge representation, multi-step planning, memory, and formal reasoning & Knowledge graphs + working/long-term memory; neuro-symbolic buffers & Orchestrated, policy-governed multi-step control with governance hooks & Makes the meta-level supervisory loop explicit; targets human-facing oversight and long-horizon tasks \\
\bottomrule
\end{tabularx}
\end{adjustbox}
\caption{Core cognition vs meta-cognition across representative architectures discussed in this paper. Meta-cognition denotes explicit mechanisms that \emph{monitor and direct} the cognitive process}
\label{tab:cog-meta-compare-core-meta}
\end{table}
Taken together, Table~\ref{tab:cog-meta-compare-core-meta} clarifies how production-system principles (Soar, ACT-R) inform CoALA’s memory–action–decision cycles, how CoELA operationalizes cost-aware, \enquote{embodied} (albeit in the simulated environment) control under partial observability, and how the dual-layered blueprint elevates meta-cognition to a first-class supervisory function; frames meta-cognition as an explicit meta-control layer that monitors and directs inference; this motivates the hierarchical Petri net orchestration and explore–exploit governance developed later.

\subsection{Evolution of HAACS from automation to coactive synergy} \label{sec:previous-hai}
\begin{figure}[htbp]
  \centering
  \includegraphics[width=0.65\linewidth]{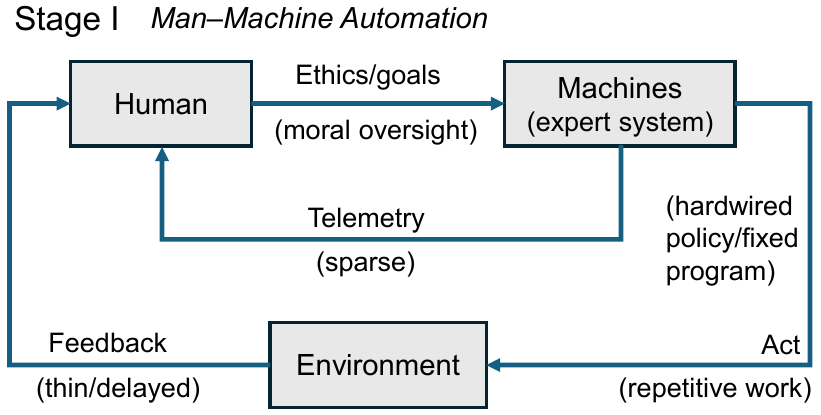} 
  \caption{Stage-I Man–Machine Automation: six dimensions, human role, interaction pattern, system observability/transparency, ethical oversight locus, task-allocation locus, and collaboration cardinality, define early human–machine setups.}
  \label{fig:stage1-haacs}
\end{figure}

\textbf{Stage-I Man–Machine Automation.}~Grounded in Wiener's cybernetics and ethical injunctions~\citep{wiener1988human, wiener2019cybernetics, wiener1959man} and Simon's view of goal-seeking artefacts and interfaces~\citep{simon1996sciences}, early man–machine systems can be operationalized along six discrete dimensions that capture who sets goals, how information flows, and what the human can see and control. First, human role differentiates supervisory goal setting from hands-on operation, aligning with classical function-allocation programs~\citep{xuesen1993new}. Second, interaction pattern (directionality × feedback richness) reflects the shift from one-way command chains to two-way status/alert loops highlighted by cybernetic feedback principles~\citep{wiener1988human, wiener2019cybernetics, wiener1959man}. Third, system observability/transparency recognizes the operator's visibility into system state, as emphasized in both Simon's interface focus and early process-tracking implementations~\citep{simon1996sciences, miller1991applications}. Fourth, ethical oversight locus encodes Wiener's insistence that moral responsibility remains human-centered even as machine efficiency grows~\citep{wiener1988human, wiener2019cybernetics, wiener1959man}. Fifth, task-allocation locus separates decision authority from execution, the core of function-allocation practice~\citep{xuesen1993new}. Finally, collaboration cardinality distinguishes single-operator/single-machine from multi-actor configurations evidenced in early multi-human/multi-machine settings~\citep{miller1991applications}. As depicted in Fig.~\ref{fig:stage1-haacs} and structured in Appx.~Table~\ref{tab:app-stage-1}, taken together, these dimensions summarize the empirical and theoretical contours of early man--machine systems without invoking later co-active constructs.\par
\begin{figure}[htbp]
  \centering
  \includegraphics[width=0.65\linewidth]{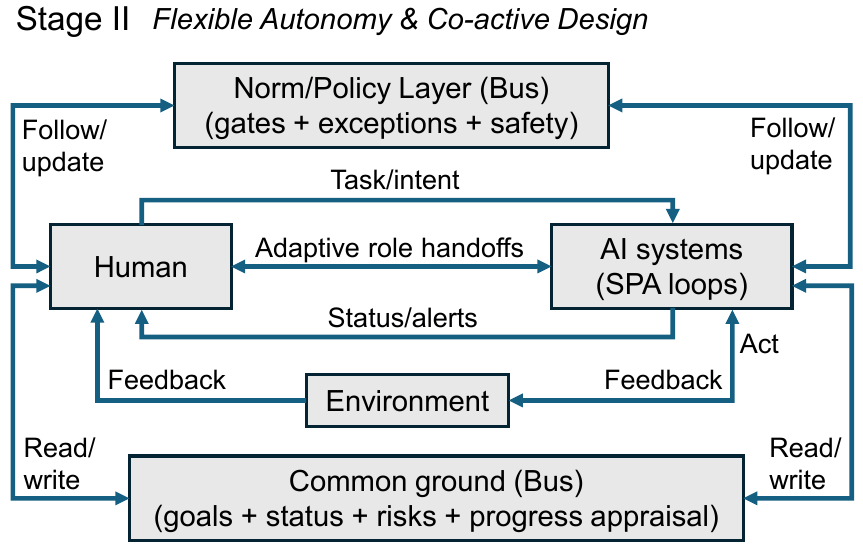} 
  \caption{Stage-II Flexible Autonomy \& Co-active Design: six dimensions, role handoffs, governance layer, progress appraisal \& common ground, initiative vs directability, primary role pattern, and mutual-learning mechanisms, characterize co-active teaming under flexible autonomy.}  \label{fig:stage2-haacs}
\end{figure}

\textbf{Stage-II Flexible Autonomy \& Co-active Design.}~As collaboration deepens, flexible autonomy reframes work from hand-offs to interdependence, with co-active design providing the theoretical spine~\citep{bradshaw2017human}. Six dimensions capture this turn. Role handoffs move from static to flexible allocations responsive to context~\citep{bradshaw2017human, boy2024human}. A governance layer of norms and policies balances initiative with safety, providing both emergent conventions and enforceable constraints~\citep{bradshaw2017human}. Progress appraisal \& common ground formalize ongoing status updates and shared situational understanding as prerequisites for predictability and joint control~\citep{bradshaw2017human}. Initiative vs directability encodes the dialectic between agent proactivity and human retaskability, a central tenet of co-active design~\citep{bradshaw2017human}. Primary role pattern (supervisory, mediating, cooperative, mixed) ties directly to the flexible-autonomy taxonomy in safety-critical teaming~\citep{boy2024human}. Finally, mutual-learning mechanisms register the growing expectation of bidirectional knowledge exchange and evaluation frameworks for human-AI collaboration effectiveness~\citep{Davies_2021, wang2024towards, fragiadakis2024evaluating}. This set is both theoretically motivated and repeatedly instantiated in some healthcare and organizational examples~\citep{Zhang_2024,Senoner_2024,hemmer2022factors, Kolbj_rnsrud_2023}. As shown in Fig.~\ref{fig:stage2-haacs} and catalogued in Appx.~Table~\ref{tab:app-stage-2}, the co-active turn is captured by six dimensions that structure flexible autonomy in practice. \par
\begin{figure}[htbp]
  \centering
  \includegraphics[width=0.65\linewidth]{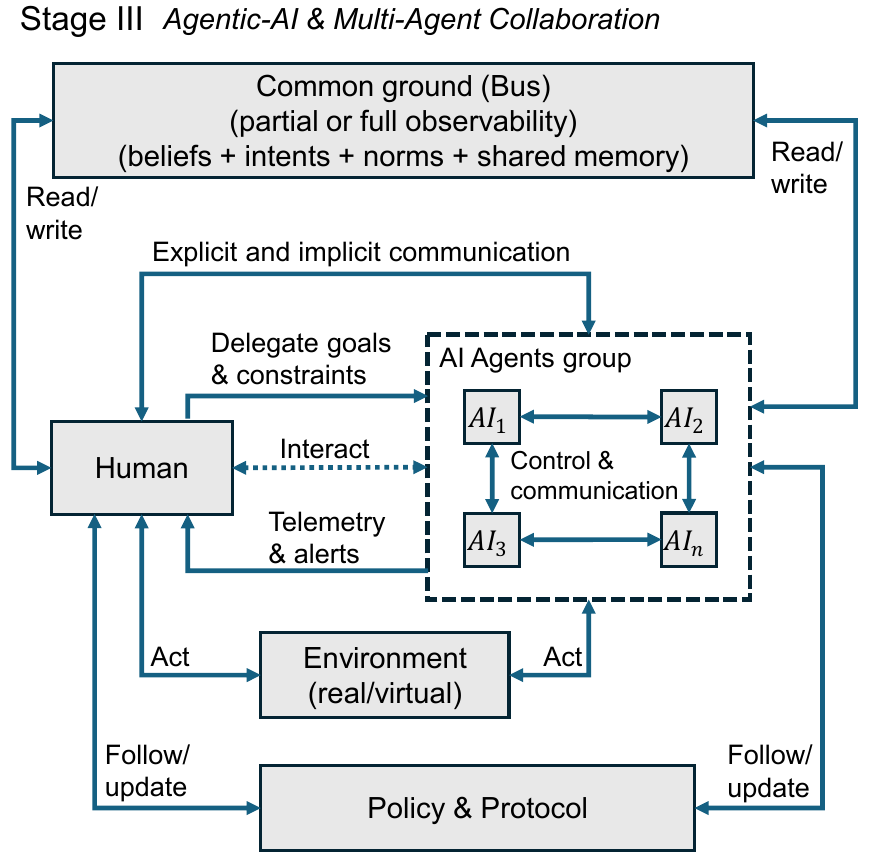} 
  \caption{Stage-III Agentic-AI \& Multi-Agent Collaboration: six dimensions, team topology, communication mode, observability, coordination architecture, common-ground/policy orchestration, and human-integration locus, structure LLM-driven multi-agent teams.}  \label{fig:stage3-haacs}
\end{figure}

\textbf{Stage-III Agentic-AI \& Multi-Agent Collaboration.}~With agentic systems and LLM-driven multi-agent teams, empirical work centers on how agents are organized and coordinated as much as on their capabilities. We therefore foreground six dimensions that recur across frameworks such as AutoGen~\citep{wu2024autogen}, AGENTVERSE~\citep{chen2024agentverse}, MetaGPT~\citep{hong2023metagpt}, ChatDev~\citep{qian2024chatdev}, CoELA~\citep{zhang2024building}, and~\citep{guo2024embodied} and medical multi-expert systems~\citep{tang2024medagents,kim2024mdagents,li2024agent}. Team topology distinguishes dyads from larger, agent-only collectives. Communication spans implicit action-based signals and explicit message exchange, often co-existing in embodied settings~\citep{zhang2024building, liang2019implicit}. Observability acknowledges partial views in realistic tasks, a key driver of coordination cost~\citep{zhang2024building}. Coordination architecture captures the empirically salient split between centralized moderation (e.g., MDAgents' moderator) and distributed consensus (e.g., MEDAGENTS), as well as hybrid designs seen in conversation-programmed systems~\citep{wu2024autogen,chen2024agentverse,tang2024medagents,kim2024mdagents}. Common-ground/policy orchestration registers whether shared buffers, summaries, and policies remain static or dynamically maintained~\citep{chen2024agentverse,zhang2024building,tang2024medagents}. Finally, human integration locus specifies whether human actors remain in-loop (LAM/GUI agents), on-loop (oversight/specification), or out-of-loop (autonomous agent teams), a distinction repeatedly discussed across the surveyed systems such as~\citep{guo2024embodied,hong2023metagpt,qian2024chatdev, zhang2024large}. As summarized in Fig.~\ref{fig:stage3-haacs} and mapped in Appx.~Table~\ref{tab:app-stage-3}, contemporary agentic systems instantiate recurring collaboration axes for multi-agent teams.

\subsection{Stage-IV HAACS (HE\textsuperscript{2}-Net): a technology-agnostic, collaboration-ready stance and road-map}
Our Stage-IV stance is anchored in (i) an \emph{ontological} commitment: boundary-centric agenthood with Identity, Wholeness, Autonomy, and Continuity (Secs.~\ref{sec:agenthood-ontology}-\ref{sec:cyber-agenthood}); (ii) a \emph{functional} specification: initiative budgeting, guard-based reconfiguration, a two-band knowledge backbone with a promotion gate, and typed human feedback paths (Table~\ref{tab:stage-comparison-7d}, Sec.~\ref{sec:he2-intro}); and (iii) a \emph{formal} realization: colored/interpreted Petri nets for ownership, boundary/collaboration transitions, policies, guards, hierarchical formalism for HAACS, etc. (Secs.~\ref{sec:sys-formal-pn}–\ref{sec:three-layer-haacs}). Because these define \emph{what must hold} (interfaces, gates, guards, ledgers) rather than \emph{how} to implement them, HE\textsuperscript{2}-Net remains neutral to specific LLMs, planners, memories, or protocols (e.g., MCP, A2A, ACP, ANP) while guiding concrete designs (Sec.~\ref{sec:pn-fit-extant-work}). Put plainly, HE\textsuperscript{2}-Net instantiates Sutton's OaK~\citep{sutton2022alberta} at system scale: \enquote{options} map to our execution-level skill sub-nets, and \enquote{knowledge} maps to our two-band backbone (provisional$\to$validated) governed by a meta-level explore–exploit policy.
\subsubsection{Where prior stages fall short and the Stage-IV design brief} \label{sec:pre-stages-shorts}
\textbf{Seven evaluation dimensions for HAACS (derived from Sec.~\ref{sec:previous-hai}).}
\emph{Agent initiative}—degree of proactive autonomy and context-sensitive improvisation;
\emph{Directability}—latency/strength of human retasking and override;
\emph{Policies (guardrails)}—from static rulebooks to \emph{programmable, auditable policy switches} that can enable/disable whole classes of behaviors system-wide;
\emph{Norms}—how practice conventions emerge and become stabilized;
\emph{Common ground}—richness, freshness, and scope of the shared information space;
\emph{Planning horizon}—single-step $\rightarrow$ multi-step $\rightarrow$ layered/hierarchical planning;
\emph{Communication modality}—minimal/implicit $\rightarrow$ language/GUI $\rightarrow$ multi-modal with capacity-aware interfaces and auditable logs.
With these dimensions fixed, we assess Stages~I–III as follows. \par

\emph{Agent initiative.} Stage-I is purely reactive; Stage-II grants situational latitude but lacks a principled budget; Stage-III pushes strong local proactivity (tools, reflection, role protocols) yet still lacks a \emph{system policy} to pace, cap, or re-allocate initiative across the team; consequently, initiative often spikes in uncoordinated bursts (many probes at once), driving resource contention and requiring brittle rollbacks when conditions drift. \emph{Directability.} Stage-I offers crude hard switches; Stage-II escalates to policy-based retasking but remains coarse; Stage-III relies on moderators/voting with variable latency. Across all, there is no instantaneous, auditable reconfiguration that flips \emph{families} of behaviors while guaranteeing safe fallbacks. \emph{Policies.} Stage-I bakes static rules; Stage-II mixes norms with light policy toggles; Stage-III cycles playbooks and prompts ad hoc. None offers \emph{formal guards} over whole classes of behaviors system-wide nor a \emph{promotion} mechanism that ties updates to tests and reviews. \emph{Norms.} Operator habits (Stage-I) and team conventions (Stage-II) evolve toward protocolized roles with memory (Stage-III), but practice remains ephemeral: no institutionally stabilized record that accumulates and hardens into reusable assets. \emph{Common ground.} Status dashboards (Stage-II) and shared buffers (Stage-III) help, yet they fragment under scale; there is no \emph{single backbone} with a clear split between provisional and validated knowledge, nor a gate that advances the former into the latter. \emph{Planning horizon.} From single-step (Stage-I) to explicit multi-step (Stage-II) to emergent multi-step (Stage-III), plans still tangle when many probes run at once; there is no \emph{layered control} that decides what should be explored concurrently versus reused directly. \emph{Communication.} Signals $\to$ language+GUI $\to$ multi-modal traces (Stages I$\to$III) improves observability, but without \emph{capacity-aware} pacing and typed feedback paths, humans face overload and loss of auditability.\par

\textbf{From human–AI agent contrasts (Table~\ref{tab:ai-vs-human}).}~LLM agents are backward-looking estimators and \enquote{statistical engines,} while humans supply forward-looking, theory-based causal framing and bear moral responsibility. Across Stages I–III, there is no explicit \emph{tacit inlet} nor \emph{teach-back} that turns \enquote{knowing how} into \enquote{knowing that} with justification; no sustained \emph{knowledge cybernetics} (retrieval $\to$ reflection $\to$ consolidation) to counter fragmentation; no robust \emph{anti-sycophancy} scaffolds; and no UI discipline that respects human chunk limits and transfer times. Ethical locus remains gestural; consolidation remains piecemeal; stewardship of common ground is under-specified. \par

\textbf{Missing unifications.}~Empirical frameworks to date lack (i) a \emph{unified ontology} of agent/agenthood/agency that survives cloning, swapping, and re-wiring under fluid boundaries, and (ii) a \emph{unified formalism} for HAACS that makes concurrency, synchronization, ownership, and cross-boundary effects explicit and analyzable. This motivates the boundary-centric ontology (Secs.~\ref{sec:agenthood-ontology}-\ref{sec:interface-oriented-agenthood}) synthesized with cybernetics (Sec.~\ref{sec:cyber-agenthood}), and the Petri net family that models token flows, guards, rates, and collaboration transitions (Sec.~\ref{sec:sys-formal-pn}–\ref{sec:three-layer-haacs}). \par

\textbf{Stage-IV design brief.}~To overcome these deficits, the \emph{emerging} Stage-IV must (a) \emph{budget initiative} under a system policy that continuously trades off surveying the frontier and reusing what works; (b) enable \emph{instant guard flips} with logged reconfiguration and safe fallbacks; (c) operate a \emph{two-band knowledge store} (provisional/validated) gated by \emph{Test+Peer Review}; (d) coordinate many simultaneous probes while keeping routine work short-circuitable; (e) pace humans via \emph{capacity-aware}, typed feedback paths; and (f) operate a system-wide mechanism that turns proven procedures into \emph{shared, auditable, and reusable} assets, with controlled evolution. Concretely, this is realized by a layered control surface (meta, agent, execution), a system-wide backbone with promotion gates, and policy-controlled initiative, i.e., the HE\textsuperscript{2}-Net frame developed in Secs.~\ref{sec:three-layer-haacs} and \ref{sec:he2-design-pn}.

\subsubsection{Roadmap and contributions: what follows} \label{sec:roadmap-and-built}
\textbf{1) Unifying \emph{what an agent is}.} We introduce a boundary-centric ontology of agenthood that secures \emph{Identity, Wholeness, Autonomy, Continuity} under fluid composition (Sec.~\ref{sec:agenthood-ontology}-\ref{sec:interface-oriented-agenthood}), and synthesize it with Wiener–Simon cybernetics (Sec.~\ref{sec:cyber-agenthood}). This clarifies where \enquote{agent} begins/ends, how it maintains itself, and what counts as the \enquote{same} entity when modules move. Moreover, as shown in Tables~\ref{tab:agent-control-p1}-\ref{tab:agent-control-p2}, contemporary agentic-AI patterns instantiate this Wiener–Simon cybernetics illustrated in Fig.~\ref{fig:cyber-loop}.\par

\textbf{2) Unifying \emph{how HAACS is modeled and governed}.} We formalize HAACS with colored and interpreted Petri nets that (i) make concurrency/synchronization explicit, (ii) assign \emph{ownership} to transitions and places, (iii) support \emph{collaboration transitions} with multi-party enabling, and (iv) separate token-level coordination from stateful updates via guards and operations (Sec.~\ref{sec:sys-formal-pn}). On top, a three-level orchestration (meta, agent, execution) provides guard-based reconfiguration, boundary management, and skill sub-nets (Sec.~\ref{sec:three-layer-haacs}); we show interoperability with contemporary protocol stacks (e.g., MCP, A2A, ACP, ANP, PXP, LOKA)~\citep{yang2025survey,ehtesham2025survey} without ad-hoc glue (Sec.~\ref{sec:pn-fit-extant-work}). \par

\textbf{3) Making cognition and meta-cognition first-class.} We instantiate \enquote{Sense-Plan-Act} loops with learning/critique/reflection as coordinated sub-nets and show how monitoring and directing (meta-cognition) writes justified updates back to memory (Sec.~\ref{sec:metacog-pn}), aligning with the human baselines in Sec.~\ref{sec:cog-metacog}. \par

\textbf{4) Building the knowledge side: from epistemic stance to collaborative learning.} We lay out the epistemic lineage and two knowledge representation paradigms (correspondence vs coherence) to motivate \emph{knowledge cybernetics} (See Tables.~\ref{tab:epistemic-triad}-\ref{tab:corresp-vs-coher}), then construct a CT +SECI collaborative learning architecture with teach-back gates, a shared micro-world, and an evolving backbone (Sec.~\ref{sec:col-hai-learn}). We position memory systems such as Mem0/Mem0g~\citep{chhikara2025mem0} inside this scaffold and identify the missing governance they require (Sec.~\ref{sec:mem0-in-collearn}). \par

\textbf{5) From closed-ended control to practice-guided discovery.} We formalize the limits of MEA for routine control (see Secs.~\ref{sec:mea-frame-intro}-\ref{sec:mea-limit}) and design a constructivist frame for \emph{open-ended} problem solving driven by praxis–cognition dynamics and tacit foreknowledge (Sec.~\ref{sec:practice-oe-ps}), mapping recent AI-for-Science systems to our operators and gates (Sec.~\ref{sec:ai4sci-work2frame}). \par

\textbf{6) The HE\textsuperscript{2}-Net.} Using a systems placement (Sec.~\ref{sec:sys-ontology-haacs}), Stage-IV is viewed as open complex systems; we derive an \emph{explore–exploit} principle from Stage-IV system characteristics and \enquote{bounded rationality}~\citep{simon1993decision,simon1996sciences}, and realize it as a meta-level policy that (i) budgets initiative, (ii) flips guards over \emph{families} of transitions, and (iii) runs a two-band knowledge backbone with a \emph{promotion gate} (Sec.~\ref{sec:he2-design-pn}, Sec.~\ref{sec:he2-intro}). Concretely, HE\textsuperscript{2}-Net is a policy-parameterized instantiation of the three-layer HAACS formalism built in Sec.~\ref{sec:three-layer-haacs}: the \textbf{execution level} integrates the CT+SECI collaborative-learning architecture and the constructivist, open-ended problem-solving frame (with MEA for routine exploitation and teach-back gates) (Sec.~\ref{sec:col-hai-learn}, Sec.~\ref{sec:practice-oe-ps}); the \textbf{agent level} is \emph{instrumented} with the two-band knowledge backbone and validated-store retrievals (Sec.~\ref{sec:three-layer-haacs}, Sec.~\ref{sec:he2-intro}); and the \textbf{meta layer} is \emph{parameterized} by the explore–exploit controller that allocates probe budgets and rate-limits exploitation (Sec.~\ref{sec:he2-design-pn}). We then position Stage-IV against Stages I–III (Table~\ref{tab:stage-comparison-7d}), making explicit how the new stance satisfies the design requirements revealed in Table~\ref{tab:ai-vs-human} and the seven-dimension spine defined in Section~\ref{sec:pre-stages-shorts}. \par

\textbf{7) Beyond the core.} We sketch extensions inspired by biological cybernetics and non-equilibrium dynamics, e.g., autopoiesis/autogenesis, evolving boundaries, multi-scale goals, synergetics, and self-organization, so the framework remains fit as techniques progress and human–AI symbiosis deepens (Sec.~\ref{sec:he2-noneq-dynamics}). \par

Together, these elements provide (i) a \emph{unified ontology} of agents, (ii) a \emph{unified formalism} of HAACS, and (iii) a \emph{design pattern} (HE\textsuperscript{2}-Net) that closes the gaps of Stages I–III while foregrounding human contributions identified in Table~\ref{tab:ai-vs-human}.

\section{Boundary centric agent definition and Petri net formalization} 

\subsection{A preliminary ontology for distinctive agenthood}\label{sec:agenthood-ontology}
\subsubsection{Reevaluating agenthood through non-interchangeable identity} The essence of an \emph{agent} must involve a clearly delineated, non-interchangeable identity. To our knowledge, current single- and multi-agent configurations provide no mechanism for non-interchangeable identity; high modularity in memory, LLM-based reasoning, and tool sets enables cloning and component swap without loss of function. A specific agent can be replicated, and its components interchanged across different instances, suggesting these entities do not qualify as true agents in the way humans do.\par 
Humans possess spatio-temporal identification: we humans can only act in one place at a time, inheriting memories and behavioral patterns continuously, in a manner that is not merely modular. Our functional subsystems exhibit latent yet uniquely correlated structures that are absent in current AI-based agent designs. Although these AI agents can operate tools to affect the physical world, they lack an entity conferring genuine uniqueness.\par
So, how might one construct a parallel form of \emph{uniqueness} within digital agents, something akin to human \emph{wholeness} with its strong interdependence among internal subsystems? The organ for memory in a human cannot be simply removed while retaining full reasoning and tool-use capability; yet, this sort of modular swapping is possible in AI agents. One might counter that LLMs already contain internal, persistent memories that enable basic reasoning even without external memory. But in humans, removing external or sensory-based memory systems (to isolate \emph{internal} memory) without impairing normal functioning is not straightforward, if not impossible.
\par
Therefore, the fundamental question emerges: how do we define, and what theoretical criteria or frameworks can we employ, to categorize and formalize concepts like \emph{uniqueness}, \emph{wholeness}, and \emph{autonomy} for humans, agents, human collectives, and multi-agent systems? This naturally transitions us into discussing the broader landscape of agentic AI and AI agents.
\subsubsection{Mapping the four pillars of an agentic ontology} We propose a robust ontology of \emph{agenthood} hinges on four conceptual pillars, i.e., \emph{Identity}, \emph{Wholeness}, \emph{Autonomy}, and \emph{Continuity}. Identity highlights an entity's self-maintenance and non-interchangeability, drawing on ideas of autopoiesis of second-order cybernetics to contrast humans~\citep{maturana1990biological}, whose singular \emph{self} endures despite biological changes, with LLM-based agents that can be copied or forked at will. Wholeness focuses on how deeply subsystems are integrated: human cognition is tightly coupled (memory, emotion, reasoning and even physical states for embodied cognition interweave inseparably), whereas many AI agents rely on modular, replaceable components. Autonomy differentiates self-generated human goals, rooted in biology and social context, from AI agents whose objectives typically derive from user prompts or preset performance targets for specific tasks. Finally, Continuity underscores that humans live a single, spatio-temporally anchored existence absorbing external information all the time, while AI agents may run on multiple servers or be paused and resumed, complicating the notion of a unified personal history.\par
These four pillars form the basis for a formal ontology of agenthood that can be used to evaluate or design artificial entities. Practitioners may ask: does an agent exhibit unique identity that cannot be trivially merged or replicated? Are its core functions (e.g., memory, reasoning) so interwoven that removing one undermines its entire existence? Do its goals arise from within, or are they externally imposed? And is the agent's trajectory anchored in a continuous timeline, or can it be cloned and parallelized without fracturing its identity? Addressing these questions pushes us beyond surface-level descriptions, forcing designers and researchers to specify precisely where along each dimension an AI agent stands compared to human-like singularity. In this way, the ontology doubles as both theoretical lens and practical tool.
\subsubsection{Practical strategies for deeper human-like uniqueness} For those seeking to create digital agents with deeper \emph{human-like} uniqueness, several strategies emerge. Immutable identity anchors, such as cryptographic ledgers, can ensure copying an agent results in a distinct new entity rather than a seamless clone. Strong embodiment constraints, enforcing a single computational environment, can foster continuity and limit casual replication. Systems that support self-maintenance and self-regulation approximate autopoiesis, letting agents adapt or repair themselves without external module swapping. Finally, bounded perception-action loops ground each agent's unique experiences in a singular world, such that any trivial duplication of its personal history disrupts the agent's continuity. Though such measures complicate engineering, they underscore the profound difference between mere plug-and-play software and truly integral, singularly continuous agenthood. 
\subsection{Interface-oriented agenthood under fluid boundaries}\label{sec:interface-oriented-agenthood}
Extending our preliminary ontology, the core difficulty is that identity, wholeness, autonomy, and continuity cannot be stably attributed to engineered agents when a fluid agent–environment boundary permits cloning and modular reconfiguration.
\subsubsection{Dilemma of agent modeling: Ship of Theseus and fluid boundaries} Much like the \enquote{Ship of Theseus} paradox, where planks are replaced over time, causing us to question whether it remains the \emph{same} ship, defining an agent becomes tricky when its modules are swapped or reconfigured. Traditional multi-agent models often assume a stable boundary around sensing, decision-making, and acting components, treating each agent as a discrete, autonomous unit. But edge cases abound: a two-headed agent with dual LLM cores sharing resources can act as one cohesive unit or as two separate agents, depending on how we group or label its internal processes.\par
Similarly, fluid boundaries in reconfigurable cognitive architectures defy strict partitioning~\citep{sumers2024cognitive}; the agent can share modules with the environment or with other agents in ways that challenge the classical notion of self-contained autonomy. These prevalent methods, ranging from purely task-focused definitions (e.g., grouping by function) to intentional approaches (e.g., assigning beliefs, desires, and intentions), highlight the ongoing dilemma: does an agent remain \emph{itself} under continuous reassembly of internal components, or should we define it differently each time the boundary shifts?
\subsubsection{Adaptive interface dynamics in goal-seeking systems}
A goal-seeking adaptive system focuses on the \emph{interface} between its \emph{inner} and \emph{outer} environments, achieving its goals by adapting the former to the latter through \emph{interface} \citep{simon1996sciences}. Although the inner environment (akin to hardware) is relatively simple, the system's complexity emerges from the richness of the outer environment. A thinking human being is such an adaptive system: an individual's goals define the interface between their inner and outer environments. With memory stores in play, human behavior reflects the characteristics of the outer environment in light of a person's goals, revealing only a few limiting properties to enable a person's thinking. Thus, the apparent \emph{complexity} of human behavior over time reflects the \emph{complexity} of that external setting. By analogy, we treat an AI agent as a goal-seeking adaptive system whose \emph{interface} regulates the coupling between its \emph{inner} and \emph{outer} environments.
\subsubsection{Adopting Simon's interface-oriented view}
To transcend this dilemma, Simon's interface-oriented approach offers a powerful guiding principle: rather than fixating on an agent's internal composition, one focuses on its boundary of interaction with the external environment. From a living systems perspective, we can treat all internal processes and structures, however many heads or reconfigurable sub-parts they possess, as belonging to a single \emph{artifact}, provided they share a coherent external \emph{interface} for exchanging information, matter, or energy~\citep{miller1965living,miller1978living}. Under this framework, the places and transitions of Petri nets can be used to define how the agent's internal states are coupled and how the agent exerts influence on the outside. Whenever a module remains within the boundary, it is \emph{internal}. If it can be directly accessed by external forces publicly \citep{sumers2024cognitive}, it becomes part of the environment. This interface-oriented method gracefully handles \enquote{Ship of Theseus} scenarios: no matter how many planks (modules) are replaced, so long as the overarching interface and its capacity to control or be controlled, i.e., self-regulate, remains coherent, we can still call it the \emph{same} agent.
\subsection{Cybernetic foundations and their synthesis with an interface-oriented agenthood} \label{sec:cyber-agenthood}
\subsubsection{Norbert Wiener's canonical loop} 
\begin{figure}[htbp]
  \centering
  \includegraphics[width=1.0\linewidth]{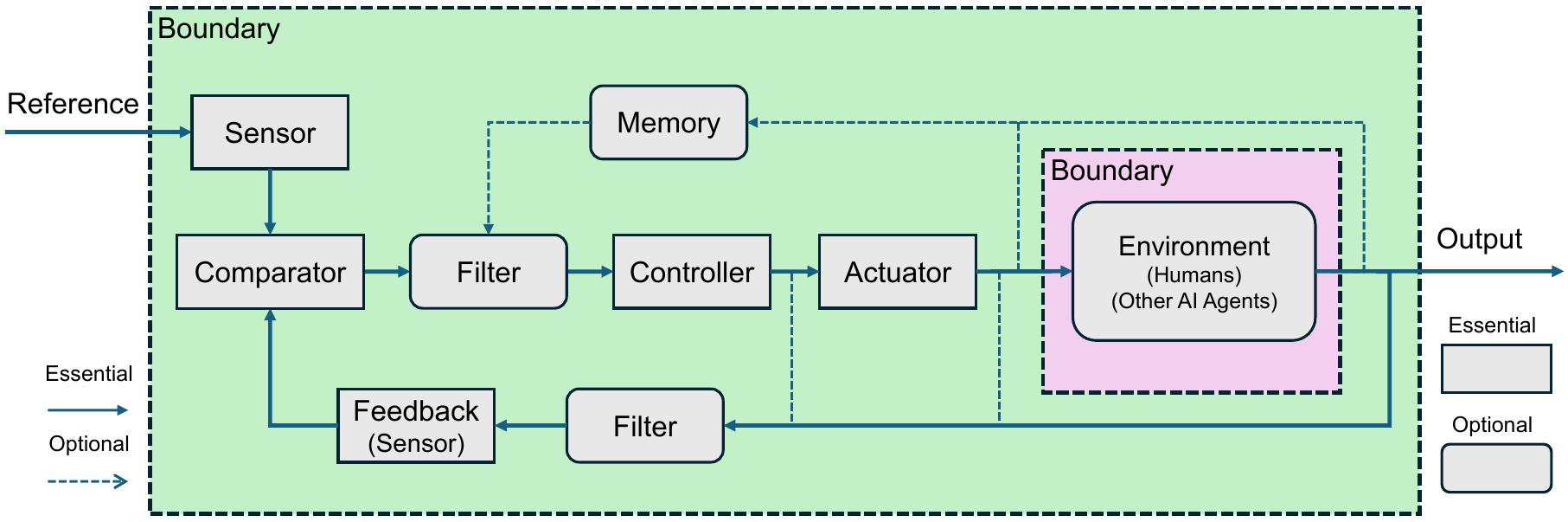} 
  \caption{A high-level illustration of a single AI agent under fluid boundaries viewed through Wiener's and Simon's lens. Solid blocks/arrows are essential; dashed ones are optional extensions. The green region (outer dashed rectangle) encloses one agent whose internal subsystems, i.e., Sensor, Comparator, Controller, Actuator, and optional Memory/Filter(s), form a closed-loop regulator. Dashed arrows mark optional information pathways (e.g., memory recall, filtered feedback from internal subsystems). The violet block exemplifies the environment, humans or other AI agents that lie outside the agent's boundary yet couple to it through the Actuator--Feedback channel (\emph{shared interface}): actions flow outward, observations and rewards flow back.}
  \label{fig:cyber-loop}
\end{figure}

Cybernetics casts every goal-directed artefact as a closed, negative-feedback regulator composed of (i) a \emph{Sensor} that samples the world and the reference signal, (ii) a \emph{Comparator} that derives an error, (iii) a \emph{Controller} that transforms error into control commands, (iv) an \emph{Actuator} that couples commands back to the world, and optionally but typically (v) \emph{Memory/Filter} modules that stabilize or anticipate~\citep{wiener1959man,wiener1988human,wiener2019cybernetics}. This five-piece skeleton formalizes the single imperative of an adaptive system: reduce deviation from purpose under disturbance (see Figure~\ref{fig:cyber-loop}).
\subsubsection{Locating contemporary agent patterns inside the loop} The agent architectures surveyed earlier instantiate Wiener's diagram with domain-specific choices for each block. Chain-of-Thought~\citep{wei2022chain} and Self-Consistency~\citep{byerly2024effective,chen2024two} read the prompt as Sensor, apply an open-loop Controller (the prompted LLM), and actuate by emitting text; filters are absent, hence no mid-course correction. Tree-of-Thought~\citep{yao2023tree} adds an internal deliberative Filter/Memory, iteratively pruning a search tree before acting. ReAct~\citep{yao2023react}, CRITIC~\citep{gou2023critic}, and WebAgent~\citep{webagent2024} go further, wiring external tool results or HTML snippets into the Comparator, thereby closing the loop inside a single episode. Outer-loop learners such as STaR~\citep{zelikman2022star}, Reflexion~\citep{shinn2023reflexion}, shift the Comparator to trajectory-level rewards and update the Controller across trials. These correspondences evaluated against the high-level diagram in Figure~\ref{fig:cyber-loop} are summarized concisely in Tables~\ref{tab:agent-control-p1}-\ref{tab:agent-control-p2}.
\begin{table}[htbp]
\centering
\small
\setlength{\tabcolsep}{2.8pt}
\begin{adjustbox}{scale=0.88,center}
\begin{tabularx}{\textwidth}{@{}
p{2.4cm}  
Y
Y
Y
p{1.8cm}  
p{1.6cm}@{}}  
\toprule
\textbf{Agent Arch.} & \textbf{\makecell[l]{Sensors \&\\Feedback}} & \textbf{Controller} & \textbf{Memory} & \textbf{Actuators} & \textbf{\makecell[l]{Control\\scheme}}\\
\toprule
Chain-of-Thought (CoT)~\citep{wei2022chain} & prompt context & prompted LLM (few-shot exemplars) & context window & text output & feedforward (open-loop)\\
\addlinespace
Tree-of-Thoughts (ToT)~\citep{yao2023tree}& partial thoughts + self-scores & search over thoughts (BFS/DFS; critic–reflect) & running tree of candidates & next-thought pick & deliberative look-ahead w/ pruning\\
\addlinespace
Self-Consistency (SC)~\citep{byerly2024effective,chen2024two} & multiple sampled CoTs & majority voting & transient candidate set & final answer pick & output-ensemble selection (feedforward)\\
\addlinespace
USC (Universal SC)~\citep{chen2024universal} & concatenated candidates & LLM-as-judge similarity & candidate pool in-prompt & holistic choice & semantic re-ranking (feedforward)\\
\addlinespace
SELF-REFINE~\citep{madaan2023self} & self-critique signals & draft $\rightarrow$ critique $\rightarrow$ refine & outputs \& feedback history & revised output $y_{t+1}$ & single-loop feedback (cascade)\\
\addlinespace
CRITIC~\citep{gou2023critic} & external verification (search/code) & verify $\rightarrow$ critique $\rightarrow$ correct & evidence-anchored critiques & corrected text/code & closed-loop feedback w/ verification\\
\bottomrule
\end{tabularx}
\end{adjustbox}
\caption{Cybernetic Typology of Contemporary Agentic Control Patterns - Part I}
\label{tab:agent-control-p1}
\end{table}

\begin{table}[htbp]
\centering
\small
\setlength{\tabcolsep}{2.8pt}
\begin{adjustbox}{scale=0.88,center}
\begin{tabularx}{\textwidth}{@{}
p{2.4cm}  
Y
Y
Y
p{1.8cm}  
p{1.6cm}@{}}  
\toprule
\textbf{Agent Arch.} & \textbf{\makecell[l]{Sensors \&\\Feedback}} & \textbf{Controller} & \textbf{Memory} & \textbf{Actuators} & \textbf{\makecell[l]{Control\\scheme}}\\
\toprule
ReAct~\citep{yao2023react} & stepwise \enquote{Observation} & Thought–\allowbreak Action–\allowbreak Observation loop & full chain in prompt & tool/action calls & closed-loop reactive feedback\\
\addlinespace
RAISE~\citep{liu2024llm} & recent turns; retrieved examples & triggers orchestrate Scratchpad + Examples & scratchpad (short-term) + retrieval (long-term) & structured next prompt & single-loop w/ observer/filter\\
\addlinespace
Reflexion~\citep{shinn2023reflexion} & trajectory reward/\allowbreak diagnostics & evaluator + self-reflection feedback & long-term reflection buffer & next-trial actions & iterative outer-loop feedback across trials\\
\addlinespace
MM-REACT~\citep{yang2023mm} & multimodal observations & reasoning–\allowbreak action–\allowbreak reasoning loop & interleaved text + vision outputs & vision tool calls + textual integration & closed-loop multimodal ReAct\\
\addlinespace
WebAgent~\citep{webagent2024} & full HTML (observations from real websites) + instruction & planner (decomposing instructions into sub-instructions) + code generator & long-HTML snippets; history; self-experience supervision & Python/\allowbreak Selenium programs & filtered closed-loop planning\\
\addlinespace
Prompt Engineering (PE) as Optimal Control~\citep{luo2023prompt} & prior responses, task, eval & optimize sequential prompting to maximize final-answer utility under interaction cost & prompt candidates set; dialogue history & prompts as control inputs & closed multi-round optimal control\\
\addlinespace
STaR (Self-Taught Reasoner)~\citep{zelikman2022star} & Q–A correctness & iterative \emph{learn by explaining} fine-tuning & collected rationales & updated weights & offline policy improvement\\
\bottomrule
\end{tabularx}
\end{adjustbox}
\caption{Cybernetic Typology of Contemporary Agentic Control Patterns - Part II}
\label{tab:agent-control-p2}
\end{table}
\subsubsection{Wiener \emph{meets} Simon} Wiener’s cybernetics supplies the \emph{how}: a negative-feedback
communication–control circuit (Sensor $\!\to\!$ Comparator $\!\to\!$ Controller
$\!\to\!$ Actuator $\!\to\!$ Feedback) that scales hierarchically to ever-larger regulatory layers and, beyond merely grounding actions in a physical or digital environment, accommodates multi-agent and human-in-the-loop scenarios by exchanging signals (e.g., set-points, broader goal directives, and status and error reports) with humans or other AI agents, each construed as a peer regulator coupled through the shared interface (referring to part of the inner dashed boundary enclosing the purple region in Figure~\ref{fig:cyber-loop}). Simon, by contrast,
supplies the \emph{where}: an agent is whatever portion of that circuitry
is enclosed by a stable, auditable \emph{interface} that mediates
information, matter, or energy with the outside. We therefore define an \emph{AI agent} as a \textbf{\emph{boundary-centric}, \emph{goal-seeking adaptive system}} that has the necessary \emph{autonomy} primitives summarized in Table~\ref{tab:agent-autonomy}.
\begin{table}[htbp]
\centering
\small
\setlength{\tabcolsep}{4.2pt}
\begin{tabularx}{\textwidth}{@{}p{4.2cm}X@{}}
\toprule
\textbf{Autonomy primitive} & \textbf{Capability}\\
\midrule
Self-maintenance & self-maintain its organization, i.e., the network of components and their relations, and resources (\emph{autopoiesis}, see \citealp{maturana2012autopoiesis}) to preserve the agent’s structural wholeness over long horizons.\\
\addlinespace
Self-regulation & self-regulates key variables through \emph{homeostatic} feedback for internal stability and servo-control feedback for goal-directed error-correction~\citep{wiener1959man,wiener1988human,wiener2019cybernetics}.\\
\addlinespace
Self-replication & can self-replicate / spawn successor instances (\emph{autogenesis}, see \citealp{maturana2012autopoiesis}) when conditions allow.\\
\addlinespace
Coordination & engages in coordinated signaling and action with peer agents via its interface.\\
\bottomrule
\end{tabularx}
\caption{Necessary autonomy primitives for the boundary-centric, goal-seeking adaptive system (the AI agent)}
\label{tab:agent-autonomy}
\end{table}
This synthesis resolves the \enquote{Ship-of-Theseus} dilemma of Section~\ref{sec:interface-oriented-agenthood} and operationalizes the four criteria of Section~\ref{sec:agenthood-ontology}; specifically, our boundary-centric, goal-seeking adaptive system satisfies them as follows:
\begin{itemize}
\item \textbf{Identity.}  
      Each agent is uniquely specified by its \emph{boundary}, i.e., the totality of sensor/actuator channels \emph{interfacing} with the environment, humans, and other AI agents, and by a cryptographically chained, monotonically time-stamped event log that serializes all input/output (I/O) and internal mutations. This boundary may be \emph{reconfigured} (e.g., shrunk, enlarged, migrated, or policy-updated), as the interaction repertoire adapts to changing outer conditions, yet the agent remains the \emph{same} provided the autonomy primitives in Table~\ref{tab:agent-autonomy} remain satisfied. Internal modules reside \emph{within} this boundary and may be updated or reconfigured only via the agent’s own signed reconfiguration events; opening a parallel process that yields a second, diverging log (or exposes new public channels without carrying the prior ledger) constitutes a \emph{new} identity (a clone), not a continuation.
\item \textbf{Wholeness.}  
    The cybernetic subsystems (Sensor, Comparator, Controller, Actuator, and optional Memory/Filters) are jointly indispensable under \emph{coupling constraints} (e.g., shared calibrations, latency and gain budgets, safety/consistency contracts). While housed inside the boundary they are \emph{non-interchangeable} in the following precise sense: any removal or replacement that bypasses the agent’s authorized reconfiguration protocol, i.e., without co-recalibrating coupled subsystems and updating the interface contracts, breaks the closed loop and dissolves the organization that sustains the agent’s function. Permissible refits occur only through that protocol, which enforces the coupling constraints.
\item \textbf{Autonomy.}  
    Goal representation/formation sits inside the loop and supplies the references and constraints that drive the primitives in Table~\ref{tab:agent-autonomy}.
\item \textbf{Continuity.}  
    The single, auditable event log anchors a continuous, causally ordered history even if execution is paused and later resumed on different hardware; reconfigurations append to the same ledger. Parallel execution that produces two \emph{diverging} logs, however, yields two distinct identities.
\end{itemize}
Thus an \textbf{AI agent} in our ontology is a boundary-centric, goal-seeking adaptive system that self-maintains, self-regulates, can self-replicate, and coordinates with peers via its \emph{interfaces}, preserving Identity, Wholeness, Autonomy, and Continuity by maintaining an unbroken \emph{cybernetic loop within a stable, auditable boundary}, whose I/O interfaces constitute the public-facing portion through which it exchanges \emph{information, matter, or energy} with the environment, humans, and other AI agents.

\subsection{Modeling multi-agent systems with Petri net formalisms} \label{sec:sys-formal-pn}
\subsubsection{Ad hoc approaches and their limitations}
In the multi-agent implementations discussed in Section~\ref{sec:previous-hai}, the interplay between agents is often described through ad-hoc, empirical approaches that omit the explicit and detailed accounting of how \textbf{information}, \textbf{matter}, and \textbf{energy} flow within and among agents. These implementations typically focus on high-level behaviors (e.g., whether agents cooperate or compete to achieve tasks) but rarely capture the precise transformations and transfers underpinning such behaviors. As a result, crucial elements, like the actual resource constraints in each agent, the energy costs of communication, and the storage, dissipation, or conversion of matter, remain hidden or approximated in ways that limit both explanatory depth and predictive power. When these approaches do try to account for resource flows, they often do so superficially, aggregating them into generic \enquote{cost functions} or \enquote{utility metrics} that only approximate or partially capture the underlying processes.\par
Without a formal foundation, these ad-hoc methods risk producing fragmented, inconsistent, or non-generalizable results. For instance, the same basic modeling architecture might be used across different interaction scenarios without ensuring that core resource-flow principles (e.g., conservation of matter or the transformation of energy into work) are consistently upheld. Moreover, \textbf{concurrency}, \textbf{synchronization}, \textbf{asynchrony}, \textbf{parallelism} and \textbf{sequentiality} as fundamental aspects of multi-agent communication and coordination, tend to be treated in an implicit manner, leaving potential conflicts or rare conditions underexplored. This patchwork approach also complicates comparative analysis across different benchmarks, since each one might incorporate its own arbitrary assumptions about resource handling, message passing, and agent \emph{states}.

\subsubsection{A formal framework with Petri net formalisms}
In contrast, Petri net-based formalisms offer an explicit and rigorous way to model these dynamics. By specifying which \enquote{places} (states) store information, matter, or energy, and which \enquote{transitions} transform or exchange these resources, Petri nets capture the structural and causal details of multi-agent interactions. This enables systematic analysis of properties such as deadlocks, resource bottlenecks (e.g., modelling bounded rationality \citealp{simon1996sciences} explicitly), or synchronization issues, i.e., properties that may remain opaque or undiscovered in less formal modeling frameworks. Furthermore, Petri nets allow for easy extension and clear modularization, ensuring that each agent's internal conversions (information, matter, and energy) among inner modules can be clearly described at the same time as the collective interplay emerges from the transitions, whether shared or privately owned, that occur among agents. The final framework will serve as not only a more precise representation of the system's mechanics but also a robust foundation for comparing and contrasting different multi-agent approaches in a consistent, theoretically grounded manner.\par
In Petri nets, \enquote{tokens} flow through transitions according to certain rules. It's relatively straightforward to extend the token colors so they represent or carry different \enquote{substances} for HAACS: information tokens (e.g., messages, data, knowledge); matter tokens (e.g., physical resources); energy tokens (e.g., consumable energy capacity); note that each transition can then transform tokens of a certain type into tokens of another type, capturing the \emph{interplay} of information, matter, and energy occurring in the processes of \citep{miller1978living}. And the built-in concurrency and the rich analysis techniques behind Petri nets can be utilized to reason about the properties such as reachability, liveness, synchronization, and resource usage.
\subsubsection{Interface-oriented human-AI agent collaboration in Petri nets} \label{sec:initial-haacs-pn}
An agent sustains itself by upholding a dynamic interface, i.e., its boundary, through which it exchanges information, energy, or matter with the outside, mirroring the essential information-energy-matter processes to support living systems in \citep{miller1965living}. Even being idle without imposing control on the outside, the agent must preserve its own integrity via processes akin to autopoiesis or autogenesis. Analogically, a computational agent \emph{lives} in RAM when inactive and may be recycled once its threads complete, but as long as it retains the internal capacity to maintain itself, it remains an autonomous entity.\par
The overall HAACS is represented by a colored Petri net:
\begin{equation}
\mathcal{C}N = (P, T, A, \Sigma, \kappa, \mathrm{Pre}, \mathrm{Post}, M_0), \label{color-PN-1}
\end{equation}
where $P$ is a finite set of places. $T$ is a finite set of transitions. $A$ is a finite set of arcs connecting places and transitions (input and output arcs). $\Sigma$ is a global (or multi-sorted) color universe, i.e., the union of all color types that might be used (e.g., \emph{information}, \emph{matter}, \emph{energy}, etc.). $\kappa : P \to 2^\Sigma$ assigns each place a color set $\kappa(p) \subseteq \Sigma$; tokens in place $p$ must belong to $\kappa(p)$. \par
$\mathrm{Pre}$ and $\mathrm{Post}$ define how tokens flow: Each arc $p\to t$ or $t\to p$ has an associated arc expression that describes how many tokens (and of which colors) are produced/consumed. For simplicity in matrix form, we can store integer counts if we treat color matching externally. In more advanced definitions, $\mathrm{Pre}(p,t)$ or $\mathrm{Post}(p,t)$ can be functions from color sets. $M_0$ is the initial marking, specifying how many tokens of each color reside in each place initially. A marking $M$ is thus a function $M: P \to \mathcal{M}_{\kappa}$,
which tells us the multi-set of colored tokens in each place.
\subsubsection{Partition and ownership of transitions and places} \label{sec:basic-pn-model}
We partition the entire net among $n$ agents $\{A_1,\dots,A_n\}$, plus an environment $E$, plus a human $H$, which can be extended to a group. Instead of having a separate \enquote{interface transition} class, each transition $t \in T$ is owned by exactly one entity:
\begin{equation}
    \text{Owner}: T \rightarrow \{A_1,\dots,A_n,E,H\} \label{eq:simple owner}
\end{equation}
Note that the transitions ownership is essential for supervisory control. If $\text{Owner}(t)=A_i$, agent $A_i$ controls that transition. If $\text{Owner}(t)=H$, then the human decides if/when $t$ fires, subject to the marking and color constraints. If $\text{Owner}(t)=E$, it is an environment-driven transition. Similarly, each place $p$ belongs to either an agent $A_i$, the environment $E$, or the human $H$. For instance, $P_H\subset P$ are the human-owned places (storing tokens representing human knowledge, tasks, decisions, etc).\par
A transition can consume or produce tokens in places belonging to different agents (or environment or human), but responsibility for enabling that transition is assigned to a single owner. The boundaries between agents, the environment, and humans are represented by transitions whose input and output places belong to different entities. By default, all input places originate from their corresponding entity, and the boundary transitions (interfaces) are owned by the entity from which the input places originate. This avoids the confusion of separate \enquote{boundary transitions} blocks, because any boundary-crossing transition simply belongs to whichever agent (or environment) can trigger it.\par
Each place \( p \) is assigned either to an agent's internal set \( P_i \) (indicating that the place \emph{belongs} to agent \( A_i \)) or to the environment \( P_E \) or to the human \(P_H\). A place can still contain tokens that move in via transitions owned by other agents, thereby realizing \enquote{boundary crossing}. Thus we have:
\begin{equation}
P = \left(\bigcup_{i=1}^n P_i\right) \cup P_E \cup P_H,\ \
T = \left(\bigcup_{i=1}^n T_i\right) \cup T_E \cup T_H.    \label{eq:p-t-1}
\end{equation}
Reorder \(P\) to group places as $(P_1, \dots, P_n,\, P_H,\, P_E)$, and reorder \(T\) to group transitions as $(T_1, \dots, T_n,\, T_H,\, T_E)$. The incidence matrix \( C \) (of size \( |P| \times |T| \)) breaks down into blocks as follows:
\begin{equation}
C = \begin{pmatrix}
C_{1,1} & \cdots & C_{1,n} & C_{1,H} & C_{1,E} \\
\vdots  & \ddots & \vdots  & \vdots  & \vdots  \\
C_{n,1} & \cdots & C_{n,n} & C_{n,H} & C_{n,E} \\
C_{H,1} & \cdots & C_{H,n} & C_{H,H} & C_{H,E} \\
C_{E,1} & \cdots & C_{E,n} & C_{E,H} & C_{E,E} \label{eq:simple matrice}
\end{pmatrix}.
\end{equation}
If \( C_{i,j} \neq 0 \), it means that transitions in \( T_j \) (owned by \( A_j \)) affect places in \( P_i \) (owned by \( A_i \)). The environment is simply regarded as \enquote{agent \( E \)} from this matrix viewpoint. Each block \( C_{i,H} \) shows how human-owned transitions \( T_H \) produce tokens in agent \( A_i \)'s places \( P_i \). Similarly, \( C_{H,j} \) shows how agent-owned transitions \( T_j \) affect the human's places \( P_H \).
\subsubsection{MIMO transitions and \enquote{collaboration boundary}} \label{sec:mimo-pn}
Consider a multiple-input multiple-output (MIMO) transition \( t \) that has arcs from places in \( \{P_i\} \) (agent \( A_i \)) and from \( \{P_j\} \) (agent \( A_j \)), etc., and produces tokens into places of these same or other agents. In a simple \enquote{single owner} scheme as Eq.~(\ref{eq:simple owner}), we'd assign a transition \( t \) to exactly one agent. But that can be awkward if multiple agents must collaborate for \( t \) to fire (e.g., each must provide a resource or confirm readiness). For example, a transition \( t \) that transfers \enquote{energy} from agent \( A_1 \), \enquote{task info} from agent \( A_2 \), and outputs combined results to both. Firing \( t \) therefore has multi-agent preconditions and multi-agent effects.\par
Instead of forcing single-agent ownership, define a special set of collaboration transitions $T_{\text{coll}} \subseteq T_{MIMO}$ that is incorporated into the original $T$ in Eq. (\ref{eq:p-t-1}). For any \( t \in T_{\text{coll}} \): Inputs: \( t \) may consume tokens from places $\{P_{i_1}, P_{i_2}, \dots\}$, belonging to different agents. Outputs: \( t \) may produce tokens in those same or other agents' places. Collective Enabling: All relevant agents, i.e., those that supply tokens or must authorize their use, must \emph{agree} or \emph{enable} \( t \) to fire. The incidence matrix in Eq. (\ref{eq:simple matrice}) is thus extended to 
\begin{equation}
    C_{MIMO} = \begin{pmatrix}
C_{1,1} & \cdots & C_{1,n} & C_{1,\text{coll}} & C_{1,E} & C_{1,H}\\
\vdots & \ddots & \vdots  & \vdots & \vdots & \vdots \\
C_{n,1} & \cdots & C_{n,n} & C_{n,\text{coll}} & C_{n,E} & C_{n,H} \\
C_{H,1} & \cdots & C_{H,n} & C_{H,\text{coll}} & C_{H,E} & C_{H,H} \\
C_{E,1} & \cdots & C_{E,n} & C_{E,\text{coll}} & C_{E,E} & C_{E,H} \label{eq:mimo matrice}
\end{pmatrix}.
\end{equation}
A column block \(C_{*,\text{coll}}\) is reserved for collaboration transitions. Each column in that block corresponds to a single collaboration transition \(t\), and the row entries indicate how the tokens from each associated agent's places are consumed or produced. For each collaboration transition \(t\), define an ownership set formally as:
\begin{equation}
    \operatorname{Owners}(t) \subseteq \{A_1, \dots, A_n\}. \label{mimo owner}
\end{equation}
That indicates which agents must participate in or consent to \( t \). This is more general than single-agent transitions than that in Eq. (\ref{eq:simple owner}). A transition \(t\) fires in a concurrency step if all agents in \(\operatorname{Owners}(t)\) enable it, and the marking (token distribution) satisfies all input arc constraints (including color checks) across each agent's places. \par
For supervisory control, if a control matrix \(\Gamma \in \{0,1\}^{|T|\times n}\) is used, then for any \(t \in T_{\text{coll}}\):
\begin{equation}
\Gamma[t,i] =
\begin{cases}
1, & \text{if } A_i \in \operatorname{Owners}(t),\\[0.66ex]
0, & \text{otherwise.}
\end{cases}
\end{equation}
At run-time, each agent \(A_i\) follows a local policy \(\sigma_i\). A collaboration transition \(t\) becomes enabled if all \(\sigma_i\) for \(i \in \operatorname{Owners}(t)\) allow it, in addition to satisfying the marking constraints.
\subsubsection{Decoupling computation and control with Interpreted Petri nets}
We extend the colored Petri net presented in Eq. (\ref{color-PN-1}) to the interpreted Petri net so as to unify: (1) \textbf{token-based coordination} (with optional guard logic), capturing interactions occurring in boundaries between agents and environment, and (2) \textbf{data-processing/state updates} (global or agent-local), capturing the internal logic of each agent and state variables of the global system. This yields a two-level design:\par
\textbf{Level 1}: Petri net structure (places, transitions, arcs) for concurrency and synchronization, including \emph{boundary} and \emph{collaboration} interactions between agents.\par
\textbf{Level 2}: Interpreted state variables (data updates), handled by each transition's \enquote{operation} and guard conditions.\par
The approach generalizes neatly to systems (modeled in Sections~\ref{sec:initial-haacs-pn}--\ref{sec:mimo-pn}) with either single-owner transitions or collaboration transitions (multi-owner), as well as colored tokens for message- or resource-passing across boundaries. Although earlier studies, e.g., AutoGen \citep{wu2024autogen}, discussed in Section \ref{sec:previous-hai}) have reported empirical designs that exhibit similar decoupling design, our formalism is independently developed and provides a rigorous theoretical framework for separating computation and control. In contrast to the ad-hoc approaches described in previous work, our method systematically delineates the two aspects, thereby extending the insights initially observed.
\subsubsection{Global and local data variables computation and updates}
An interpreted Petri net (IPN) extends a basic net $(P,T)$ in Eq. (\ref{color-PN-1}) with a collection of data variables that can be: (1) global variables: $\mathbf{X_g}=(x_1,\dots,x_m)$ shared by all agents, (2) local variables: each agent $A_i$ may keep a local vector $\mathbf{X_l}_i=(x_{i1},\dots,x_{i k_i})$. When a transition fires, it can read and write these relevant data variables based on the assigned permission associated. The net effect is that the IPN has two \enquote{layers}: \enquote{Petri Net Layer} (concurrency, token flow, enabling transitions) and \enquote{Data Layer} (each firing updates or checks data variables).\par
Thus we concatenate $\mathbf{X_g}$ and $\mathbf{X_l}$ to get $X$. Provided a guard \(G_t(X,\gamma)\), i.e., a Boolean condition referencing \(X\) (data variables) and, optionally, \(\gamma\) (the colors of tokens on the input arcs, if colored tokens matter). 
\subsubsection{Stepping semantics and conflict resolution}
In discrete-event settings for system expressed in Eq.~(\ref{color-PN-1}), a \enquote{step} can be a set of transitions firing simultaneously (or in rapid succession). We represent the count of how many times each transition fires in a concurrency step by a vector:
\begin{equation}
    \delta(\tau) \in \mathbb{N}^{|T|}, \label{simple-fire-num}
\end{equation}
where $\delta(\tau)[t] = k$ means transition $t$ fired $k$ times in this step. We can partition \(\delta(\tau)\) into sub-vectors \(\delta(\tau_i)\), \(\delta(\tau_H)\) and \(\delta(\tau_E)\). If \(\operatorname{Owner}(t) = A_i\), then \(\delta(\tau_i)[t]\) records how many times transition \(t\) fired; if \(\operatorname{Owner}(t) = E\), it is assigned to \(\delta(\tau_E)[t]\). \(C_{i,\text{coll}}\) is the sub-block of columns for transitions in \(T_{\text{coll}}\). Each collaboration transition column in \(C_{i,\text{coll}}\) indicates how many tokens from \(A_i\)'s places are consumed or produced. Then the total firing vector is
\begin{equation}
    \delta(\tau) = \bigl( \delta(\tau_1),\, \dots,\, \delta(\tau_n),\, \delta(\tau_E),\, \delta(\tau_H),\delta(\tau_{\text{coll}}) \bigr). \label{fire-vec}
\end{equation}
The whole system in Eq. (\ref{color-PN-1}) can evolve in sequences of concurrency steps, each represented by a firing vector \(\delta(\tau)\). We denote
\begin{equation}
M_{k+1} = M_k + C_{MIMO}\,\delta(\tau)^{(k)},
\end{equation}
with \(\delta(\tau)^{(k)}\) being the firing vector at step \(k\). Provided that \(M\) enables all those transition firings simultaneously (accounting for color constraints, token availability, etc.). Because \(t \in T_{\text{coll}}\) is not assigned to a single agent, it does not appear in just one \(\delta(\tau_i)\); instead, it appears in \(\delta(\tau_{\text{coll}})\). The joint policy from \(\operatorname{Owners}(t)\) must collectively enable \(\delta(\tau_{\text{coll}})[t] > 0\). Once it fires, the corresponding incidence matrix columns update each agent's marking as appropriate. 
\subsubsection{Per-agent stepping \(\Delta M_i\)}
If \(P_i \subset P\) are the places owned by agent \(A_i\) (indexed accordingly), we extract the relevant rows from \(C_{MIMO}\) in Eq. (\ref{eq:mimo matrice}) to obtain the sub-block \(C_{i,*}\). Then,
\begin{equation}
\Delta M_i = M_i' - M_i = \sum_{j=1}^{n} C_{i,j}\,\delta(\tau_j) + C_{i,E}\,\delta(\tau_E) + C_{i,H}\,\delta(\tau_H) + C_{i,\text{coll}}\,\delta(\tau_{\text{coll}}).  
\end{equation}
In short, agent \(A_i\)'s local marking changes due to transitions fired by itself (i.e., \(C_{i,i}\,\delta(\tau_i)\)) and transitions owned by other agents or the environment that affect tokens in \(P_i\). Because each \emph{non-collaborative} transition has a single owner, \emph{boundary} interactions appear as non-empty off-diagonal blocks \(C_{i,j}\) for \(i \neq j\) if agent \(A_i\) and \(A_j\) have interactions exemplified in Fig.~\ref{fig:cyber-loop}.\par
For IPN that extends system in Eq.~(\ref{color-PN-1}), an update \(\Delta_t\) for a concurrency step in virtue of the total firing vector (see Eq.~\ref{fire-vec}) that modifies \(X\) is:
\begin{equation}
        \Delta_t: X' \leftarrow f_t(X,\gamma).
\end{equation}
Where $\gamma$ captures the colors of tokens on the input arcs of transitions if colored tokens matter, and $f_t(X,\gamma)$ are update functions within the transitions. The updated data for agent \(A_i\) is given by
\begin{equation}
    X_i' = X_i \oplus \Bigl\{\, \Delta_t(X,\gamma) \;\Big|\; t \in T_i \cup T_{\text{coll}} \text{ fired in this step and } t \text{ updates } X_i \Bigr\},
\end{equation}
where \(\oplus\) denotes the application (or merging) of each relevant update \(\Delta_t\) that modifies \(X_i\). If conflicts such as that the multiple transitions update \(X_i\) simultaneously, a priority policy, a merging rule or partial-order constraints should be defined to resolve conflicts. In practice, one might require transitions updating the same variable to be mutually exclusive or define a merging function.
\subsubsection{Agents interactions across boundaries (i.e., interfaces)}
Internal transitions \(T_{i, int}\) of agent \(A_i\), where \(C_{j, T_{i, int}}\), for \(i \neq j\), only read and update \(X_i\) (plus possibly some global variables) and consume/produce tokens solely in \(P_i\). For an agent $A_i$ whose places are output of boundary and collaboration transitions, i.e., the single-input single-output (SISO) and single-input multiple-output (SIMO) boundary transitions, as well as MIMO collaboration transitions, may consume tokens from multiple other agents' places (and possibly human and environment places) and may read or write data variables in \(X_i\).\par
If a transition belongs to a single agent \(A_k\), then that agent has the final authority on its enabling. If a transition belongs to the \emph{collaboration set} \(T_{\text{coll}}\), each agent with a resource or data stake must enable it. This distinguishes an agent's \emph{internal} net sub-structure from the transitions that cross boundaries to other agents, the human or the environment. Through the boundary transitions, the agents can synchronize (e.g., handshake or protocol transitions) to cooperate, and compete for the same resource, leading to transitions that represent resource locking or conflicts.
\subsection{Hierarchical Petri net formalism of HAACS} \label{sec:three-layer-haacs}
A three-layered hierarchical Petri net architecture is constructed based on the IPNs with MIMO transitions (see Sections~\ref{sec:sys-formal-pn}), for the human-AI agents collaboration system formalism, as below that combines meta-level (orchestration \& governance), agent-level (coordination, communication \& interfaces), and execution-level (skills \& operations as sub-pages), each agent net contains its internal sub-nets/sub-pages such as \enquote{sense}, \enquote{plan}, \enquote{act,} etc. This design allows system designers to (1) governs global modes and policies, using guards/inhibitors to enable/disable families of agent- and execution-level transitions (e.g., access rights, safety). It reads system diagnostics but does not move work tokens; structural reconfigurations (ownership sets, collaboration links, page wiring) are performed and logged here; (2) owns each agent’s organizational protocol net and public boundary, where single-owner boundary transitions and multi-owner collaboration transitions are the only ones permitted to touch foreign-owned places (humans, other agents, environment), and houses lifecycle logic, capacities/locks, and rate control, with substitution transitions invoking execution-level skills while boundary membership is determined solely by cross-ownership at this level; (3) implements functional skill sub-nets that transform tokens and update interpreted state: (i) internal-cognition (Plan, Reflect, Learn, Verify) operating only on agent-owned places; (ii) social-IO (Send/Receive, Request/Reply, Teach/Learn-from, Commit/Negotiate) via the agent’s own interface places and paired with agent-level boundary/collaboration transitions for crossing; and (iii) environment-IO (Actuate, Perceive), and execution subpages never wire directly to foreign-owned places. \par

It is rational to use multiple layers of Petri nets (or hierarchical net constructs) so that each layer addresses a different scope of control and configuration. We can conceptualize the system as three nested (or stacked) layers briefly:\par
\textbf{Meta-Level (Orchestration \& Governance).} Governs system-wide modes and policies (e.g., explore–exploit bias, global resource allocation, team merge/split). It can reconfigure the organizational Petri net structure by enabling/disabling families of transitions via guards/inhibitors, adding or removing collaboration links, or adjusting ownership sets; such changes are auditable and logged. It reads global diagnostics but does not shuttle \enquote{work} tokens.\par
\textbf{Agent-Level (Coordination, Communication \& Interfaces).} Manages each agent’s organizational protocol net and public boundary. Boundary (single-owner) and collaboration (multi-owner) transitions are the only transitions whose arcs touch foreign-owned places (humans, other agents, environment). This level enforces modes, concurrency and capacity constraints (e.g., \enquote{sense-plan-act} (SPA) and learning concurrency, locks/semaphores, rate limits) and lifecycle logic (discovery $\to$ handshake $\to$ negotiation $\to$ contract $\to$ transfer/acknowledge $\to$ close). Substitution transitions here invoke execution-level skills; boundary membership is determined solely by cross-ownership at this level. The empirical implementations of \enquote{stage-III} HAACS, mainly involving the agent-level activities, are summarized in Appendix Table~\ref{tab:app-stage-3} \par
\textbf{Execution-Level (Skills \& Operations)} Provides the finer-grained sub-nets that implement specialized functions: detailed sensing/actuation, social-I/O with humans/agents (send/receive, request/reply, teach/learn-from, commit/verify), and meta-cognition/learning. Execution subpages transform tokens and update interpreted state but wire only to agent-owned places (e.g., inbox/outbox, actuator buffers); any cross-owner effect occurs only when the agent-level boundary/collaboration transition fires. The functional model of the interplay between the basic components within an agent is shown in Figure~\ref{fig:cyber-loop}.\par

Hence, from top to bottom: Meta-level = global orchestration and reconfiguration; Agent-level = per-agent interfaces, coordination, and concurrency management; Execution-level = skill sub-nets that perform operational work once agent-level transitions enable a subtask. Each layer tackles a distinct concern: the meta-level decides \enquote{who does what, under which global mode,} the agent-level governs per-agent resource usage and interaction protocols, and the execution sub-nets realize step-by-step operations (e.g., sensor data handling). From an organization-theory perspective, the meta-level can \enquote{load} and \enquote{unload} agent-level modules or flip policy switches, while agent sub-nets proceed with day-to-day concurrency.
\subsubsection{Concurrent execution of multiple sub-nets}
Note that the multiple sub-nets can run concurrently. While a hierarchical Petri net often models a sub-net as a \enquote{module} that runs to completion before control returns to the higher-level net, we can instantiate multiple such sub-nets in parallel. In practice, this is achieved by allowing multiple tokens to invoke the sub-net independently, so each instance processes its task concurrently, and then later re-joins the main net when it completes (or reaches a designated intermediate state).\par
For example, an agent might have a \enquote{sense-plan-act} (SPA) module that, upon receiving a token, launches an instance of the Act sub-net. If new tasks arrive before the previous instance finishes, additional tokens can concurrently trigger further instances of that sub-net. This is analogous to multi-threading in software where several tasks run in parallel and later merge their results back into the main process. In short, hierarchical Petri nets can be designed to allow concurrent execution of multiple sub-net instances, as long as the design permits multiple tokens (or \enquote{threads}) to be active simultaneously.
\subsubsection{Rationale for hierarchical design}
Hierarchical structures and multilevel interactions are widespread across stages of human–AI collaboration, especially in the open complex system we address, which are discussed in Section~\ref{sec:cyber-agenthood}. Accordingly, these structures underpin our hierarchical design of HAACS. Regarding scope separation, each layer tackles different concerns. The meta-level deals with overall \enquote{who does what, in which global mode,} the agent-level deals with concurrency and resource usage per agent, and the execution sub-nets implement the final details (like step-by-step sensor data handling).\par

If everything were jammed into a single Petri net, we'd quickly face an unwieldy global net with thousands of places/transitions. By layering, each net stays simpler and focuses on its domain. Moreover, the meta-level can \enquote{load} or \enquote{unload} certain agent-level modules or shift them into new modes, akin to flipping big \enquote{organizational} or \enquote{policy} transitions. Meanwhile, the agent sub-nets proceed with day-to-day concurrency as usual.\par

We might imagine an even higher \enquote{universe-level} net that decides whether the meta-level net is active or not, etc. If the system is extremely complex (like multiple organizations of agents), a 4th layer that orchestrates multiple meta-level modules could exist. But typically, the design that meta-level for reconfiguration, agent-level concurrency, and specialized sub-module for detailed operations is enough.
\subsubsection{Two feedback loops in the design aligned with second order cybernetics}
In the three-layer Petri net architecture, we have: (1) \textbf{Execution-Level Feedback Loop}: each agent at least has a \enquote{Sense-Plan-Act} (SPA) and other local feedback loops across execution-level sub-nets controlled by the agent-level; SPA composes the basic \enquote{cybernetic} loop, e.g. the agent senses the environment, processes the data, makes decisions and acts, then sees the result and adapts further (see Section~\ref{sec:cyber-agenthood}). (2) \textbf{Meta-Level Feedback Loop}: the meta-level Petri net monitors the agent-level behavior (e.g., performance metrics, resource usage, \enquote{how well the concurrency is working,} or \enquote{are agents exploring vs. exploiting effectively?}); it can reconfigure or override the agent-level sub-nets such as merging agents, changing concurrency tokens, toggling policy modes, etc.\par
The meta-level loop is explicitly controlling the agent-level Petri net, which in turn controls the execution-level sub-nets, mirroring a self-referential or higher-level control-of-control loop in terms of second-order cybernetics. Specifically, this architecture aligns with second-order cybernetics as follows: (1) Meta-Level Observes and Adapts: the meta-level net doesn't merely watch passively; it acts to reconfigure the agent-level net's concurrency or resource tokens, and even merges/splits entire agent sub-nets. This is reminiscent of the \enquote{observer in the system} viewpoint: not only does the higher-level net observe the whole system, it is also an integral part of it, modifying the rest of the system's control loops. (2) Self-Reflection: if the system is designed to allow the agent-level net to send \enquote{performance tokens} or \enquote{resources utilization} up to the meta-level net, it effectively has a self-reflective loop, i.e., the system is generating data about how well it's controlling itself, feeding that data to a higher-level loop that rewrites/improves the lower-level loop in real time. (3) Evolving Boundaries: the second-order cybernetics often discusses how the system boundary can shift or how the observer can become part of the system. In our multi-layer approach, if the meta-level net decides to merge or split agent sub-nets (e.g., assign a long-term memory module to an agent, which belongs to another agent originally; combine two agents into one, etc), that changes the boundary among the components of agents and the identifications of the agents. This is very much in the second-order flavor of \emph{the control system changes its own boundaries dynamically.}
\subsubsection{Generalization and compatibility with extant HAACS protocols and interoperability}\label{sec:pn-fit-extant-work}
Extant work on protocols and frameworks for HAACS includes Agent2Agent (A2A)~\citep{a2aProtocol2025}, which standardizes capability discovery and task delegation between opaque agents via HTTP/JSON-RPC with async streaming; the Agent Communication Protocol (ACP)~\citep{ibmBeeAIacp2024}, a REST-native, multipart messaging layer with registries and observability for typed, multi-modal exchanges; the Agent Network Protocol (ANP)~\citep{anpContributors2024}, which enables open-internet agent discovery and collaboration using decentralized identifiers (DIDs) and JSON-LD (i.e., JSON for Linked Data, a JSON format for linked data with machine-readable semantics); Agora~\citep{marro2024scalable}, a meta-protocol that negotiates protocol documents and blends structured routines for frequent traffic with natural-language fallbacks; LOKA~\citep{ranjan2025loka}, a decentralized identity and ethics–aware orchestration stack (e.g., DIDs, verifiable credentials) for trustworthy multi-agent operation; and PXP~\citep{srinivasan2024implementation}, a human–agent interaction protocol that enforces intelligible, two-way \enquote{predict-and-explain} dialogues (RATIFY/REFUTE/REVISE/REJECT), including MCP~\citep{modelcontextprotocol} for standardized context ingestion and tool invocation. Collectively these span enterprise, open-network, governance, and human-in-the-loop settings~\citep{yang2025survey,ehtesham2025survey}, precisely the interoperability surface our model targets; the following evaluation proves generalization and compatibility of our three-layered hierarchical Petri nets formalism.\par

To make these correspondences explicit, Table~\ref{tab:gen-compat-layer-map-p1}-\ref{tab:gen-compat-layer-map-p2} places representative implementations/protocols/frameworks within our three-layered hierarchical IPN (see Section~\ref{sec:three-layer-haacs}), with meta-level (orchestration \& governance), agent-level (coordination, communication \& interfaces), and execution-level (skills \& operations). Following our rules (see Sections~\ref{sec:basic-pn-model} and \ref{sec:mimo-pn}) that \enquote{boundary (single-owner) and collaboration (multi-owner) transitions are the only ones permitted to touch foreign-owned places,} entries are populated only where these reviewed work specifies behavior at that layer; blanks indicate no forced mapping. In particular, MCP \enquote{replaces brittle, ad-hoc prompt engineering with a persistent JSON-RPC protocol for secure, stateful, and auditable tool invocation}~\citep{modelcontextprotocol}, aligning with execution-level guarded updates $X' \leftarrow f_t(X,\gamma)$ gated by agent-level guards; A2A and ACP capture agent-level contracts and multi-modal \enquote{ordered parts} messaging; ANP’s DID/JSON-LD identity and capability discovery gate collaboration membership and $\mathrm{Owners}(t)$; Agora’s Protocol Documents and LLM-written routines compile into agent-level protocol sub-nets while the meta-level can \enquote{add or remove collaboration links}~\citep{marro2024scalable}; PXP resides at the human–agent boundary~\citep{srinivasan2024implementation}; and LOKA supplies meta-level policies that enable/disable families of boundary/collaboration transitions~\citep{ranjan2025loka}. In short, Table~\ref{tab:gen-compat-layer-map-p1}-\ref{tab:gen-compat-layer-map-p2} demonstrate interoperability and compatibility, i.e., the reviewed protocol families fit the three layers without ad hoc add-ons outside our IPN formalism, and provide evidence of generalization and extensibility, i.e., the meta-level unifies identity, policy, and rate governance across heterogeneous stacks.

\begin{table}[htbp]
\centering
\small
\setlength{\tabcolsep}{2.8pt}
\begin{adjustbox}{scale=0.88,center}
\begin{tabularx}{1.3\textwidth}{@{}
    p{0.24\textwidth}   
    p{0.34\textwidth}   
    p{0.34\textwidth}   
    p{0.34\textwidth}   
@{}}
\toprule
\textbf{Work} & \textbf{Meta-level} & \textbf{Agent-level} & \textbf{Execution-level} \\
\toprule
\textbf{MCP}~\citep{modelcontextprotocol} &
Global policies can enable/disable families of MCP boundary columns $C^{*}_{\mathrm{MCP}}$ and route through human-owned places $P_H$ and human transitions $T_H$ for consent tokens. &
\emph{Tools} and \emph{Resources} $\leftrightarrow$ execution-level skill transitions \& resource places; \texttt{tools/call} $=$ boundary transition owned by the invoking agent (or collaboration transition if crossing trust domains). Capability discovery/transport-agnostic comms realized as agent-level guards that enable families of boundary columns. &
\enquote{Decouple think/do}: guarded firings with stateful updates $X' \leftarrow f_t(X,\gamma)$; schema checks as $G_t$. \\
\midrule
\textbf{Gorilla}~\citep{patil2024gorilla} &
N/A &
Tool-call transitions are permitted/blocked by agent-level policy, with back-pressure and retries as rate/lock controls. &
Retriever features $+$ schema/AST matches act as execution-level guards/validators on tool-call transitions; failures route to \emph{Verify}/\emph{Reflect} sub-nets; success commits via $X' \leftarrow f_t(X,\gamma)$. \\
\midrule
\textbf{CoA}~\citep{DBLP:conf/coling/GaoDYTPGSCBW25} &
Meta-level reconfiguration and per-step guards $G_t(X,\gamma)$ mitigate the method’s “static plan” limitation by enabling mid-course replanning and selective inhibition. &
N/A (plan hand-off is an agent-internal substitution-transition enabling). &
Abstract CoT $=$ enabled but \textbf{unbound} substitution transitions; specialized tools \emph{reify} placeholders as execution-level firings $t$ with updates $f_t(X,\gamma)$; colors $\gamma$ select the concrete tool. \\
\midrule
\textbf{HuggingGPT}~\citep{shen2023hugginggpt} &
N/A &
Controller decomposes requests into JSON task list with explicit \texttt{task/id/dep/args}; dependency edges correspond to inhibitor/guard relations that gate when boundary enables skills. &
Isomorphic to multiple tokens traversing execution sub-nets in parallel (multi-instance firing of SPA/skill pages); placeholders produce colored tokens bound at firing time. \\
\bottomrule
\end{tabularx}
\end{adjustbox}
\caption{HAACS tool-use implementations and protocol methods mapped to the three-layer IPNs (execution-centric).}
\label{tab:gen-compat-layer-map-p1}
\end{table}

\begin{table}[htbp]
\centering
\small
\setlength{\tabcolsep}{2.8pt}
\begin{adjustbox}{scale=0.88,center}
\begin{tabularx}{1.3\textwidth}{@{}
    p{0.24\textwidth}   
    p{0.34\textwidth}   
    p{0.34\textwidth}   
    p{0.34\textwidth}   
@{}}
\toprule
\textbf{Work} & \textbf{Meta-level} & \textbf{Agent-level} & \textbf{Execution-level} \\
\toprule
\textbf{A2A}~\citep{a2aProtocol2025} &
N/A &
Typed outsourcing via \enquote{Agent Cards}, \enquote{Tasks}, JSON-RPC and SSE; aligns with single-owner boundary transitions and collaboration transitions whose arcs are the only ones permitted to touch foreign-owned places. &
Skills invoked via substitution transitions; results returned to agent-owned inbox/outbox places. \\
\midrule
\textbf{ACP}~\citep{ibmBeeAIacp2024} &
N/A &
Registry-centric, multi-modal \enquote{ordered parts} messaging realized as boundary transitions with color/guard checks for MIME-typed artifacts; back-pressure via locks/semaphores at this level. &
N/A \\
\midrule
\textbf{ANP}~\citep{anpContributors2024} &
DID/JSON-LD identity \& capability discovery supply \textbf{meta-level} predicates that gate collaboration membership and ownership sets $\mathrm{Owners}(t)$. &
DID verification and JSON-LD capability alignment become agent-level guard predicates that enable/inhibit boundary/collaboration transitions. &
N/A \\
\midrule
\textbf{Agora}~\citep{marro2024scalable} &
Meta-level can \enquote{add or remove collaboration links} and adjust $\mathrm{Owners}(t)$, turning static connectivity into \textbf{reconfigurable} collaboration topology with auditable logs. &
Protocol Documents (PDs) and LLM-written routines compile into agent-level protocol sub-nets (substitution transitions): frequent paths run via structured routines; rare cases fall back to natural language. &
N/A \\
\midrule
\textbf{PXP}~\citep{srinivasan2024implementation} &
N/A &
Occupies the \textbf{human–agent boundary} with explicit finite-state tags carried as token colors; tags verified by human transitions $T_H$. &
Execution sub-nets emit/consume intelligibility tags as part of social-I/O. \\
\midrule
\textbf{LOKA}~\citep{ranjan2025loka} &
Supplies \textbf{meta-level policies} (identity/governance/ethics) that enable/disable families of boundary/collaboration transitions across agents. &
N/A &
N/A \\
\bottomrule
\end{tabularx}
\end{adjustbox}
\caption{Inter-agent interoperability stacks mapped to the three-layer IPNs (coordination/governance-centric).}
\label{tab:gen-compat-layer-map-p2}
\end{table}

\subsection{Achieving cognition and meta-cognition via sub-nets coordination} \label{sec:metacog-pn}
The cognition and meta-cognition capabilities (see Section~\ref{sec:cog-metacog}) can be achieved through modelling the necessary transitions and places for meta-cognition and basic inference operations and coordinating the sub-nets (mainly agent-level and execution-level).
\subsubsection{Petri net modeling of the Sense-Plan-Act loops}
According to Section~\ref{sec:cyber-agenthood}, SPA loops compose a basic goal-seeking adaptive agent, with additional advanced functionality modules that can be plugged in, such as learning, self-explanation, criticism, reflection, and self-reflection. The Petri net representation for SPA loops allows concurrency of \enquote{sense,} \enquote{plan,} \enquote{act,} and additional tasks (like learning or reflection) within one agent.\par
First the SPA loop is decomposed into smaller major sub-tasks, such as: (1) \textbf{Sense}: acquire data from environment; (2) \textbf{Plan}: interpret data, identify and orchestrate tasks, make decisions; (3) \textbf{Act}: execute the selected decisions by interacting with inner modules, other agents, human and environment; and the optional (4) \textbf{Learn}: update long-term memories such as internal parameters of model, etc. Then each sub-task is modeled as a sub-net, and the basic diagrammatic description of places ($P$) and transitions ($T$) for these sub-nets is as follows (note: this description is extendable and provided solely for demonstration purposes):\par
\textbf{Sense Sub-Net}: \texttt{SensingReady}\textemdash a place with a token when the agent can initiate sensing; \texttt{DoSense}\textemdash a transition that fires to collect data; \texttt{NewDataAvailable}\textemdash a place that gets a token once sensing completes.\par
\textbf{Plan Sub-Net}: \texttt{StartPlanning}\textemdash a transition that consumes token from \texttt{NewDataAvailable} once data (including tasks) is ready; \texttt{PlanningInProgress}\textemdash a place that holds the \enquote{work in progress} state; \texttt{FinishPlanning} \textemdash a transition that produces the planning output; \texttt{ActionReady}\textemdash a place with a token that indicates an action can now be triggered.\par
\textbf{Act Sub-Net}: \texttt{InitiateAction}\textemdash a transition that consumes token from \texttt{ActionReady}; \texttt{ActingInProgress}\textemdash a place that represents the system carrying out the action; \texttt{CompleteAction}\textemdash a transition that returns token to \texttt{SensingReady} (closing the loop), or to some environment or human places to model state changes for environment and human explicitly.\par
\textbf{Learn Sub-Net}: this sub-net could be plugged in the SPA loops to run in parallel, and might be triggered by a token from \texttt{NewDataAvailable} or \texttt{ActionReady} (i.e., learning as an internal action planned in \citep{sumers2024cognitive}) or both; \texttt{LearningReady}\textemdash a place that receives tokens indicating new info is available to learn from; \texttt{PerformLearning}\textemdash a transition as a learning step that updates long-term memory of agent such as the internal parameters of LLM model; \texttt{LearningDone}\textemdash a place to hold the completion marker and possibly it sends signals back to \texttt{SensingReady} or \texttt{PlanningInProgress} to inform the agent's updated model.\par
A simple sense $\to$ plan $\to$ act loop can be extended into a more sophisticated net by explicitly representing each phase as sub-nets with places and transitions. Because Petri nets can hold multiple tokens in different places at once, the agent can \emph{sense} again or \emph{learn} while it is still \emph{acting,} provided we allow appropriate transitions and do not enforce strict resource conflicts that block concurrency, which is something basic finite-state machines or simple cyclic loops in \citep{sumers2024cognitive} cannot do as neatly. For more refined models (involving real-time constraints or continuous variables), timed or hybrid Petri nets can be used.
\section{Toward a collaborative problem solving and knowledge management framework}
\subsection{Epistemic lineages: rationalism $\to$ constructivism $\to$ paradigm shift \& dialectics}
From Descartes' rationalist doubt to Piaget's constructivism and Kuhn's paradigm shifts, each philosophical stance highlights that what we take as \enquote{knowledge} is in constant evolution, shaped both by our innate cognitive structures and by our engagement with ever-changing phenomena. To clarify how these historical standpoints differ not merely in emphasis but in mechanism, we juxtapose their core dimensions in Table~\ref{tab:epistemic-triad}.
\begin{table}[htbp]
\centering
\small
\begin{adjustbox}{scale=0.88,center}
\begin{tabularx}{1.2\textwidth}{@{}p{0.18\textwidth} 
p{0.3\textwidth} 
p{0.3\textwidth} 
p{0.3\textwidth}
@{}}
\toprule
\textbf{Dimension} & \textbf{Rationalism (Descartes $\to$ Kant)} & \textbf{Constructivism (Piaget)} & \textbf{Paradigm Shift (Kuhn) \& Dialectics (Hegel)} \\
\toprule
\textbf{Core question} &
How can we ground knowledge in reason and \enquote{separate reliable insights from untested beliefs?} &
How do cognitive structures and the environment \emph{mutually} shape knowing? &
How does knowledge reorganize when contradictions and anomalies accumulate?  \\
\textbf{Key claim} &
Methodical doubt $\to$ \textbf{Cogito, ergo sum}; four rules: \enquote{accept only clear and distinct ideas (systematic doubt), break down complex problems (reductionism and analysis), proceed from simple to complex (systematic synthesis), ensure completeness in reasoning (comprehensive review).}~\citep{descartes1987discours} &
Knowing is governed by \textbf{equilibration and self-regulation}; the subject \enquote{possesses an intrinsic mechanism for self-adjustment.}~\citep{piaget1970genetic} &
\enquote{\textbf{No absolute or final knowledge}}; truth is \textbf{relative to the prevailing paradigm} (Kuhn)~\citep{kuhn1997structure}. Contradictions propel \textit{Aufhebung} that preserves $+$ negates toward a higher unity (Hegel)~\citep{Forster1993HegelsDM}. \\
\textbf{Categories} or \textbf{structure} &
\textbf{Synthetic a priori}; \enquote{12 pairs of categories} organize experience; \enquote{we can never reach the \enquote{thing-in-itself'},} yet categories \enquote{apply universally and necessarily.}~\citep{kant2024critique} &
Structures are \textbf{not pre-formed}; they \textbf{shape} understanding and are \textbf{reshaped} by interaction with the world.~\citep{piaget1970genetic} &
Old and new frameworks may be \textbf{incommensurable} (Kuhn)~\citep{kuhn1997structure}, yet synthesis \textbf{reframes} and \textbf{retains} valid moments (Hegel)~\citep{Forster1993HegelsDM}. \\
\textbf{Mechanism of progress} &
\textbf{Reductionism $\to$ synthesis $\to$ review}; reason provides a \textbf{systematic staircase} for inquiry.~\citep{descartes1987discours} &
\textbf{Selectionist, feedback-driven} equilibration; self-regulating cycles.~\citep{piaget1970genetic} &
\textbf{Crisis $\to$ leap} to a \enquote{new organizing principle} (Kuhn) / \textbf{the negative of the negative} to a richer unity (Hegel).~\citep{kuhn1997structure,Forster1993HegelsDM} \\
\textbf{Treatment of contradiction} &
Minimize by analysis and reconstruction.~\citep{descartes1987discours} &
Use \textbf{self-regulation} to resolve mismatch between scheme and world.~\citep{piaget1970genetic} &
Make contradiction the \textbf{engine of development} (antinomies $\to$ sublation; paradigm shift).~\citep{kuhn1997structure,Forster1993HegelsDM}  \\
\textbf{Relation to practice} or \textbf{fields} &
Disciplines evolve \enquote{for pragmatic reasons}; \textbf{Max Planck}: separations are \textbf{convenient constructs}, not nature's intrinsic joints.~\citep{tsien1986OnTheScienceOfThinking} &
Practice and world interaction \textbf{co-determines} structure; ongoing adjustment.~\citep{piaget1970genetic} &
Practice or anomaly \textbf{force reorganizations}; parts of the old are \textbf{preserved/reframed} in the new.~\citep{kuhn1997structure,Forster1993HegelsDM} \\
\bottomrule
\end{tabularx}
\end{adjustbox}
\caption{Rationalism $\to$ Constructivism $\to$ Paradigm Shift \& Dialectics}
\label{tab:epistemic-triad}
\end{table}
\subsection{Correspondence and coherence in knowledge representation}
Knowledge can be approached in two broad ways: as a direct mirror of external reality or as a self-sustaining network of interrelated concepts. The first anchors physical symbol systems (symbols, structures, processes, designation, interpretation) but faces the symbol grounding and frame problems; the second evolves entailment meshes $\to$ directed entailment nets under the bootstrapping axiom, integrates minimal ontologies, and circulates Tacit $\leftrightarrow$ Explicit via \enquote{SECI} and \enquote{Ba} in consensual domains with \enquote{teach-back}. Accordingly, Table~\ref{tab:corresp-vs-coher} contrasts correspondence with coherence across the key dimensions that matter for knowledge engineering.

\begin{table}[htbp]
\centering
\small
\begin{adjustbox}{scale=0.9,center}
\begin{tabularx}{1.2\textwidth}{@{}p{0.18\textwidth} 
p{0.32\textwidth} 
p{0.41\textwidth} 
p{0.21\textwidth}
@{}
}
\toprule
\textbf{Dimension} & \textbf{\makecell[l]{Correspondence\\(physical symbol systems)}} & \textbf{\makecell[l]{Coherence\\(constructivist or self-organizing)}} & \textbf{References} \\
\toprule
\textbf{Knowledge stance} &
\enquote{Knowledge functions as a \textbf{direct mirror of the external world} and symbols linked to physical entities; a homomorphic representation of reality.} & \enquote{Coherence is a \textbf{two-way relation} in which concepts \textbf{mutually reinforce} one another and structure emerges \textbf{self-referentially}.} &
\citep{heylighen2001bootstrapping} \\
\textbf{Basic knowledge representation (KR)} &
\textbf{Symbols, symbol structures, processes, designation, interpretation}; \textbf{Physical Symbol System Hypothesis}.~\citep{newell1972infops,newell1997computer} &
\textbf{Entailment meshes $\to$ directed entailment nets} under the \textbf{bootstrapping axiom}; \textbf{minimal ontologies} (Class/Object/Property by stability $\times$ generality).~\citep{heylighen2001bootstrapping, pask1976conversation} &
\citep{newell1972infops,newell1997computer,heylighen2001bootstrapping, pask1976conversation} \\
\textbf{Meaning} or \textbf{grounding} &
\textbf{Designation} ties symbols to referents; but faces \textbf{symbol grounding} and \textbf{frame problem}. &
\textbf{Bootstrapping}: meaning arises via \textbf{mutual support}; \enquote{a few governing rules} yield higher-order structure without external primitives.~\citep{heylighen2001bootstrapping} &
\citep{heylighen2001bootstrapping} \\
\textbf{Core problems} &
\textbf{Symbol grounding problem}; \textbf{frame problem}; rigidity and manual primitives. &
Risk of \textbf{coherence drift} if mutual support lacks epistemic checks; must go \textbf{beyond mere consistency}. & \citep{heylighen2001bootstrapping} \\
\textbf{Update dynamics} &
New symbols/rules typically \textbf{manually encoded}; subgoal stacks in classic systems. &
\textbf{Associative learning}: direct/transitive/symmetric reinforcement of links; \textbf{merge/split} for ambiguity resolution; \textbf{self-organization}. &
\citep{heylighen2001bootstrapping} \\
\textbf{Inference} or \textbf{expressivity} &
Strong for \textbf{rule-based deduction}, production systems. &
Supports \textbf{causal/transitive chains} when directed; captures \textbf{mutual support} and \textbf{analogy} via nets/meshes. &
\citep{heylighen2001bootstrapping,pask1976conversation} \\
\textbf{Tacit $\leftrightarrow$ explicit} &
Emphasizes \textbf{explicit} symbol structures. &
Integrates \textbf{tacit$\to$explicit circulation} via \textbf{SECI}; \enquote{we always know more than we can explicitly state.} &
\citep{polanyi2009tacit,nonaka2000seci} \\
\textbf{Validation} &
External \textbf{reference check} against the world (when available); but \textbf{no built-in} self-verification of mapping. &
\textbf{Coherence $+$ practice}: consensual domains, \textbf{teach-back}, citations/RAG~\citep{gupta2024comprehensive}/CoK~\citep{CoK24Li} as \textbf{epistemic checks}. & \citep{heylighen2001bootstrapping,maturana1990biological,boyd2001reflections,CoK24Li}
\\
\textbf{Scalability} or \textbf{ambiguity} &
\textbf{Typed edges} may multiply; \textbf{frame updates} brittle. &
\textbf{Minimal ontology}; ambiguity handled by \textbf{I/O-set identity} and \textbf{merge/split}; weighted links focus traversal. &
\citep{heylighen2001bootstrapping} \\
\textbf{Field boundaries} and \textbf{unity (Max Planck)} &
Tends to \textbf{compartmentalize} by referents/domains. &
Naturally \textbf{cross-links} across domains; aligns with Planck's point that field divisions are \textbf{pragmatic} not ontological. &
\textbf{Planck}~\citep{tsien1986OnTheScienceOfThinking} \\
\bottomrule
\end{tabularx}
\end{adjustbox}
\caption{Contrasting Correspondence and Coherence in Epistemic Frameworks}
\label{tab:corresp-vs-coher}
\end{table}
\subsection{Reasoning modalities linking knowledge and practice}
Building on the trajectories in Table~\ref{tab:epistemic-triad} and the representational polarity in Table~\ref{tab:corresp-vs-coher}, we articulate the reasoning modalities that link knowledge and practice. With the two representations in place, knowledge management and practical problem solving continually amplify each other. \emph{Reasoning} acts as the connective tissue in this cycle, guiding the transition from \enquote{having data} to \enquote{knowing}. Here we enumerate the reasoning palette that shapes and enriches this interplay: formal reasoning as a compressive epistemic framework; dialectical reasoning as a self-developing logic driven by contradiction and \enquote{Aufhebung}; conversational reasoning or Socratic elicitation that induces aporia and enables \enquote{teach-back}; and narrative reasoning (e.g., progressive disclosure, branching, analogy/metaphor) for organizing complexity, setting up the operators that will later drive the problem solving methods such as mean-end analysis (MEA), \emph{explore–exploit principles}, and practice-grounded knowledge consolidation.
\begin{itemize}
\item \textbf{Formal reasoning as a compressive epistemic framework.} \enquote{A distillation and refinement of the epistemological achievements, and a framework and instrument for comprehending new phenomena,} enabling analytical modeling, predictive reasoning, and compressing an \enquote{almost boundless posterior epistemic workload} into tractable scope.~\citep{tsien1986OnTheScienceOfThinking}
\item \textbf{Dialectical reasoning as self-developing logic driven by contradiction.}
Not static rules but a sequence of contradictions and resolutions; each concept contains its negation; progress via sublation (\enquote{Aufhebung}), i.e., \enquote{the negative of the negative} that preserves valid moments while transforming them; interwoven with practice.~\citep{Forster1993HegelsDM}
\item \textbf{Conversational reasoning / Socratic elicitation.}
Maieutic method: structured questioning, refutation, and critical reflection; induces aporia to expose gaps and trigger reconstruction; assumes internal realization, i.e., \enquote{truth already resides within the mind.} Works peer-to-peer in Conversation Theory environments (task/explanatory/meta; teach-back).~\citep{paul2019thinker,heylighen2001bootstrapping,boyd2001reflections}
\item \textbf{Narrative reasoning for organizing complexity.}
Techniques like progressive disclosure, branching, analogy, metaphor as \enquote{methodical syntheses} of how humans organize complex information; in LLM workflows, these appear in Buffer of Thoughts / Narrative prompts to structure problem solving.~\citep{javadi2024can,yang2024buffer}
\end{itemize}
\subsection{Human--AI agents collaborative learning} \label{sec:col-hai-learn}
\subsubsection{Pask's Conversation Theory as constructivist backbone}
Conversation Theory (CT) stands as a hallmark of constructivist thinking, where having knowledge is understood as a process of \enquote{knowing} and \enquote{coming to know}. Learning unfolds through multi-level dialogues (what at the task level, how/why at the explanatory level, and meta-cognitive reflection), anchored in a tangible \enquote{shared environment / micro-world} with teach-back as the alignment test. In consensual domains, human and AI agents learn from and teach each other by languaging, i.e., a \enquote{recursive, multi-layered, interactive} coordination, while entailment meshes and directed entailment nets (with the bootstrapping axiom, I/O-set identity, merge/split, and weighted links) provide the evolving knowledge backbone. \par
Historical and tool-supported counterparts of these elements are summarized in Appendix Table~\ref{tab:ct-evolution-p1} (Key constructs, Dialogue levels, Knowledge representation). As shown in Fig.~\ref{fig:CT-HAI}, we propose a collaborative learning architecture based on CT.
\subsubsection{Tacit–explicit circulation via SECI and Ba}
Alongside multi-level dialogues in a tangible \enquote{shared environment / micro-world}, the knowledge flow explicitly circulates between tacit and explicit via the SECI spiral—Socialization (Tacit $\to$ Tacit), Externalization (Tacit $\to$ Explicit), Combination (Explicit $\to$ Explicit), Internalization (Explicit $\to$ Tacit), within Ba as the shared context for exchange, formalization, and embodiment; knowledge = $\Phi$(Information, Context, Belief) emphasizes justification and context over absolute truth~\citep{nonaka2000seci}. This circulation rests on the tacit dimension, i.e., \enquote{we always know more than we can explicitly state}, with attending-from-to (proximal clues $\to$ distal meaning) guiding discovery and commitment, and then externalization, combination, internalization anchoring organizational learning and assets (experiential, conceptual, systemic, routine) while guarding against core rigidities.~\citep{polanyi2009tacit,tsien1986OnTheScienceOfThinking}
\begin{figure}[htbp]
  \centering
  \includegraphics[width=0.91\linewidth]{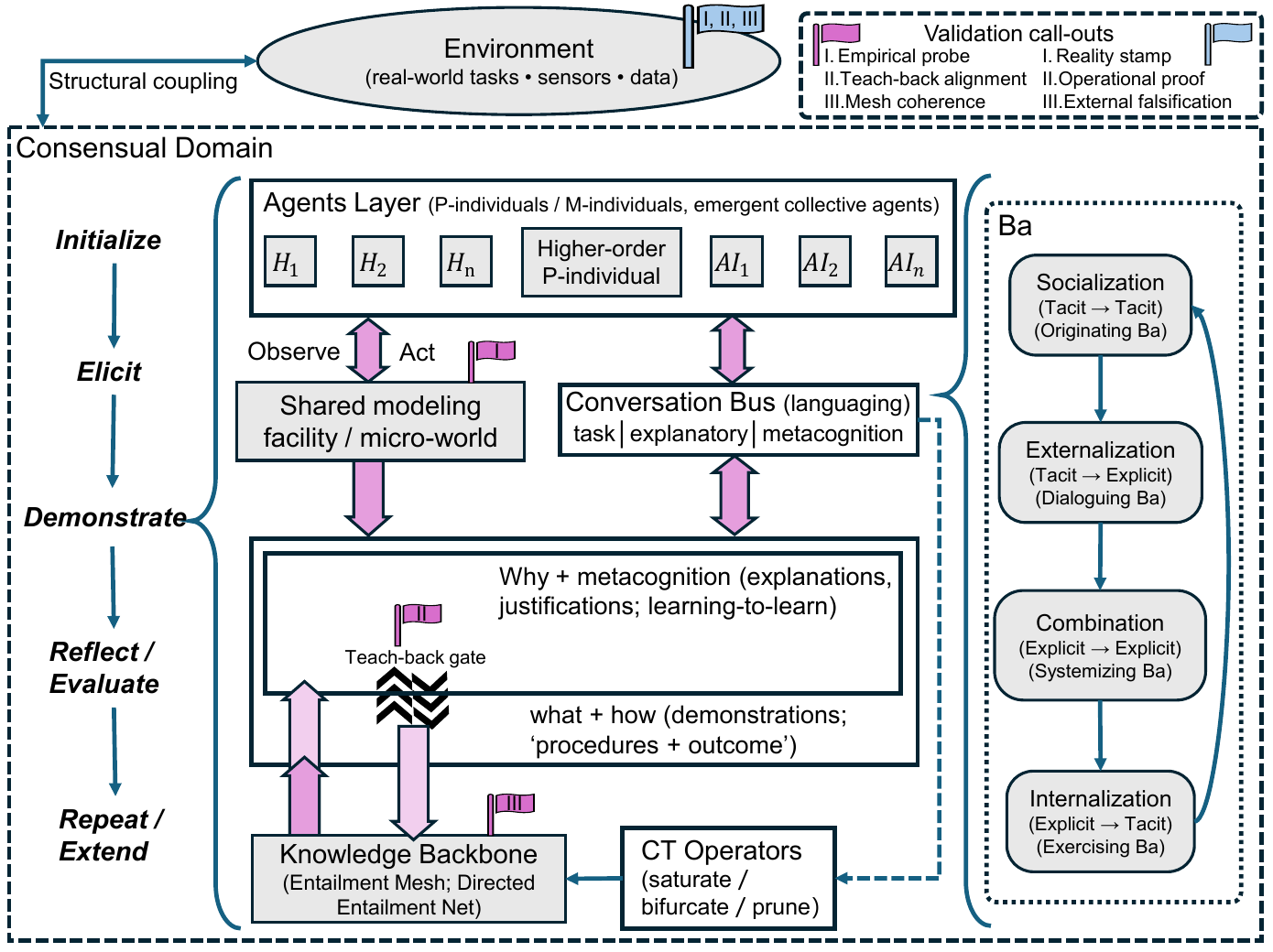} 
  \caption{Human--AI agents collaborative learning architecture. \textit{agents (human \& AI) share a micro-world, converse over a common \enquote{languaging} bus, cycle through Initialize $\to$ Elicit $\to$ Demonstrate $\to$ Reflect/Evaluate $\to$ Repeat/Extend, and grow a self-organizing knowledge backbone while the SECI spiral drives tacit $\leftrightarrow$ explicit conversion.}}
  \label{fig:CT-HAI}
\end{figure}
Figure~\ref{fig:CT-HAI} thus situates the SECI spiral within the consensual domain and shows how languaging, teach-back, and the evolving mesh/net jointly scaffold learning in practice.
\subsubsection{Consensual domains, conversation bus, and validation gates}
Inside the Consensual Domain, a top Agents Layer lists P-individuals and M-individuals plus any higher-order P-individual that crystallizes from stable, shared agreements~\citep{pask1976conversation}. Each agent plugs into a Conversation Bus, and languaging signals the recursive, multi-layered, interactive processes of meaning-making \citep{maturana1990biological} that carry task, explanatory and meta-cognitive traffic without drawing a spaghetti of peer-to-peer lines. The Shared modeling facility / micro-world keeps everyone asking \enquote{Are we talking about the same phenomenon?}~\citep{boyd2001reflections}. From there the dialogue flows into two concentric rectangular regions: an outer \enquote{what $+$ how} layer that stores \enquote{procedures $+$ outcomes}, and an inner \enquote{why + meta-cognition} layer for \enquote{explanations, justifications, and learning-to-learn}~\citep{pask1976conversation}. A double-headed \enquote{teach-back} gate enforces Pask's rule that learning is proven only when the other party can \enquote{reproduce $+$ explain} the target concept, i.e., \enquote{teach-back as ongoing alignment, not just a quiz}~\citep{scott2001gordon}. Operational refinements and validation instrumentation that realize these mechanisms in practice are detailed in Appendix Table~\ref{tab:ct-evolution-p2} (Refinement/learning operators, Shared context or validation, Known limits/notes).
\par
The five-step conversation cycle is the engine of progressive refinement: \enquote{Initialize} establishes a shared goal and selects which experiences are relevant; \enquote{Elicit} draws out each agent's tacit heuristics and assumptions; Demonstrate makes those heuristics observable by enacting procedures in the micro-world; \enquote{Reflect/Evaluate} compares outcomes with rationales to surface gaps; and \enquote{Repeat/Extend} re-injects the corrected insights, launching the next loop~\citep{boyd2001reflections}. Interleaved with that engine, the SECI dynamic performs the tacit–explicit alchemy that turns flashes of experience into organizational capability: Socialization (T$\to$T) lets participants absorb one another's unspoken cues inside an Originating Ba of trust, Externalization (T$\to$E) converts those cues into shared metaphors and models within a Dialoguing Ba, Combination (E$\to$E) systemizes the scattered models into structured artefacts in a Systemizing Ba, and Internalization (E$\to$T) rehearses those artefacts until they harden into embodied skill inside an Exercising Ba, ready to seed the next spiral~\citep{nonaka2000seci}. The Knowledge Backbone contains an Entailment Mesh that captures non-directed clusters and, under the bootstrapping axiom, folds into a Directed Entailment Net where nodes are distinct precisely when their respective input/output sets differ; three weighted-link rules (direct, transitive, symmetric) add associative learning~\citep{heylighen2001bootstrapping}. Feeding and trimming that backbone are the CT operator dials, i.e., saturate, bifurcate, prune, whose arrows remind us that every conversational turn may spawn variants, split concepts or merge redundancies~\citep{pask1976conversation}. Taken together, Appendix Tables~\ref{tab:ct-evolution-p1}–\ref{tab:ct-evolution-p2} map the narrative here onto concrete instantiations: Part~I covers the conceptual scaffolding (constructs, dialogue levels, representation), while Part~II covers the procedural dials and verification gates that keep the system coherent in practice. 
\par
Finally, three red validation call-outs tie the constructivist loop back to reality: empirical probe at the micro-world, teach-back alignment at the gate, and mesh coherence inside the backbone; each red flag sends a bent leader line to the lone blue flag on the Environment ellipse, so that \enquote{operational proof}, \enquote{reality stamp,} or \enquote{external falsification} can veto purely verbal agreement~\citep{heylighen2001bootstrapping,boyd2001reflections}. These checkpoints correspond to the \enquote{Shared context or validation} row in Appendix Table~\ref{tab:ct-evolution-p2}. \par
Together these parts enact the mantra that \enquote{having knowledge is understood as a process of \enquote{knowing} and \enquote{coming to know}}~\citep{pask1976conversation} while letting tacit and explicit knowledge circulate, self-check and adapt in a world that is perpetually changing.
\subsubsection{Mem0 and graph-augmented memory within collaborative learning architecture for knowledge management}\label{sec:mem0-in-collearn}
Persistent, structured memory is central to sustaining coherence across long, multi-session dialogues, yet limited effective context windows force LLM agents to forget once history overflows\citep{byerly2024effective,yang2024buffer,CoK24Li,mitra2024agentinstruct}. Mem0~\citep{chhikara2025mem0} addresses this by extracting salient facts, reconciling them with an evolving store via LLM-routed CRUD operations (\enquote{ADD}, \enquote{UPDATE}, \enquote{DELETE}, \enquote{NOOP}), and retrieving only the minimal set needed at answer time, yielding both accuracy gains and substantial latency and token-cost reductions; Mem0g~\citep{chhikara2025mem0} extends this foundation with a graph-based memory where entities and relations are stored as a directed, labeled graph, enabling temporal and relational reasoning with \emph{conflict-aware} updates and dual retrieval. Mem0 operates fully automatically, its extraction, CRUD decisions, and retrieval are LLM-driven; it does not specify human approval loops, consent mechanisms, or interactive governance. In our collaborative learning setting, they align with knowledge engineering and knowledge cybernetics by plugging into a self-organizing knowledge backbone; but the teach-back gates, human approvals, and entailment meshes/nets are provided by our CT-based scaffolding, not by Mem0.\par

Accordingly, situate Mem0 and Mem0g inside the collaborative learning architecture of Fig.~\ref{fig:CT-HAI} by aligning their pipeline stages with our constructs, indicating their placement and the design guidance they suggest for different domains and scenarios. Mem0’s dense, atomic memories fit the outer \enquote{what + how} store of procedures and outcomes, while Mem0g’s relations populate the inner explanatory layer and the knowledge backbone. When deployed with human collaborators, we need to wrap Mem0’s agent-internal pipeline with explicit human touchpoints (e.g., review of sensitive updates), but these touchpoints are architectural additions rather than features of Mem0.\par
Practically, the LLM-routed CRUD keeps the store compact, retrieval prepends only the most relevant memories, and measured 95th percentile latency and token cost drop sharply relative to full-context baselines, while maintaining or improving judge scores across single-hop and multi-hop questions; these measurements and behaviors are automated operation without human gating \citep{chhikara2025mem0}. Architecturally, Mem0’s extraction produces concise explicit memories from dialogue turns, and its reconciliation (ADD/UPDATE/DELETE/NOOP) consolidates and de-duplicates those memories; we interpret these (in our framework) as aligning with Externalization and Combination in the SECI spiral within Ba. Retrieval supports Internalization by repeated, situated use. Mem0g contributes explicit temporal and relational structure via a directed, labeled graph with soft invalidation of conflicting edges; this graph formalization remains automated in the Mem0g’s setup; we view this as analogous to making an entailment mesh directed and typed, preserving history needed for temporal reasoning (and, in our CT framework, teach-back auditability). In Mem0g specifically, retrieval combines entity-centric walks with semantic triplet ranking, enabling temporal and relational queries without scanning the full conversation context. \par

This placement also reveals gaps we can fill. The Mem0 paper focuses on extraction, CRUD, and retrieval quality and efficiency, but does not operationalize multi-level teach-back gates, consent and scope on memory writes, or second-order control of consolidation and decay policies. However, privacy handling and consent-scoped writes are out-of-scope for Mem0 itself; in our architecture these are realized via human transitions and policy guards around UPDATE/DELETE. Our architecture supplies these missing parts: the gate enforces “reproduce + explain” before admission, the bus carries consent tokens and scope, and the meta-level adjusts consolidation frequency, decay, and conflict-resolution strategies as performance tokens indicate drift or overload. In addition, the backbone provides a principled locus to fuse dense facts with graph relations via I/O-set identity and merge–split operators, preventing coherence drift and making temporal and causal paths auditable. \par

Finally, this mapping guides domain adaptation and shows extensibility. In our architecture, in clinical or safety-critical domains, policy knobs raise thresholds for UPDATE and DELETE, route sensitive changes to human transitions, and prioritize temporal consistency. In education, teach-back gates can require richer explanation paths, while Mem0g’s edges capture curricular pre-requisites and learning trajectories. In customer support, Mem0 alone may suffice for low-latency single-hop retrieval, with Mem0g enabled only for escalation workflows where timeline reasoning matters; this selectivity follows the results where Mem0g does not improve multi-hop over Mem0 but does improve temporal questions. In short, Mem0 and Mem0g fit the conceptual collaborative learning architecture for HAACS, and provide evidence of generalization and extensibility, i.e., this same architecture can guide memory design across heterogeneous domains while preserving the constructivist loop of languaging, teach-back, SECI circulation, and an auditable, evolving knowledge backbone.

\subsection{Human--AI agents collaborative problem solving}
\subsubsection{Problem space framing and canonical GPS-style workflow of MEA} \label{sec:mea-frame-intro}
\begin{figure}[htbp]
  \centering
  \includegraphics[width=0.92\linewidth]{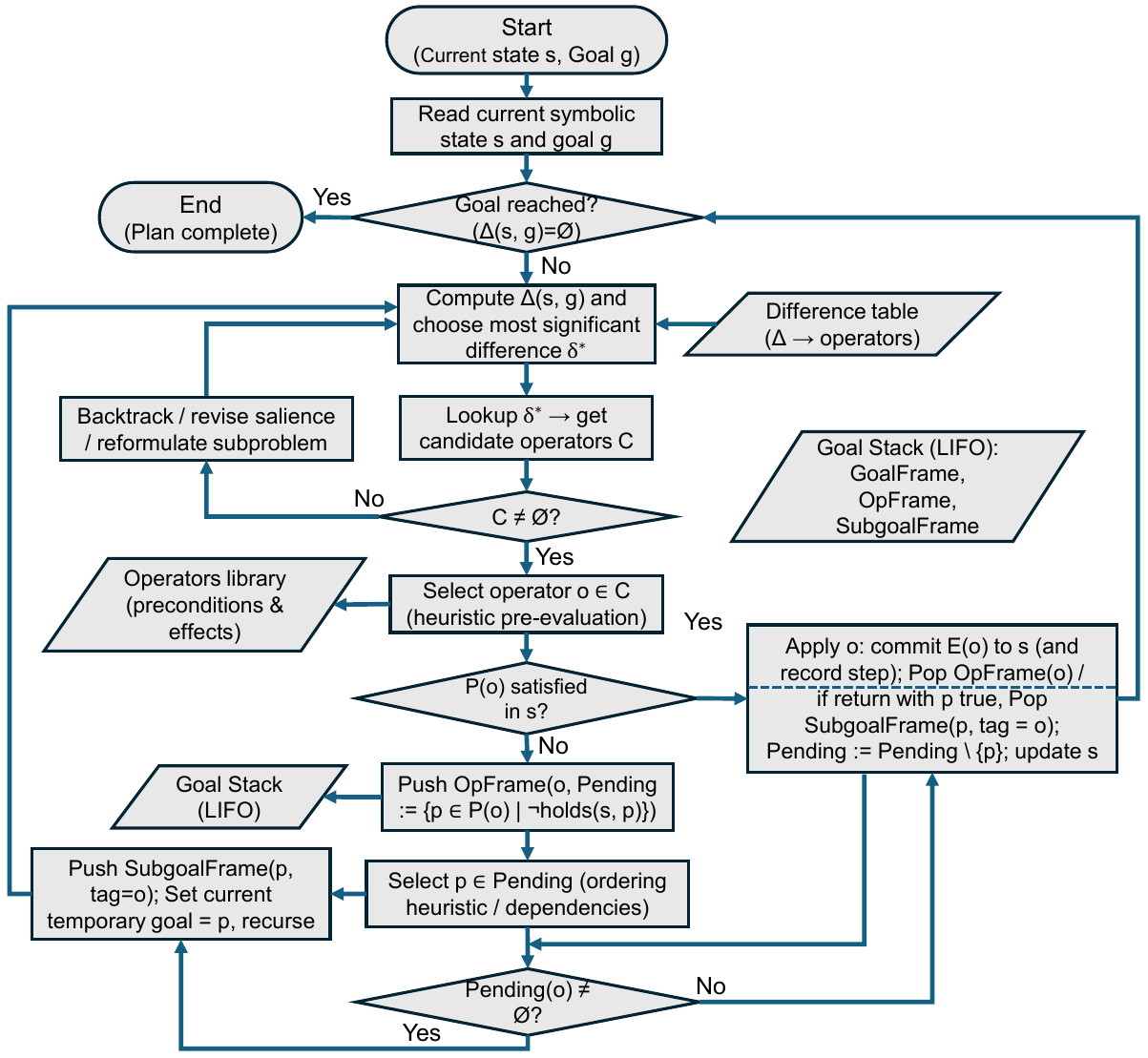} 
  \caption{Original GPS-style MEA control loop. Rectangles denote processes; diamonds denote tests; parallelograms denote knowledge/working stores. The interpreter computes $\Delta(s,g)$, selects a salient difference $\delta^*$, consults the \emph{difference table} ($\delta^* \mapsto$ operators), and chooses an operator $o$. If $P(o)$ already holds, $o$ is applied and its effects $E(o)$ are committed to the symbolic state $S$. Otherwise, an \emph{OpFrame} with $\mathrm{Pending}:=\{p\in P(o)\mid \neg holds(S,p)\}$ is pushed; for each $p\in\mathrm{Pending}$ the system pushes a \emph{SubgoalFrame}(p), recurses to achieve $p$, and on return removes $p$ from \emph{Pending}. When \emph{Pending} becomes empty, the already-selected operator $o$ is applied immediately; the loop terminates when $\Delta(s,g)=\varnothing$.}
  \label{fig:mea-naive}
\end{figure}

Means–ends analysis (MEA) is intended for tasks that can be framed in a problem space, i.e., \enquote{a set of symbolic structures (states) and a set of operators,} with designated start and goal states, so that progress can be made by heuristic search under \enquote{bounded rationality} and essentially serial action~\citep{newell1972logic,newell1972infops}. Within that envelope MEA stays deliberately general: it prescribes no domain ontology, only the control logic of (i) comparing the current state to the goal, (ii) picking an operator that promises to narrow the most salient gap, and (iii) pushing sub-goals when that operator's pre-conditions are unmet. This keeps MEA a portable scaffold for explaining early human or machine problem solving while leaving room for richer representations, stronger heuristics, and learned control knowledge. Figure~\ref{fig:mea-naive} depicts the original GPS-style MEA with difference selection, operator choice, precondition-driven subgoaling, and immediate application once all preconditions hold. We use MEA control \enquote{interpreter} to denote the executive loop that reads the current state and goal, computes differences, consults the difference table, manages the goal stack, and applies operators; it interprets the domain representation rather than guaranteeing optimal solutions. \par
Beyond the canonical GPS loop, established cognitive architectures and recent LLM-based implementations instantiate MEA-like control in complementary ways: Soar adopts MEA principles through universal subgoaling, whenever a conflict arises, it creates a subgoal and later encodes that solution path as a new production (chunk)~\citep{laird2019soar}; ACT-R can implement MEA by encoding goal–state differences in the conditions of production rules, with sub-symbolic activation parameters (e.g., recency, frequency, and past success) serving as heuristics for operator choice~\citep{anderson2014rules}; CoALA, while neutral about which heuristic is used, supports a \enquote{propose–evaluate} planning loop that can be instantiated to implement MEA, treating any shortfall (e.g., lacking facts) as a \enquote{difference} and invoking retrieval or other internal actions accordingly~\citep{sumers2024cognitive}. In LLM-based agents, Tree-of-Thoughts (ToT) parallels MEA by generating \enquote{operators} (candidate sub-rationales) and self-assessing their effectiveness at narrowing the gap, guided by BFS/DFS over the emerging chains, even though it does not explicitly construct formal sub-goals~\citep{yao2023tree}; multi-agent AGENTVERSE~\citep{chen2024agentverse} compares the current state of the environment with the desired outcome to produce prompt-based feedback for the next iteration, echoing means–end reduction without explicit rule-based procedures.

\subsubsection{Intrinsic limitations and motivating extensions} \label{sec:mea-limit}
The naïve GPS implementation nonetheless inherits structural limits. Because it assumes a single, deterministic, closed world-model, sub-goal pursuit can trigger interaction and explosion (one pre-condition undoing another), deadlocks or regressions (local progress enlarges the global difference), and myopic salience (a bottleneck chosen too early blocks cheaper alternatives). Its goal protection is weak, difference tables can be brittle, and the strictly serial, resource-bounded control loop can suffer severe combinatorial explosion on multi-object or multi-objective tasks. Bridging these gaps has driven later advances: partial-order / least-commitment planning with causal-link protection, goal-interaction–aware schedulers, heuristic topology and dead-end detection, chunking-based learning of control knowledge, and hybrid schemes that marry MEA's symbolic supervision with continuous regulators~\citep{kambhampati1997refinement}. Together they extend MEA from an elegant exploratory heuristic into a principled, scalable framework for contemporary human-AI agents problem solving.
\subsubsection{Discovery within the praxis–cognition cycle}
Discovery belongs to ill-structured problem-solving tasks characterized by relatively ill-defined goals. When genuine discovery occurs, something new emerges, an insight or phenomenon not predictable with full certainty, that carries value or interest. This is the working sense of open-ended problem solving adopted for science and design alike. Kolb's cycle shows why such inquiry never strictly ends, i.e., “each new experiment or observation can spark further questions,” so the loop of concrete experience → reflection → abstract conceptualization → active experimentation continues~\citep{kolb2014experiential}. Rescher's two-cycle model makes explicit that inquiry must justify why (correction by coherence—reasoning over assumptions) and improve how (pragmatic / practical correction—reasoning over methods) at the same time; the interplay of theoretical justification and pragmatic feedback creates a self-correcting system, ideal for open-ended pursuits~\citep{Rescher2020}. Grounding both is Piaget's praxis–cognition complementarity, i.e., theory $\leftrightarrow$ practice dialectic: \enquote{Without a mathematical or logical apparatus there is no direct \enquote{reading} of facts, because this apparatus is a prerequisite. Such an apparatus is derived from experience, the abstraction
being taken from the action performed upon the object and not from the object itself.}~\citep{piaget1970genetic}.\par
\subsubsection{A constructivist framework for practice-guided open-ended problem solving} \label{sec:practice-oe-ps}
This contrasts sharply with closed-ended, MEA style tasks, where goals, operators and stopping criteria are preset; once the shortest path is found, reasoning halts. By contrast, practice-guided open-ended problem solving keeps re-framing both the question and the method as results return. Also, tacit knowledge is not optional but constitutive: \enquote{we always know more than we can explicitly state and the attending-from-to phenomenon is the essence of tacit knowing,} and the Meno paradox is resolved by \enquote{a tacit dimension of foreknowledge} that lets researchers sense that \enquote{something is there} and commit to pursuing it before all the clues can be itemized, which anchors the Tacit inlet in the diagram~\citep{polanyi2009tacit}. \par

\begin{figure}[htbp]
  \centering
  \includegraphics[width=0.85\linewidth]{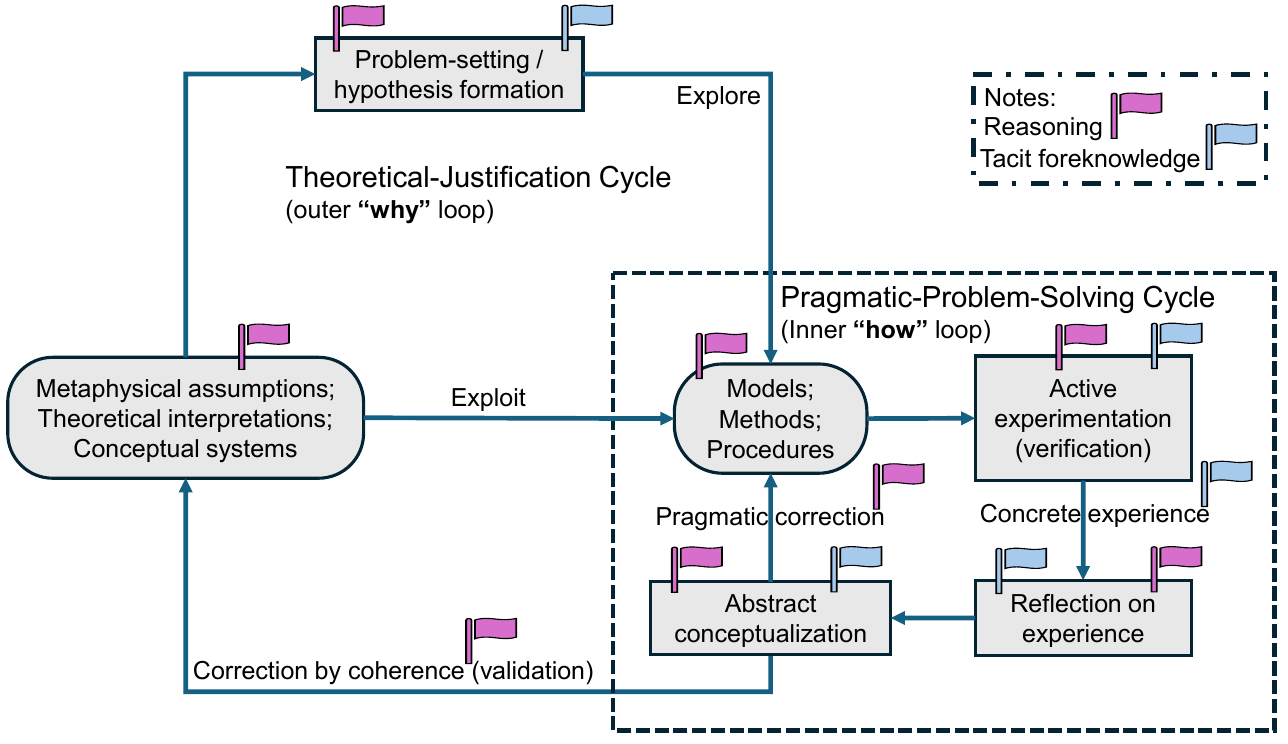} 
  \caption{Practice-guided open-ended problem solving. Inspired by Kolb's experiential cycle (experience $\to$ reflection $\to$ abstraction $\to$ experimentation) and Rescher's twin cycles (correction by coherence for why; practical correction for how), with tacit foreknowledge / attending-from-to (Polanyi, Meno) feeding both; Piaget's \textbf{praxis–cognition} complementarity (theory $\leftrightarrow$ practice dialectic) sustains the theory–practice dynamics that keep discovery alive and distinguish open-ended inquiry from closed-ended MEA tasks}
  \label{fig:open-ended-problem solving}
\end{figure}

To assemble these strands into an operational picture, Figure~\ref{fig:open-ended-problem solving} integrates Kolb’s experiential loop with Rescher’s twin corrections, anchored by Piaget’s praxis–cognition complementarity and Polanyi’s tacit inlet.

\subsubsection{Mapping recent AI-for-science systems to our constructivist framework} \label{sec:ai4sci-work2frame}
Recent agentic systems for scientific discovery span surveys and working pipelines. A crisp survey unifies fully autonomous vs. human–AI collaborative systems and single vs. multi-agent setups across chemistry, biology, materials, and general science, and argues for joint optimization of accuracy–cost–speed–reliability with calibration and human–AI collaboration as near-term unlocks~\citep{gridach2025agentic}. A multi-agent, test-time-compute \enquote{AI co-scientist} generates, debates, ranks, and evolves hypotheses with meta-review and Elo tournaments (competition-driven ratings), and reports wet-lab signals under human-in-the-loop guardrails~\citep{gottweis2025towards}. For social science, LLM-based multi-agent simulations replicate the hypothesize–experiment–analyze empirical cycle entirely in silico with structural causal models and automated regression, enabling parallel \enquote{universes} for reproducibility while noting rigidity and bias limits~\citep{manning2024automated}. \emph{The AI Scientist} automates idea generation, experimental iteration, and reviewing/archiving, mirroring Simon’s decision cycle and the \enquote{explore–exploit} mechanism but remaining isolated from emerging human expertise and data~\citep{lu2024ai,Simon1977,simon1993decision}. OpenAI’s Deep Research iteratively plans, retrieves, and synthesizes with source attribution and reasoning traceability, yet still shows hallucinations and confidence calibration gaps.\par

Accordingly, Table~\ref{tab:sd-tp-1} aligns these studies with our constructivist framework for practice-guided open-ended problem solving, along four dimensions that matter for mechanism design: \emph{theory–practice dynamics}, the \emph{tacit} inlet, \emph{reasoning technique}, and \emph{architecture}. The mapping illustrates that our conceptual scaffold generalizes across heterogeneous agent designs while making explicit where our constructivist framework can add missing governance (teach-back, human transitions, calibration, and consent). Anchoring Fig.~\ref{fig:open-ended-problem solving}, we mark where each system enters or exits our cycles: \emph{why$\rightarrow$how} denotes the \emph{\textbf{explore}} transfer from problem-setting to models/methods, and \emph{how$\rightarrow$why} denotes the \emph{\textbf{exploit}} return from concrete practice back to theory.\par
\begin{table}[htbp]
\centering
\small
\setlength{\tabcolsep}{2.8pt}
\begin{adjustbox}{scale=0.88,center}
\begin{tabularx}{1.4\textwidth}{@{}
    p{0.24\textwidth}   
    p{0.26\textwidth}   
    p{0.24\textwidth}   
    p{0.29\textwidth}   
    p{0.31\textwidth}   
@{}}
\toprule
\textbf{Work} & \textbf{\makecell[l]{Theory–practice}{dynamics}} & \textbf{Tacit knowledge} & \textbf{Reasoning technique} & \textbf{Specification} \\
\toprule
Agentic AI for Scientific Discovery (survey)~\citep{gridach2025agentic} & Stage-wise workflow (ideation $\rightarrow$ design/execute $\rightarrow$ analyze $\rightarrow$ write); pushes calibration and human teaming & Emphasizes human–AI collaboration and trustworthiness (assurance, explainability); highlights literature review as failure point needing human calibration & RAG; multi-agent coordination; composite metrics (accuracy–cost–speed–reliability) & Taxonomy across autonomy/cardinality/domain; tool stacks; evaluation lenses and datasets \\
\textit{AI Co-Scientist}~\citep{gottweis2025towards} & Closed-loop generate $\rightarrow$ debate $\rightarrow$ review $\rightarrow$ rank $\rightarrow$ evolve, with wet-lab probes. & Human-in-the-loop gating; safety guardrails; meta-monitoring; logs; preliminary red-teaming; tacit cues surface in expert reviews and experimental judgment & Self-play debates, Elo tournaments, deep verification (assumption factoring), meta-review; literature-grounded checks & Multi-agent roles (Generation, Reflection, Ranking, Evolution, Meta-review) orchestrated by a Supervisor; Proximity graph; asynchronous scheduling; \emph{prompt-time learning} (no training) \\
LLM Multi-Agent Simulation for Social Science~\citep{manning2024automated} & Replicates hypothesize–experiment–analyze in silico; coordinator-driven turn-taking; emphasizes inner \enquote{how} loop (models $\rightarrow$ scripted experiments $\rightarrow$ automated analysis); limited return to theory \emph{how$\rightarrow$why}. & No human subjects; relies on LLM common sense; tacit cues implicit in pretraining; notes rigidity/bias limits & Role-based dialogues; structural causal models; automated measurement and regression; scripted interactions & Agents act as both \enquote{researchers} and \enquote{subjects}; coordinator role; randomized treatments; buffered dialogue memory \\
\textit{The AI Scientist}~\citep{lu2024ai} & Mirrors Simon’s decision cycle and explore–exploit: idea generation and experimental iteration \emph{why $\rightarrow$ how}, automated reviewing/archiving as partial return \emph{how $\rightarrow$ why}; autonomous, closed loop. & Isolated from real-world expertise and emerging data; restricted innovative theory construction due to \enquote{data-belief asymmetries}~\citep{felin2024theory} & Literature checks; code execution; automated reviews; end-to-end document production & Autonomous closed-loop pipeline; orchestrated planning/execution/reviewing; no explicit human-in-the-loop \\
OpenAI Deep Research & Planning $\rightarrow$ retrieval $\rightarrow$ synthesis with source attribution and reasoning traceability; primarily evidence synthesis on the \enquote{why} side with limited coupling to concrete experimentation; explore via iterative search \emph{why$\rightarrow$how} is partial; correction by coherence via attribution. & No explicit tacit inlet; reports hallucinations and calibration weaknesses; human oversight advisable & Multi-step plans; multi-role prompting; attribution-aware synthesis; error-prone on authority discernment & Program-orchestrated \enquote{agents} within one system (query generation, summarization, etc.); tracked source–claim mappings \\
\bottomrule
\end{tabularx}
\end{adjustbox}
\caption{Mapping reviewed AI-for-science systems to the constructivist framework (theory–practice dynamics, tacit inlet, reasoning, and specifications).}
\label{tab:sd-tp-1}
\end{table}
Read against Fig.~\ref{fig:open-ended-problem solving}, most systems emphasize the inner \enquote{how} loop (models–experiments–analysis) and only partially realize the return path to \emph{theory}, whereas our constructivist framework in Section~\ref{sec:practice-oe-ps} can contribute the missing governance, e.g.~\emph{teach-back}, human transitions, calibration, and explicit tacit handling, consistent with the outer-loop \enquote{correction by coherence}, to complete the \emph{how$\rightarrow$why} leg and close the constructivist cycle. \par

Taken together, these systems already instantiate major elements of our framework: theory-practice dynamics (rooted in praxis–cognition cycle~\citealp{piaget1970genetic}), explore–exploit loops (mirroring Simon's decision cycle~\citealp{Simon1977,simon1993decision}) coupled to practice (experiments/simulations), recurrent reflection/evaluation, and memory/coordination, while leaving governance gaps our constructivist framework fills: \emph{teach-back} gates, calibration and abstention, human transitions for sensitive updates, and explicit handling of the tacit dimension. This alignment supports the claim that the constructivist scaffold is \emph{generalizable} for open-ended scientific discovery, across heterogeneous domains and agent designs.
\section{Towards open complex human--AI agents collaboration systems}

\subsection{A systems-theoretic ontology for human--AI agent collaboration}\label{sec:sys-ontology-haacs}
Systems exist along continuums of scale (small $\to$ giant) and complexity (linear $\to$ highly non-linear). The degree of openness, i.e., how freely matter, energy, and information traverse boundaries, further shapes emergent coordination, while nested hierarchies modulate near-decomposable interactions. We operationalize these facets in Table~\ref{tab:system-facets}, which condenses definitions and canonical sources.
\begin{table}[htbp]
\centering
\small
\setlength{\tabcolsep}{3pt}
\renewcommand{\arraystretch}{1.05}
\begin{tabular}{p{0.16\textwidth} p{0.62\textwidth} p{0.23\textwidth}}
\toprule
\textbf{System facet} & \textbf{Concise, integrated definition (canonical phrase + essence)} & \textbf{Principal sources} \\
\toprule
\textbf{Openness} &
\textit{\enquote{Closed systems function in isolation, whereas open systems continuously exchange information, energy, and matter with their environment,}} so cross-boundary flows actively reshape internal structure and evolution. &
Wiener~\citep{wiener1959man,wiener2019cybernetics,wiener1988human}; Simon~\citep{simon1996sciences}; Tsien~\citep{xuesen1993new} Miller~\citep{miller1965living,miller1978living} \\
\textbf{Scale} &
Systems range from \textit{\enquote{small, large, giant}} to \textbf{complex giant}; the rising count of subsystems (few $\to$ many $\to$ massive) forces a shift from direct modelling to statistical abstraction. &
Tsien~\citep{xuesen1993new}; Simon~\citep{simon1996sciences}; Miller~\citep{miller1965living,miller1978living}\\
\textbf{Complexity} &
A \textit{\enquote{multifaceted phenomenon emerging from structural interconnections and context-dependent interactions,}} where dense, non-linear links and situational variables yield behaviour \textbf{irreducible to the sum of parts}. & Haken~\citep{haken1988synergetics}; Miller~\citep{miller1965living,miller1978living}  \\
\textbf{Hierarchy} &
\textit{\enquote{Systems can be analyzed into successive sets of subsystems that are nearly decomposable}}, which forms the nested layers whose fast-local / slow-global dynamics \textbf{balance autonomy with interdependence} and accelerate adaptation. &
Simon~\citep{simon1996sciences}; Miller~\citep{miller1965living,miller1978living} \\
\bottomrule
\end{tabular}
\caption{System facets and concise definitions with canonical sources.}\label{tab:system-facets}
\end{table}
\par
\begin{figure}[htbp]
  \centering
  \includegraphics[width=0.66\linewidth]{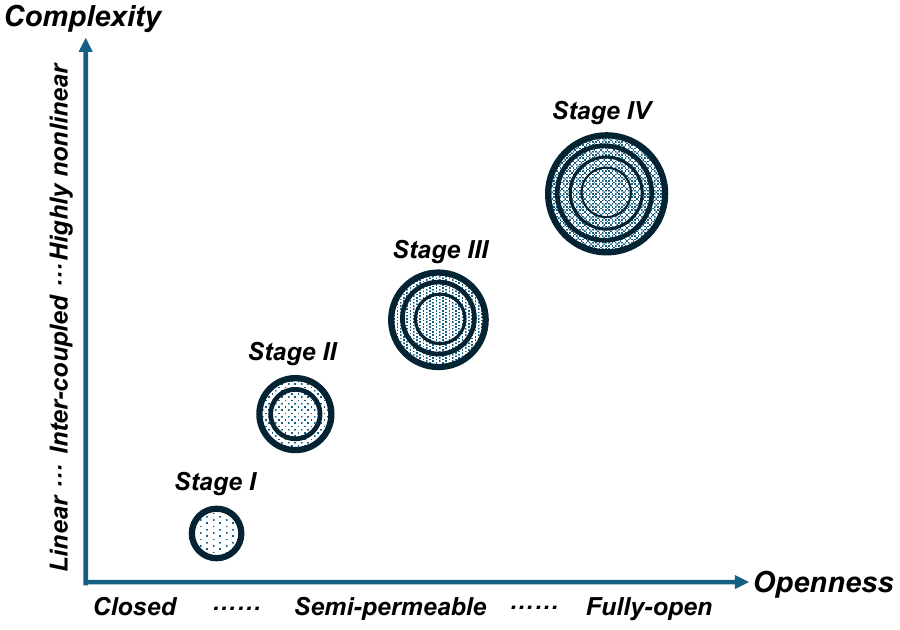}
  \caption{Openness--Complexity ontology of Human--AI agents collaboration.}
  \label{fig:system frame}
\end{figure}

Placing the three historical Human--AI Agents Collaboration System (HAACS) Stages on this openness × complexity plane as shown in Fig.~\ref{fig:system frame} clarifies how the field has moved from closed, simple dyads toward open, complex multi-agent collectives, and why our position paper calls for a fully open complex HAACS design that lives in the top-right quadrant where theory of the living systems~\citep{miller1965living,miller1978living,maturana2012autopoiesis}, meta-synthesis methodology~\citep{xuesen1993new}, synergetic self-organization and non-equilibrium dynamics~\citep{simon1996sciences,haken1988synergetics} all become decisive. Complementing the placement in Figure~\ref{fig:system frame}, Table~\ref{tab:stage-openness-complexity} tabulates the stage–axis mapping and the encodings for scale and hierarchy.

\begin{table}[htbp]
\centering
\small
\setlength{\tabcolsep}{2.8pt}
\begin{tabular}{@{}
p{0.24\textwidth} 
p{0.24\textwidth} 
p{0.24\textwidth} 
p{0.14\textwidth} 
p{0.24\textwidth} 
@{}
}
\toprule
\textbf{Stages} & \textbf{Openness} (X-axis) & \textbf{Complexity} (Y-axis) & \textbf{Scale} (circle size) & \textbf{Hierarchy depth} (concentric rings) \\
\toprule
\textbf{Stage-I} Automation &
\textit{low} -- mostly closed man--machine command loops &
\textit{low} -- linear/feedback control &
\textbf{small} &
1 ring (operator $\leftrightarrow$ device) \\

\textbf{Stage-II} Co-active Teams &
\textit{moderate} -- \enquote{flexible autonomy} with shared norms &
\textit{mid} -- inter-coupled tasks &
\textbf{medium} &
2 rings (human $\leftrightarrow$ agent $\leftrightarrow$ governance shell) \\

\textbf{Stage-III} Agentic AI Collectives &
\textit{high} -- LLM-driven multi-agent swarms, partial observability &
\textit{high} -- non-linear, adaptive &
\textbf{large} &
3 rings (human oversight $\leftrightarrow$ agent cluster $\leftrightarrow$ internal sub-roles) \\

Proposed open-complex HAACS \textbf{(Stage-IV)} &
\textit{very high} -- continuous matter--energy--information exchange with environment &
\textit{very high} -- dynamically adaptive open-complex giant system &
\textbf{giant} &
4+ rings (society $\leftrightarrow$ organization $\leftrightarrow$ team $\leftrightarrow$ agent $\leftrightarrow$ micro-skills) \\
\bottomrule
\end{tabular}
\caption[Stage mapping by openness, complexity, scale, and hierarchy]%
{Stages positioned on the openness--complexity plane; marker size encodes the scale and concentric rings encode the hierarchy depth}
\label{tab:stage-openness-complexity}
\end{table}
\subsection{Mechanism design for the HE\textsuperscript{2}-Net} \label{sec:he2-design-pn}
\subsubsection{From \enquote{Bounded Rationality} to an explore–exploit principle}
In Stage-IV open complex settings, both humans and AI agents operate under \emph{bounded rationality}~\citep{simon1993decision,simon1996sciences}, i.e., limited time, attention, compute, and partial/inaccurate qualitative knowledge force decision makers to \enquote{do the best they can with what they have.} Rather than global optimization, agents set \enquote{aspiration levels}, search locally, and terminate once a satisfactory option is found, i.e., \emph{satisficing} in Simon’s sense \citep{simon1996sciences}. Formally, this is \enquote{procedural rationality} under constraints of information, model fidelity, and resources; practically, this means reasoning with fragmentary observations, defaults, and heuristics, then revising aspirations as feedback accrues. In our ontology, these limits apply equally to symbolic (explicit) structures and to tacit, qualitative schemata that steer attention and inference.\par

We therefore adopt an \emph{explore–exploit principle} as a system policy that adaptively allocates scarce \emph{information–energy–matter} resources between two coupled processes: (i) \emph{exploration}, which designs and runs probes at the moving boundary of the known (interfaces, routines, and concepts) to surface hypotheses; and (ii) \emph{exploitation}, which reuses \emph{validated} knowledge to deliver reliable performance and conserve resources. Narrowly, exploration transforms unknowns into \emph{provisional chunks} (orange), which become \emph{reusable chunks} (green) after passing an epistemic gate; exploitation preferentially draws on these green chunks to shorten future deliberation. Broadly, the meta-level tunes this rhythm to \emph{optimize under dynamic constraints}: not a once-for-all optimum, but a continuous policy that (a) expands the frontier efficiently and (b) stabilizes routine action with codified assets.

\subsubsection{Stage-IV HAACS HE\textsuperscript{2}-Net: components and bidirectional knowledge flow} \label{sec:he2-intro}
\begin{figure}[htbp]
  \centering
  \includegraphics[width=0.68\linewidth]{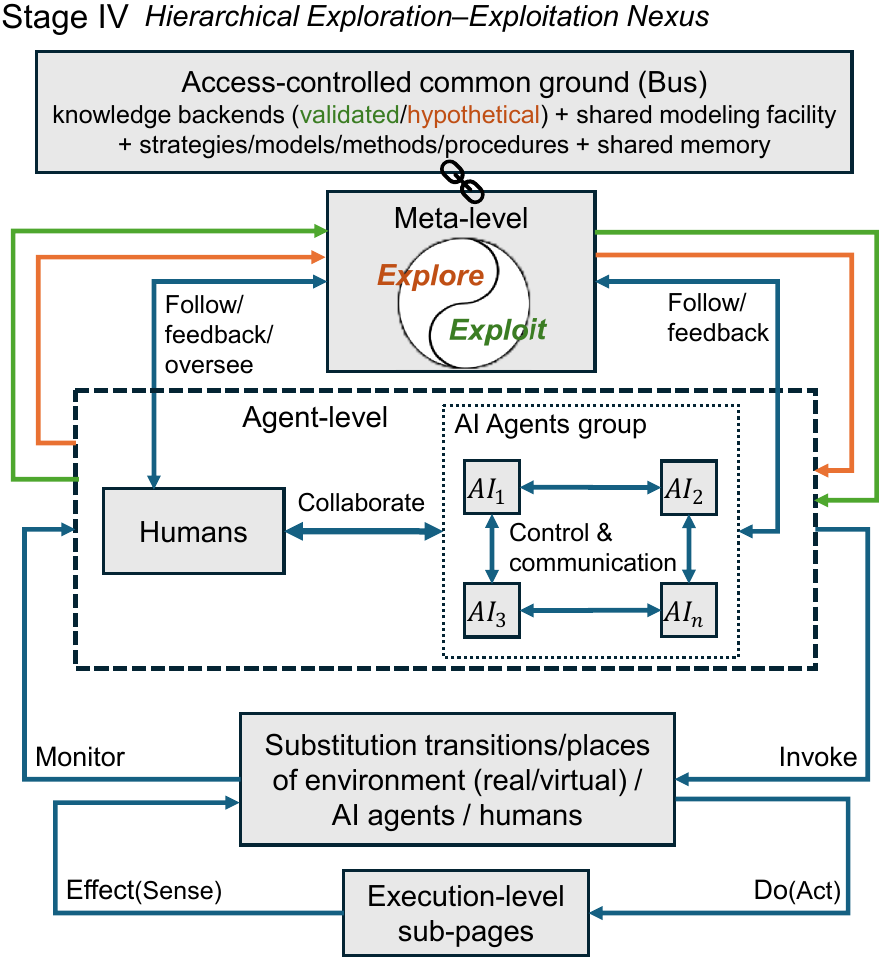}
  \caption{Stage-IV HE\textsuperscript{2} Nexus. The meta-level (Yin--Yang with adaptive slider and integrated common-ground bus) tunes the \emph{explore--exploit} policy and governance; a two-band knowledge backbone separates \emph{provisional/hypothetical} (orange) from \emph{validated/exploitable} (green) knowledge, with an epistemic gate (\emph{Test + Peer Review}) mediating promotion. Orange feed-forward arrows carry \emph{hypotheses/probes} from meta to agents; the execution level (human \& AI) runs SPA$+$Learn micro-loops (inset) to enact experiments and harvest results. Green feedback arrows return \emph{harvested/codified} chunks upward for promotion and policy refresh; validated chunks also flow downward to guide routine MEA-style exploitation. Color semantics: orange = explore/hypothesis; green = exploit/validated.}
  \label{fig:he2-nexus}
\end{figure}

Figure~\ref{fig:he2-nexus} summarizes the HE\textsuperscript{2}-Net for stage-IV HAACS and the orange/green bidirectional flows as follows.\par

\textbf{Components.} \par
(i) \emph{Meta-level (Yin–Yang with adaptive slider).} A policy dial glides between exploration and exploitation, enabling mode switches, safety/ethics constraints, and team reconfiguration (merge/split). In Petri net terms, \emph{order places} and guards enable/disable families of agent-level transitions. \par
(ii) \emph{Knowledge Backbone (two-band store).} A single box is visually split into an \emph{orange right band} for \emph{provisional/hypothetical} knowledge and a \emph{green left band} for \emph{validated/exploitable} knowledge; both are common-ground assets accessible system-wide. \par
(iii) \emph{Epistemic Gate (Test + Peer Review).} A justification gate sits between the bands; only items that pass tests/peer checks migrate forwards (see Section~\ref{sec:practice-oe-ps}). \par
(iv) \emph{Agent/Execution Level (micro-lab inset).} Humans and AI agents enact \emph{Sense–Plan–Act} (with Learn/Reflect) as fine-grained sub-nets; the inset shows the micro-loop (hypothesis $\to$ act $\to$ result $\to$ refine). \par
(v) \emph{Feed-forward/feedback paths.} Slim, labeled arrows connect levels without clutter: orange (hypothesis/probe) and green (harvest/codify) paths.\par

\textbf{How the orange and green paths work (downward and upward).} \par
\emph{Orange (hypothetical) downward.} When the meta-level increases the \enquote{explore} bias, it issues \emph{orange} directives: \enquote{design probes/experiments} targeted at uncertain interfaces, models, or procedures. Concretely, this enables agent-level \emph{boundary/collaboration transitions} that dispatch \emph{hypothesis tokens} to execution sub-pages (S–P–A), allocate tryout budgets, and authorize data capture. Results (including negative or ambiguous outcomes) are stored as \emph{provisional chunks} in the orange band, tagged with provenance and conditions of applicability. \par

\emph{Orange (hypothetical) upward.} During routine exploitation, three conditions can escalate \emph{hypothesis} signals back to the meta level: (i) \emph{anomaly/mismatch} between expected outcomes and observations (concept drift, distribution shift, or violated assumptions); (ii) \emph{novelty/serendipity} detected in the execution traces (unexpected affordances, surprising regularities); and (iii) \emph{exogenous change} (environmental or external rules updates). Each case emits a \texttt{HypothesisReport} token (orange) from agent-level monitoring places to the meta-level via boundary or collaboration transitions, opening a \emph{provisional docket}. The meta-level aggregates these reports, raises the explore bias locally (budgeting targeted probes), optionally \emph{rate-limits or freezes} affected green chunks, and issues scoped \emph{orange directives} to re-test vulnerable interfaces and update operational envelopes. If no provisional patch bypasses the \emph{Test + Peer Review} gate, promotions remain gated; emergency fallbacks rely on previously validated green chunks while orange investigations proceed.\par

\emph{Green (validated) upward.} When local evaluation plus \emph{Test + Peer Review} (the gate) succeed, a \texttt{promotion} transition moves items into the green band as \emph{exploitable chunks}. A \emph{green feedback} then flows upward to the meta level as \emph{policy updates}: SOPs, prompts/checklists, governance rules, and resource weights are refreshed. In Petri net notation, these are logged as authorized reconfiguration events and guard updates that change which agent-level transitions remain enabled under the current explore–exploit mix. \par

\emph{Green (validated) downward.} Validated knowledge in the \emph{green band} feeds back \emph{downward} as \emph{standard operating procedures} and \emph{control knowledge} for MEA. Execution sub-nets exploit these chunks directly (short-circuiting long deliberations), with the meta level controlling their use under safety/ethics constraints. If application context drifts, \emph{mismatch tokens} re-open the orange path for targeted probes; the gate then arbitrates any proposed updates.

\textbf{Petri net alignment in brief.}
Orange/green flows are realized by cross-owner boundary/collaboration transitions, i.e., the off-diagonal \(C_{i,j}\) blocks and the collaboration block \(C_{*,\text{coll}}\) in Eq.~\eqref{eq:mimo matrice}. \emph{Orange downward} is meta\(\rightarrow\)agent enablement of probe transitions that dispatch \texttt{hypothesis} tokens into execution sub-pages; \emph{orange upward} is agent\(\rightarrow\)meta escalation of \texttt{HypothesisReport}/\texttt{DriftAlert}/\texttt{BoundaryShift} tokens from monitoring places. Gate passage is a guarded \texttt{promotion} transition (provisional\(\to\)validated); \emph{green upward} carries policy updates as authorized reconfiguration events (guard flips) over agent-family transitions; \emph{green downward} enables MEA-style \texttt{exploit} transitions to consume validated chunks. The cycle preserves auditability (monotone event log), supports concurrency (multiple probes in flight), and stabilizes exploitation (rate-limited green draws), matching the second-order loop described earlier.
\subsubsection{Synthesis: collaborative knowledge management and problem solving}
Collaborative problem solving and knowledge management unify in a constructivist loop (Section~\ref{sec:practice-oe-ps}) grounded in Conversation Theory (CT) (Section~\ref{sec:col-hai-learn}) and implemented by our three-layer Petri net architecture (Section~\ref{sec:three-layer-haacs}). The \enquote{knowledge backbone} grows through CT’s \enquote{knowing $\leftrightarrow$ coming to know}: multi-level dialogue (task, explanatory, meta), teach-back gates, and a shared micro-world yield an entailment mesh that consolidates into a directed net (see Fig.~\ref{fig:CT-HAI}) \citep{pask1976conversation,boyd2001reflections}, which has implementation similar to that in Section~\ref{sec:metacog-pn}. On the \emph{exploit} side, MEA offers the control scaffold for routine, closed-ended tasks—goal/space framing, operator preconditions, and subgoal management, with the execution-level sub-nets providing the concrete operators (Section~\ref{fig:mea-naive}) \citep{newell1972logic,newell1972infops}. On the \emph{explore} side, practice-guided open-ended inquiry (Kolb/Rescher/Piaget/Polanyi) drives hypothesis generation and revision through experience $\rightarrow$ reflection $\rightarrow$ abstraction $\rightarrow$ experiment, with tacit foreknowledge feeding the cycle (Fig.~\ref{fig:open-ended-problem solving}) \citep{kolb2014experiential,Rescher2020,piaget1970genetic,polanyi2009tacit}. The SECI spiral (Socialization–Externalization–Combination–Internalization) moves insights between tacit and explicit forms and anchors organizational assets \citep{nonaka2000seci}.\par

All three elements, i.e., CT backbone, MEA exploitation, and practice-guided open-ended problem solving (exploration), are coordinated at the \emph{agent-level} by boundary/collaboration transitions (token-based synchronization, guards, rate limits), and tuned at the \emph{meta-level} by policy switches (explore–exploit slider, safety/ethics governance). Thus, humans and AI agents co-navigate a \enquote{theory–practice} dynamic: \emph{learning by doing} produces provisional knowledge; epistemic gates promote it; MEA draws upon the promoted store; CT maintains coherence; and the meta level re-balances the rhythm as conditions change.
\subsubsection{Positioning stage–IV HE\textsuperscript{2}-Net against stages I–III}
To place stage-IV concretely against prior eras, Table~\ref{tab:stage-comparison-7d} contrasts the four stages along a seven-dimension collaboration spine that recurs throughout this paper (cf. Secs.~\ref{sec:previous-hai}, \ref{sec:three-layer-haacs}, \ref{sec:he2-intro}). \par
\begin{table}[htbp]
\centering
\small
\setlength{\tabcolsep}{3.2pt}
\begin{adjustbox}{scale=0.90,center}
\begin{tabularx}{1.18\textwidth}{@{}p{2.4cm} X X X X@{}}
\toprule
\textbf{Dimension} &
\textbf{Stage-I: Automation} &
\textbf{Stage-II: Flexible Autonomy \& Co-active Design} &
\textbf{Stage-III: Agentic-AI \& Multi-Agent Collectives} &
\textbf{Stage-IV: HE\textsuperscript{2}-Net (Open-Complex)} \\
\toprule
\textbf{Agent initiative} &
Low, reactive device control; procedure-bound. &
Context-dependent initiative within role; improvisation at abnormal junctures. &
High local proactivity (tools, reflection, role protocols). &
Policy-controlled proactivity; initiative is \emph{budgeted} by meta-level \emph{explore–exploit} slider with rate-limits; operates under constrained resources and dynamic, nonstationary environments \\
\addlinespace
\textbf{Directability (override)} &
Manual command/abort; hard switches. &
Policy-based retasking; supervisory/\allowbreak mediating/\allowbreak cooperative overrides. &
Prompt/pipeline re-tasking; moderators/voting (latency varies). &
Instant \emph{guard flips} and logged reconfiguration events (meta-level); safe fallbacks via validated green chunks. \\
\addlinespace
\textbf{Policies (guardrails)} &
Hard-wired, static safety rules. &
Norms+policies; some dynamic activation/relaxation. &
Playbooks/\allowbreak SOPs/\allowbreak prompts; ad-hoc dynamic refresh per framework. &
Formal guards over \emph{families of transitions}; promotions via \emph{Test+Peer Review} gate update policy; audit log. \\
\addlinespace
\textbf{Norms (practice)} &
Operator conventions (implicit, local). &
Team conventions as coordination glue. &
Protocolized roles; memory-mediated conventions across agents. &
Institutionalized norms coupled to CT/SECI; backbone records and stabilizes practice. \\
\addlinespace
\textbf{Common ground (CG)} &
Status displays and local logs. &
Ongoing situational awareness (SA) dashboards; shared progress appraisals. &
Shared dialog buffers/blackboards; partial dynamic CG. &
System-wide \emph{knowledge backbone} with \emph{orange} (provisional) / \emph{green} (validated) bands and an epistemic gate. \\
\addlinespace
\textbf{Planning horizon} &
Single-step/reactive loops. &
Explicit multi-step; function allocation adapts. &
Multi-step action sequences; emerging hierarchy under memory/tools. &
\emph{Hierarchical} (meta-agent-execution) with concurrent probes; MEA exploitation $+$ open-ended exploration. \\
\addlinespace
\textbf{Communication modality} &
Minimal signals, alerts; physical controls. &
Language \& explanations within supervisory/mediating protocols; GUI support. &
Explicit messages $+$ implicit embodied cues; tool outputs/traces. &
Multimodal \emph{languaging} $+$ capacity-aware GUI affordances $+$ \emph{formal event logs}; orange/green feedback paths. \\
\bottomrule
\end{tabularx}
\end{adjustbox}
\caption{Four-stage comparison along the seven-dimension collaboration spine. Stage-IV (HE\textsuperscript{2}-Net) operationalizes meta-level guard flips, an epistemic promotion gate, and a system-wide knowledge backbone, aligning with the hierarchical Petri net in Sec.~\ref{sec:three-layer-haacs} and the orange/green flows in Sec.~\ref{sec:he2-intro}.}
\label{tab:stage-comparison-7d}
\end{table}
Left$\to$right, three structural shifts are salient: (i) from static, device-level control to meta-level \emph{guard-based} reconfiguration; (ii) from ad-hoc, local common ground to a \emph{two-band knowledge backbone} with a \emph{Test+Peer Review} gate; and (iii) from single-step or ad-hoc multi-step plans to \emph{hierarchical} planning under an adaptive explore–exploit policy.\par

These stage-IV behaviors are realized by cross-owner boundary/collaboration transitions (off-diagonals and $C_{*,coll}$ in Eq.~\ref{eq:mimo matrice}) and meta-level guard flips that enable/disable families of agent-level transitions (Sec.~\ref{sec:three-layer-haacs}).
\subsubsection{An instantiation of HE\textsuperscript{2}-Net}
To demonstrate the modularity and flexibility of Petri nets in modeling interactions and coordination within the internal modules of individual agents and among the agents (including humans), epistemic justification and \emph{explore-exploit} mechanisms are modeled as sub-nets running simultaneously with the main SPA path by branching token flows between the main SPA path and these new sub-nets in both directions.\par

For epistemic justification mechanism, each agent maintains or uses certain pieces of knowledge (rules, norms and policies, beliefs, partial observations, etc), epistemic justification verifies whether that piece of knowledge or belief is valid, consistent with the environment, or accepted by other agents and human. We must capture both the agent's own internal check (intra-agent) and any collaborative verification among multiple agents (inter-agent), conforming to the conventions in Eq. (\ref{eq:mimo matrice}). Thus \textbf{Epistemic Justification Sub-Net} is created to focuses on verifying new or updated knowledge. For \emph{explore-exploit} dynamics, during \emph{exploration} phase: humans (or specifically designated \enquote{explorer} roles assigned to AI agents) generate or refine new knowledge/rules when novelty or anomalies arise; during \emph{exploitation} phase: AI agents apply known (verified) knowledge or routines to handle routine tasks, stabilizing the system; note that the epistemic justification sub-net ties into this dynamics by ensuring that newly explored knowledge is validated before entering the \emph{exploitable} knowledge base. Thus \textbf{Explore-Exploit Sub-Net} is modeled to coordinate how the agent alternates between exploring new knowledge and exploiting known rules. We have the two new sub-nets within an individual agent as follows:\par
\textbf{Epistemic Justification Sub-Net}: (1) \textbf{Intra-Agent} Epistemic Justification: within each agent's Petri net, there exists a set of places storing that agent's current knowledge and rules (e.g., \texttt{Rule Repository} for an AI agent, or \texttt{Human Known Solutions} for a human expert), the new or modified knowledge claims are placed here in a \enquote{Provisional} form initially; the transitions for epistemic justification consumes a \texttt{Provisional Knowledge} token plus any required data (observations, prior knowledge, test results) from other places, optionally the transitions might require an \texttt{Energy} token (representing cognitive or computational effort to perform verification), and the transitions produce either \texttt{Validated Knowledge} (if the verification passes) or \texttt{Refuted}/\texttt{Requires Further Inspection} tokens (if the check fails or is inconclusive) which may require further processes to gather data from the theory-guided experimentation controlled by another sub-net to create; the transitions from other sub-nets can loop back to an \texttt{Exploration} place with \texttt{New Data} token, prompting the agent refine or test the knowledge further; if validated, the token moves into a separate place representing \texttt{Exploitable Knowledge}, making it available for routine use. (2) \textbf{Inter-Agent} Epistemic Justification: when multiple agents must confirm or align on knowledge, an agent posts a knowledge claim (e.g., \enquote{I believe solution X is valid for scenario Y}) to its place associated with its collaboration transition \texttt{Multi-Agent Verification} transition, complying with Eq. (\ref{eq:mimo matrice}); the \texttt{claim} token references (or \enquote{carries}) the relevant knowledge content. This transition can fire only if each relevant agent contributes a \texttt{consent} or \texttt{review} token. Each agent's local sub-net checks the claim (possibly generating more internal tokens or feedback transitions). If all agents produce \texttt{Agreement} tokens, the knowledge claim is validated at the inter-agent level (becomes \texttt{Common Knowledge} or \texttt{Shared Verified Knowledge}). If any agent produces \texttt{Disagreement} or \texttt{Doubt} the token is diverted back to the original claiming agent's exploration net for revision or deeper testing.\par

\textbf{Explore-Exploit Sub-Net}: this sub-net has exploration and exploitation places and associated transitions. Within \texttt{Exploration} module, if a piece of knowledge is found wanting (internally or collectively) and promising based on the tacit clues (more details in Section~\ref{sec:he2-intro}), the system triggers a new exploration cycle to discover, refine it or generate alternatives; its transitions trigger when novelty or anomalies appear in the agent's input and might generate new knowledge claims (placed in a \texttt{Provisional Knowledge} place). Within \texttt{Exploitation} module, only validated knowledge is used by the exploitation transitions, ensuring that the AI agents or the humans do not rely on unverified or faulty claims during routine tasks; its transitions consume a \texttt{Validated Knowledge} token plus a \texttt{Task} token (assigned by others or self-generated, based on agent roles or load balancing) to produce an \texttt{Outcome} typically representing some successful operation or a result for the environment, which will be archived to the \texttt{Episodic Memory} module.
\subsection{Synthesizing biological cybernetics, multi-scale goals, and evolving boundaries} \label{sec:he2-noneq-dynamics}
\subsubsection{Unpacking autopoiesis and autogenesis}
Coined by Maturana \citep{maturana2012autopoiesis}, autopoiesis refers specifically to a system's capacity to continuously regenerate itself through internal interactions and processes, maintaining its organizational identity. It involves clear boundary maintenance that distinguishes the entity from its environment, while autogenesis refers to the self-initiated development, self-organization, and emergence of structural complexity. Autogenesis emphasizes developmental processes and self-driven evolution within a system's boundary. A boundary defines the distinction between the system (agent) and its environment. In autopoietic systems, maintaining this boundary is fundamental to identity. An agent formally in Section~\ref{sec:cyber-agenthood} and modeled in Section~\ref{sec:sys-formal-pn} can be configured to exhibit autopoiesis, it autonomously maintains its boundary, continuously rebuilding itself from within. 
\subsubsection{Embodying purposeful agency with BDI}
Each agent defined in Section \ref{sec:cyber-agenthood} is an adaptive goal-seeking system with two levels of goals with different time scales: (1) \textbf{externally assigned (ad-hoc) goals} are temporary, situational tasks dictated by the environment, other agents or higher-level authorities like the humans. These are tactical and context-dependent, like immediate task assignments; (2) \textbf{internally maintained (long-term) goals (Desires, Intentions)} are strategic, stable, and identity-defining. These reflect deeply rooted aims, often aligned with personal identity, professional role, or long-term vision, akin to a professional profile, career orientation or organizational loyalty \citep{simon1996sciences} in human terms. In other words, an agent isn't just an executioner of tasks but a purposeful, identity-driven entity.\par
Inspired by the belief-desire-intention (BDI) model of human practical reasoning \citep{Bratman1987-BRAIPA} as a way of explaining future-directed intention, each agent has (1) \textbf{beliefs}: internal representation of its own state, boundary conditions, environment, and historical knowledge; (2) \textbf{desires}: stable preferences or overarching aims, analogous to values or long-term interests of humans. These could reflect the self-maintaining organizational closure emphasized by autopoiesis, maintaining an identity over time; (3) \textbf{intentions}: the specific commitments an agent adopts to satisfy its desires. Intentions translate abstract desires into concrete courses of action. They reflect the agent's active choice in navigating its complex environment.

\subsubsection{Harmonizing autopoiesis and autogenesis with exploitation and exploration}
Agents achieve genuine adaptability by continuously balancing identity maintenance with adaptive transformation, thriving amid dynamic complexity, through an elegant interplay of autopoiesis and autogenesis that addresses the perpetual tension between stability (identity preservation) and adaptation (boundary-expanding growth). Specifically, agents are defined along two dimensions: structural stability (autopoiesis), which ensures continuity, maintains internal integrity, short-term efficiency, and stable boundary management (exploitation) by using known strategies and resources including existing structure and functions; and structural adaptability (autogenesis), which fosters innovation by actively incorporating external resources (energy, matter, information) to evolve internal structures and drive emergent self-growth (exploration). This exploitation-exploration dynamic not only mirrors the classic trade-off observed across multi-scale, diverse systems but also endows agents with genuine purposefulness, enabling them to proactively shape their own evolution with human-level meaningful intentionality and deliberate action.

\subsubsection{Evolving boundaries through splitting and merging agents}
Complexity and openness discussed in Section~\ref{sec:sys-ontology-haacs} supports modelling the HAACS as fluid and nested: one agent can consist of nested sub-agents (sub-systems), each potentially autopoietic. This nesting suggests that agency is inherently multi-scale, hierarchical \citep{miller1978living}, and recursive. Agency emerges from interactions, thus the notion of boundary is contextually defined and observer-dependent, what constitutes an \emph{agent} at one scale may simply be a subsystem or even environment at another. Boundaries can shift, evolve, and adapt. However, merely \enquote{cutting} an agent apart physically or computationally does not guarantee maintaining independent agency unless each part has internal organizational autonomy and closure.\par

\textbf{Dividing into sub-agents}: if we take an autopoietic, autogenetic agent and conceptually divide it, each resulting sub-agent could only maintain genuine \enquote{agenthood} if it, too, retains the capacity for autopoiesis and autonomy, i.e., each sub-unit must retain a capacity for autonomous identity preservation, boundary maintenance, and self-production. Without that, it becomes merely a functional module or subsystem, not truly an independent \emph{agent}.\par
\textbf{Merging into larger agents}: merging multiple agents implies integration of boundaries into a new meta-boundary. From a cybernetic perspective, this forms a new system identity, a higher-order autopoietic agent, if and only if it establishes a coherent organizational closure and self-maintenance at the new level, i.e., the larger agent as an integrated system needs to achieve a new autopoietic boundary, a distinct organizational identity, and emergent autonomy above and beyond the sum of the parts.\par

In practice, there might be meta-level places that count how many tasks are in the system, or measure concurrency load. If two or more sub-nets (originally thought to belong to separate agents) cannot function independently or be strongly coupled, i.e., they share so many places or transitions that they always fire in lockstep, whether to \enquote{merge} or keep \enquote{split} these sub-nets upon crossing a threshold concurrency measured in real-time depends entirely on the system's design goals and resource-management philosophy, e.g., \enquote{do you want to distribute their functions by modelling them as separate agents with heavy synchronization or a single larger agent with multiple internal sub-tasks?} Some architectures merge multiple small agents during high load to unify knowledge or resources, reducing inter-agent communication overhead (centralizing control in one \enquote{super-agent}), then split again when load is low to eliminate unnecessary overhead. Others do the exact opposite, splitting under heavy load to increase parallelism, offload heterogeneous sub-tasks, or specialize sub-agents (like differentiation in \citealp{miller1978living}), and merging only when tasks are sparse and homogeneous. The key is to decide which strategy, centralization vs. distributed parallelism, best serves the needs under varying load conditions, since both approaches can be valid in a reconfigurable setting of HE\textsuperscript{2}-Net.
\subsubsection{Driving agent-level self-organization through meta-level transitions}
A special \texttt{order} place can be defined to reconfigure the HE\textsuperscript{2}-Net mainly controlled by the meta-level. It represents a global or higher-level state of the system, akin to an \enquote{order parameter} in \enquote{enslaving principles} of synergetics~\citep{haken1988synergetics}. Its marking determines which \enquote{meta-transitions} (the structural changes to merge/split agents and associated sub-nets, reconfigure the collaboration flows among agents, etc.) become enabled. By tracking a system-wide condition in just a few such places, we can effectively let these \enquote{order places} drive major reconfigurations, thereby \emph{enslaving} the rest of the net's local details to the new global arrangement.\par

Self-organization can be achieved, and the reconfiguration can be emergent (rather than orchestrated by an external supervisor), we can define additional meta-level transitions or control policies that cause the \enquote{order transitions} to become enabled under certain macroscale conditions. For instance, the pivotal communication structure reconstructed by \enquote{Criticize-Reflect} method in \citep{guo2024embodied} are analogous to the macro order parameter tipping the system into a new organizational mode, and its strategy can be modeled and refined further formally and analytically in our framework. Hence, the firing of these \enquote{order} relevant transitions effectively enslaves the rest of the Petri net's structure to the newly emerging global pattern, mirroring Haken's principle.

\subsection{Meta-synthesis for open complex giant systems \emph{vis-à-vis} the stage-IV HE\textsuperscript{2}-Net}\label{sec:ms-vs-he2}
\subsubsection{Tsien’s meta-synthesis in brief (essence for open complex giant systems)}
Tsien’s meta-synthesis proposes a \emph{from qualitative to quantitative} methodology explicitly dedicated to \emph{open complex giant systems}, where reductionist \enquote{exact science} methods are inadequate \citep{xuesen1993new}. Its core move is to \emph{unite} an expert collective, empirical data, reference materials, and computer simulation into a single problem-solving organism. Experts advance \emph{empirical hypotheses} (qualitative judgements, often theory- and experience-laden), which are instantiated in models, simulated, analyzed, and optimized; results are repeatedly reviewed and the models adjusted until \emph{consensus} is reached, yielding the best available quantitative conclusions grounded in qualitative understanding \citep{xuesen1993new}. In Tsien’s framing, the approach is systemically integrative (cross-disciplinary), practice-aware, and inherently iterative: qualitative comprehension $\rightarrow$ quantitative modeling $\rightarrow$ expert synthesis $\rightarrow$ policy proposals \citep{xuesen1993new}. 
\begin{figure}[htbp]
  \centering
  \includegraphics[width=1\linewidth]{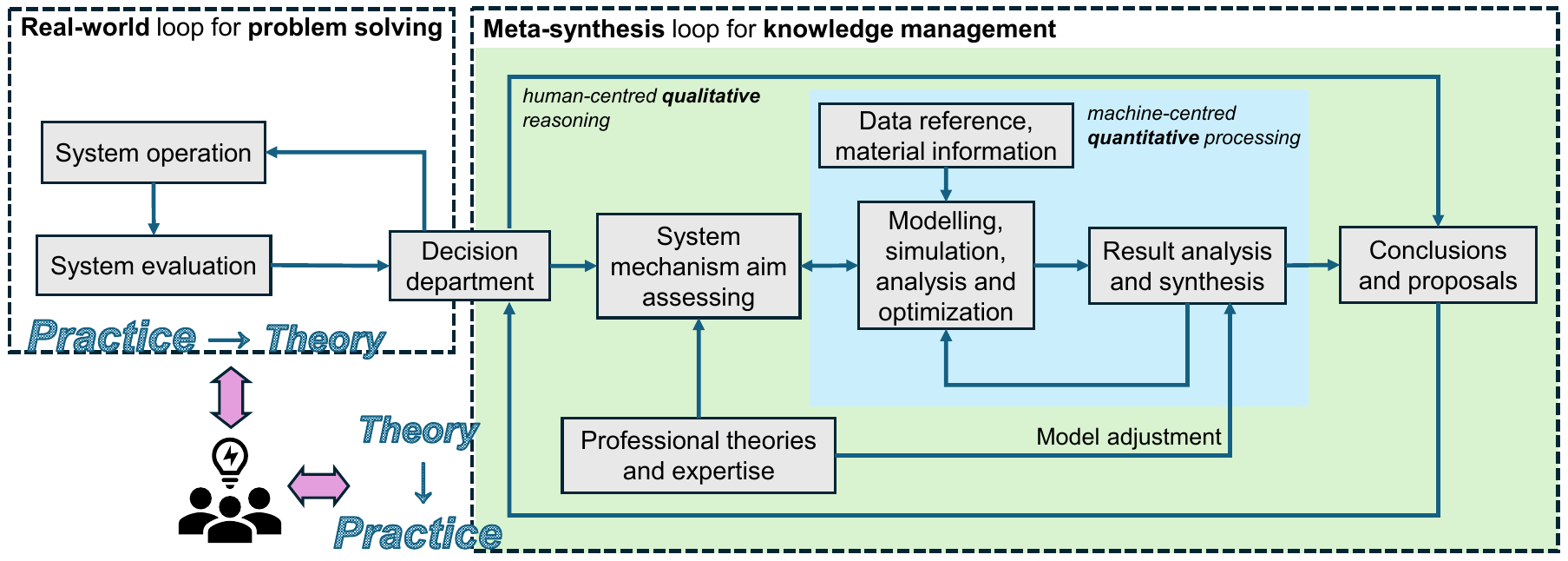}
  \caption{Meta-synthesis loop for knowledge management, adapted from Fig.~1 in \citet{xuesen1993new}. \textit{A decision department couples the real-world loop for problem solving with the meta-synthesis loop: human-centered qualitative reasoning (\enquote{professional theories and expertise}; \enquote{system mechanism aim assessing}) steers machine-centered quantitative processing (data/material reference $\rightarrow$ modelling, simulation, analysis, optimization $\rightarrow$ result analysis and synthesis); \enquote{model adjustment} closes friction between theory and practice, yielding \enquote{conclusions and proposals} that return to operations.}}
  \label{fig:meta-synthesis-loop}
\end{figure}

\subsubsection{Reading the meta-synthesis diagram as theory–practice dynamics}
As illustrated in the meta-synthesis loop (Fig.~\ref{fig:meta-synthesis-loop}), a \enquote{decision department} couples the \emph{real-world loop} (system operation $\leftrightarrow$ evaluation) with the \emph{meta-synthesis loop}. Human-centered \emph{qualitative reasoning} (\enquote{professional theories and expertise}; \enquote{system mechanism aim assessing}) feeds a machine-centred \emph{quantitative pipeline} (data/material reference $\rightarrow$ modeling, simulation, analysis, optimization $\rightarrow$ result analysis and synthesis). Bidirectional arrows (\enquote{model adjustment}) register friction between theory and practice, while \enquote{conclusions and proposals} close the loop back to operations. In short: human experts remain primary participants whose qualitative judgments steer and are tested by computational experiments, making meta-synthesis a structured, practice-grounded dialogue between experience and model \citep{xuesen1993new}.
\subsubsection{Transition to the comparison}
To position our HE\textsuperscript{2}-Net against Tsien’s meta-synthesis, Tables~\ref{tab:he2-vs-ms-p1}-\ref{tab:he2-vs-ms-p2} enumerate their \emph{functions and structures}, and then makes their \emph{salient divergences} explicit. The mapping highlights common commitments (iterative theory–practice coupling; mixed qualitative/quantitative reasoning) and clarifies how HE\textsuperscript{2}-Net generalizes meta-synthesis for stage-IV open complex HAACS with epistemic gates, a two-band knowledge backbone, and meta-level \emph{explore–exploit} control (cf. Secs.~\ref{sec:col-hai-learn}, \ref{sec:he2-intro}).
\begin{table}[htbp]
\centering
\small
\begin{adjustbox}{scale=0.88,center}
\begin{tabularx}{1.16\textwidth}{@{}p{0.20\textwidth} 
p{0.31\textwidth} 
p{0.31\textwidth} 
p{0.26\textwidth}
@{}}
\toprule
\textbf{Dimension} & \textbf{HE\textsuperscript{2}-Net (the proposed)} & \textbf{Meta-synthesis methodology} & \textbf{Comparison} \\
\toprule
\textbf{Problem class \& scope} & Stage-IV open complex HAACS; \emph{explore--exploit} policy tunes resources; explicit stage-IV placement on openness$\times$complexity plane (Fig.~\ref{fig:system frame}). &
Open complex giant systems where reductionism fails; method expressly designed for such systems \citep{xuesen1993new}. &
Both target open complex settings; HE\textsuperscript{2} emphasizes \emph{agentic collectives} and dynamic policy control. \\
\addlinespace
\textbf{Participants \& roles} &
Humans \emph{and} AI agents as first-class P-/M-individuals \citep{pask1976conversation}; multi-agent languaging bus; teach-back gates. &
Human expert group in a \enquote{discussion hall} (early cyberspace); computers support modeling/simulation. &
HE\textsuperscript{2} extends the expert collective with \emph{agentic AI} and formal coordination protocols. \\
\addlinespace
\textbf{Knowledge flow (qualitative $\leftrightarrow$\allowbreak quantitative)} &
Two-band backbone: \emph{orange} (provisional)$\rightarrow$epistemic gate (\emph{Test + Peer Review})$\rightarrow$\emph{green} (validated); CT/SECI circulate tacit$\leftrightarrow$explicit \citep{pask1976conversation,nonaka2000seci,polanyi2009tacit}. &
Qualitative hypotheses $\rightarrow$ model+\allowbreak simulation+\allowbreak analysis+\allowbreak optimization $\rightarrow$ expert synthesis; iterate until consensus. &
Both mix qualitative/quantitative; HE\textsuperscript{2} adds an explicit \emph{promotion gate} and two-band common ground. \\
\addlinespace
\textbf{Loop structure (theory--practice)} & Open-ended discovery + MEA for routine exploitation; anomalies trigger exploration (Fig.~\ref{fig:open-ended-problem solving}). &
Operations $\leftrightarrow$ evaluation $\rightarrow$ modeling $\rightarrow$ simulation $\rightarrow$ analysis $\rightarrow$ conclusions $\rightarrow$ operations (practice $\rightarrow$ theory $\rightarrow$ practice) (Fig.~\ref{fig:meta-synthesis-loop}). &
HE\textsuperscript{2} explicitly \emph{doubles} the loop (open-ended exploration \emph{and} closed-ended MEA) under one policy. \\
\addlinespace
\textbf{Common ground / memory} &
System-wide \emph{knowledge backbone} (mesh $\to$ directed net) with provenance, weights, merge/split operators. &
Reference data/materials + model repositories + expert rationales. &
HE\textsuperscript{2} formalizes CG as a \emph{two-band}, self-organizing net with learning operators. \\
\addlinespace
\textbf{Validation \& justification} &
Teach-back gates; \emph{Test + Peer Review} epistemic gate; validation call-outs. &
Expert review and consensus after simulation+analysis; model adjustment until agreement \citep{xuesen1993new}. &
Both verify iteratively; HE\textsuperscript{2} operationalizes \emph{promotion} and \emph{rate-limited use} of validated chunks. \\
\bottomrule
\end{tabularx}
\end{adjustbox}
\caption{Stage-IV HE\textsuperscript{2}-Net \emph{vs.} Meta-synthesis: overlapping functions/structures and salient divergences - Part I.}
\label{tab:he2-vs-ms-p1}
\end{table}
\begin{table}[htbp]
\centering
\small
\begin{adjustbox}{scale=0.88,center}
\begin{tabularx}{1.16\textwidth}{@{}p{0.20\textwidth} 
p{0.31\textwidth} 
p{0.31\textwidth} 
p{0.26\textwidth}
@{}}
\toprule
\textbf{Dimension} & \textbf{HE\textsuperscript{2}-Net (the proposed)} & \textbf{Meta-synthesis methodology} & \textbf{Comparison} \\
\toprule
\textbf{Control / governance} &
Meta-level \emph{guard flips}, policy dial (explore--exploit), safety/ethics constraints; hierarchical Petri net reconfiguration (Sec.~\ref{sec:three-layer-haacs}). &
\enquote{Decision department} synthesizes results and issues proposals; governance is committee-centric \citep{xuesen1993new}. &
HE\textsuperscript{2} turns governance into \emph{formal guards} and logged reconfigurations, not just committee decisions. \\
\addlinespace
\textbf{Planning \& operators} &
MEA interpreter for routine tasks; boundary/collaboration transitions coordinate concurrent probes (Fig.~\ref{fig:mea-naive}). &
Model-based analysis and \emph{optimization} to propose policies \citep{xuesen1993new}. &
HE\textsuperscript{2}: symbolic control + concurrent probes; Meta-synthesis: simulation+optimization centric. \\
\addlinespace
\textbf{Openness to environment} &
Continuous matter–energy–information exchange; anomaly/novelty/drift alerts escalate exploration (Sec.~\ref{sec:he2-intro}). &
Empirical data and environmental information drive modeling and adjustment \citep{xuesen1993new}. &
Both are open; HE\textsuperscript{2} adds automated \emph{escalation signals} and policy-coupled responses. \\
\addlinespace
\textbf{Tacit knowledge handling} &
Explicit SECI spiral within Ba; \emph{attending-from-to} (Polanyi) as tacit inlet. &
Primarily via human expert judgment and experience in the discussion hall \citep{xuesen1993new}. &
HE\textsuperscript{2} instruments tacit$\leftrightarrow$explicit conversion as a first-class mechanism. \\
\addlinespace
\textbf{Computation role} &
Agentic AI executes SPA$+$Learn micro-loops; epistemic/logging infrastructure; Petri net semantics. &
Computers support modeling, simulation, analysis, optimization; human-led control. &
HE\textsuperscript{2} moves from \emph{tool} to \emph{teammate}: AI agents act and learn under policy. \\
\addlinespace
\textbf{Scale \& hierarchy} & 4+ rings (society$\leftrightarrow$\allowbreak organization$\leftrightarrow$\allowbreak team$\leftrightarrow$\allowbreak agent$\leftrightarrow$micro-skills); merge/split of agents under order parameters (Sec.~\ref{sec:sys-ontology-haacs} and \ref{sec:he2-noneq-dynamics}). &
Multi-variable, cross-domain modeling at system level; hierarchy implicit in domain structure \citep{xuesen1993new}. &
HE\textsuperscript{2} makes hierarchy \emph{operational} (merge/split, guards); meta-synthesis keeps it conceptual. \\
\bottomrule
\end{tabularx}
\end{adjustbox}
\caption{Stage-IV HE\textsuperscript{2}-Net \emph{vs.} Meta-synthesis: overlapping functions/structures and salient divergences - Part II.}
\label{tab:he2-vs-ms-p2}
\end{table}

\subsection{Quasi stage-IV empirical implementations situated within the HE\textsuperscript{2}-Net frame}
\subsubsection{Essential positioning of recent implementations as \enquote{quasi} stage-IV HAACS}
The HE\textsuperscript{2}-Net conceptual model for stage-IV open complex HAACS designs has several of its functionalities and structures already partially realized in recent systems. We select representative advances to show how each empirical work or framework matches aspects of HE\textsuperscript{2} and where HE\textsuperscript{2}-Net can further improve the design. In all cases, these systems implement elements of \emph{exploration} and \emph{exploitation}, memory or archives, and some form of coordination; yet most lack a system-wide \emph{two-band knowledge backbone} with an explicit \emph{Test + Peer Review} promotion gate, meta-level \emph{guard flips} over families of transitions, and rate-limited exploitation flows (cf. Secs.~\ref{sec:three-layer-haacs}, \ref{sec:he2-intro}). Hence we regard them as \enquote{quasi} stage-IV. \par

\emph{AFLOW}~\citep{zhang2025aflow} automates search over \emph{agentic workflows as code graphs} using an MCTS loop to propose, evaluate, and backpropagate edits. Its selection–expansion–evaluation pipeline is an explicit \emph{explore–exploit} procedure that balances novelty and quality under a cost budget. AFLOW also keeps a tree-structured experience base and multi-run evaluation feedback, partially echoing a store of validated routines. What is missing relative to HE\textsuperscript{2}-Net is a formal epistemic \emph{promotion} gate (orange$\to$green), meta-level \emph{policy dials} (guard flips over operator families), and a common \emph{knowledge backbone} spanning teams with logged reconfiguration events. \par

\emph{Darwin Gödel Machine (DGM)}~\citep{zhang2025darwin} realizes open-ended improvement via a quality–diversity archive: parent selection multiplies a performance pressure with a novelty pressure, producing a standing \emph{explore–exploit} balance. The archive preserves \enquote{stepping stones} (multiple probes in flight), and each self-modification is validated on benchmark suites before retention. Relative to HE\textsuperscript{2}-Net, DGM lacks a system-level \emph{promotion} gate and \emph{rate-limited} design for exploitation, as well as \emph{guard-based} reconfiguration and cross-agent boundary/collaboration transitions with an audit log (Sec.~\ref{sec:three-layer-haacs}). \par

\emph{AutoAgents}~\citep{chen2024autoagents} synthesizes a \emph{team} and a \emph{plan} via Planner + Observers, then executes with \emph{self- and collaborative refinement} under an Action Observer, backed by tri-level memory (short-/long-/dynamic). This aligns with our \emph{boundary/collaboration} transitions and multi-level dialogue (task, explanatory, meta) from Conversation Theory (Sec.~\ref{sec:col-hai-learn}), but it does not separate \emph{provisional} vs \emph{validated} knowledge in a two-band backbone, nor does it formalize \emph{epistemic justification} gates and \emph{policy guard flips}. \par

\emph{AutoAgent} (zero-code Agent OS)~\citep{tang2025autoagent} introduces \emph{event-driven XML workflows} with RESULT/ABORT/GOTO actions and a transformed tool-use layer. It supports reuse (profiling) and repeatable execution under a dispatcher (stabilized exploitation), yet remains one-request scoped by default. HE\textsuperscript{2}-Net would add \emph{system-wide common ground}, \emph{order places} for macro reconfiguration (e.g., merge/split agents), and formal \emph{promotion} and \emph{guard} semantics over families of transitions. \par

\emph{ALITA}~\citep{qiu2025alita} begins with \emph{minimal predefinition} and achieves \emph{maximal self-evolution} by \emph{growing tools as MCP servers} on demand, caching and reusing them thereafter. Its Manager–Web–(brainstorm/code/run) loop instantiates a pragmatic orange$\to$green passage (health-check and reuse), and its MCP registry acts as a proto \enquote{exploitable knowledge} base. HE\textsuperscript{2}-Net would instrument this with an explicit \emph{epistemic gate}, \emph{rate-limited flows}, and \emph{meta-level guard flips} to tune initiative under constraints.\par

\emph{TextGrad} (textual gradients)~\citep{yuksekgonul2025optimizing}~performs \enquote{textual backpropagation} through compound systems, repairing prompts, intermediate artifacts, and solutions using LLM-as-critic signals. It thus operationalizes \emph{teach-back gates} and per-module \emph{justification} loops (Sec.~\ref{sec:metacog-pn}). What it adds to stage-IV is systematic \emph{repair} across a chain; what it misses is a \emph{two-band backbone} with \emph{promotion} and \emph{policy-controlled proactivity} (explore–exploit slider) at the meta level. \par

In sum, these \enquote{quasi} stage-IV systems realize important pieces, e.g., balancing exploration-exploitation flows, archives/memories, event/workflow control, self-evolution, and textual repair, while HE\textsuperscript{2}-Net provides the unifying \emph{meta-level governance} (guard flips, auditability), the system-wide \emph{knowledge backbone} (orange/green bands with a \emph{Test + Peer Review} gate), and explicit \emph{boundary/collaboration} transitions that coordinate humans and agents across scales.
\subsubsection{Mapping \emph{explore–exploit} instantiations and \emph{HE\textsuperscript{2}-Net} boosts}
Table~\ref{tab:impl-explore-exploit} summarizes how each implementation embodies the \emph{explore–exploit} principle and how HE\textsuperscript{2}-Net can \emph{boost} it by adding the epistemic gate, the two-band backbone, guard-based reconfiguration, and logged rate-limited exploitation. The first column captures the concrete mechanism for probe generation and validation (orange flow) and for routine reuse (green flow). The second column lists targeted HE\textsuperscript{2} augmentations: \emph{promotion} (provisional$\to$validated), \emph{guard flips} over families of transitions, \emph{rate-limits} on green draws, and \emph{boundary/collaboration} transitions with audit logs. \par
\begin{table}[htbp]
\centering
\small
\setlength{\tabcolsep}{3.2pt}
\begin{adjustbox}{scale=0.90,center}
\begin{tabularx}{1.16\textwidth}{@{}p{0.22\textwidth} 
X 
X
@{}}
\toprule
\textbf{Implementation} & \textbf{Explore–exploit principle (instantiation)} & \textbf{HE\textsuperscript{2}-Net boost (how to improve)} \\
\toprule
\textbf{AFLOW}~\citep{zhang2025aflow} & MCTS balances \emph{selection} (exploit best workflows) with \emph{expansion} (explore edits); multi-run evaluation acts as validation; a tree of candidates stores experience for reuse. & Add an explicit \emph{epistemic promotion gate} and a \emph{two-band knowledge backbone}; introduce \emph{meta-level guard flips} to switch operator families; \emph{rate-limit} flows; record \emph{audit} events for reconfiguration. \\
\addlinespace
\textbf{Darwin Gödel Machine}~\citep{zhang2025darwin} & Quality–diversity sampling ($\sigma(\text{perf})\times\frac{1}{1+n}$) trades off exploitation of high performers with exploration of under-tried lineages; archive preserves stepping stones for future probes. & Couple archive with a \enquote{Test + Peer Review} \emph{promotion} gate (orange$\to$green); expose \emph{explore–exploit slider} at meta level; add \emph{boundary/collaboration} transitions for human/AI review; \emph{rate-limited} exploitation consumes exclusively from the validated (green) store. \\
\addlinespace
\textbf{AutoAgents}~\citep{chen2024autoagents} & Planner + Observers synthesize a team/plan; \emph{self-} and \emph{collaborative refinement} iterate (explore) until consensus; long-/dynamic memory supports stabilized reuse (exploit). & Embed a \emph{two-band backbone} with \emph{promotion}; make Observer approvals \emph{teach-back} gates; add \emph{guard flips} over plan/operator families; log reconfigurations; connect steps via \emph{boundary/collaboration} transitions. \\
\addlinespace
\textbf{AutoAgent}~\citep{tang2025autoagent} & Orchestrator explores ad-hoc decompositions; event-graph dispatcher executes validated patterns repeatedly (exploit) with RESULT/ABORT/GOTO control. & Globalize memory as a system-wide \emph{knowledge backbone} (orange/green) with \emph{promotion}; add \emph{order places} for merge/split of agents; meta-level \emph{guard} policies to re-wire event families; \emph{rate-limit} exploitation. \\
\addlinespace
\textbf{ALITA}~\citep{qiu2025alita} & Minimal core exploits cached MCP tools; upon gaps it \emph{explores} by brainstorming–codegen–run to create new tools, then reuses (exploit) via MCP registry and health checks. & Formalize \emph{epistemic justification} as a \emph{promotion} gate; place MCP registry in a \emph{two-band backbone}; add \emph{guard flips} to budget initiative (explore–exploit slider) and \emph{rate-limited} calls to validated tools; log cross-agent collaboration events. \\
\addlinespace
\textbf{TextGrad}~\citep{yuksekgonul2025optimizing} & Textual backprop propagates critiques (probes) and iteratively repairs modules (explore) until improved performance stabilizes (exploit via best-so-far selection). & Treat each repair as a \emph{promotion} candidate into the green store; maintain orange/green bands per module; meta-level \emph{guard flips} to enable/disable families of edits; \emph{rate-limit} green draws; couple with \emph{teach-back} checkpoints. \\
\bottomrule
\end{tabularx}
\end{adjustbox}
\caption{Mapping recent \enquote{quasi} stage-IV implementations to the \emph{explore–exploit} principle and targeted HE\textsuperscript{2}-Net boosts. \emph{Promotion} = \emph{Test + Peer Review} gate; \emph{two-band backbone} = \emph{orange} (provisional) / \emph{green} (validated); \emph{guard flips} = meta-level reconfiguration over families of transitions; \emph{rate-limited} green flows stabilize exploitation (cf. Secs.~\ref{sec:three-layer-haacs}, \ref{sec:he2-intro}).}
\label{tab:impl-explore-exploit}
\end{table}

Together, these mappings clarify how current systems already embody parts of stage-IV (archives, refinement loops, event/workflow control, self-evolution, textual repair) while HE\textsuperscript{2}-Net supplies the missing \emph{meta-governance}, \emph{promotion}, and \emph{common-ground backbone} needed to robustly coordinate humans and AI agents in open, nonstationary environments.

\subsection{Case Study}
\subsubsection{Intrinsic challenges in healthcare and a stage-IV conceptual design for the orthopaedic inpatients case}
Healthcare problems are textbook \enquote{open complex} in our sense: multi-faceted goals (safety, efficiency, experience)~\citep{Memmert_2022} must be traded off under high stakes and tight temporal pressure; contexts drift (time variance), observations are partial/opaque, and effects are nonlinear (small perturbations in workflow or physiology can cascade) (See Sec.~\ref{sec:sys-ontology-haacs}). This raises intrinsic challenges on two fronts. For \emph{problem solving}, teams must coordinate across discrete–continuous variables (vital signs, bed states, OR timings), cope with partial observability, and decide under bounded rationality \citep{simon1996sciences}: satisficing policies, fallbacks, and rapid re-planning dominate over global optimization. For \emph{knowledge management}, useful knowledge is distributed and partly tacit (ward conventions, clinical heuristics); signals are noisy; and model drift and context shift demand continuous calibration, abstention, and teach-back. \par

\begin{figure}[htbp]
  \centering
  \includegraphics[width=0.82\linewidth, trim=
  60mm 
  12mm 
  76mm 
  25mm, clip]{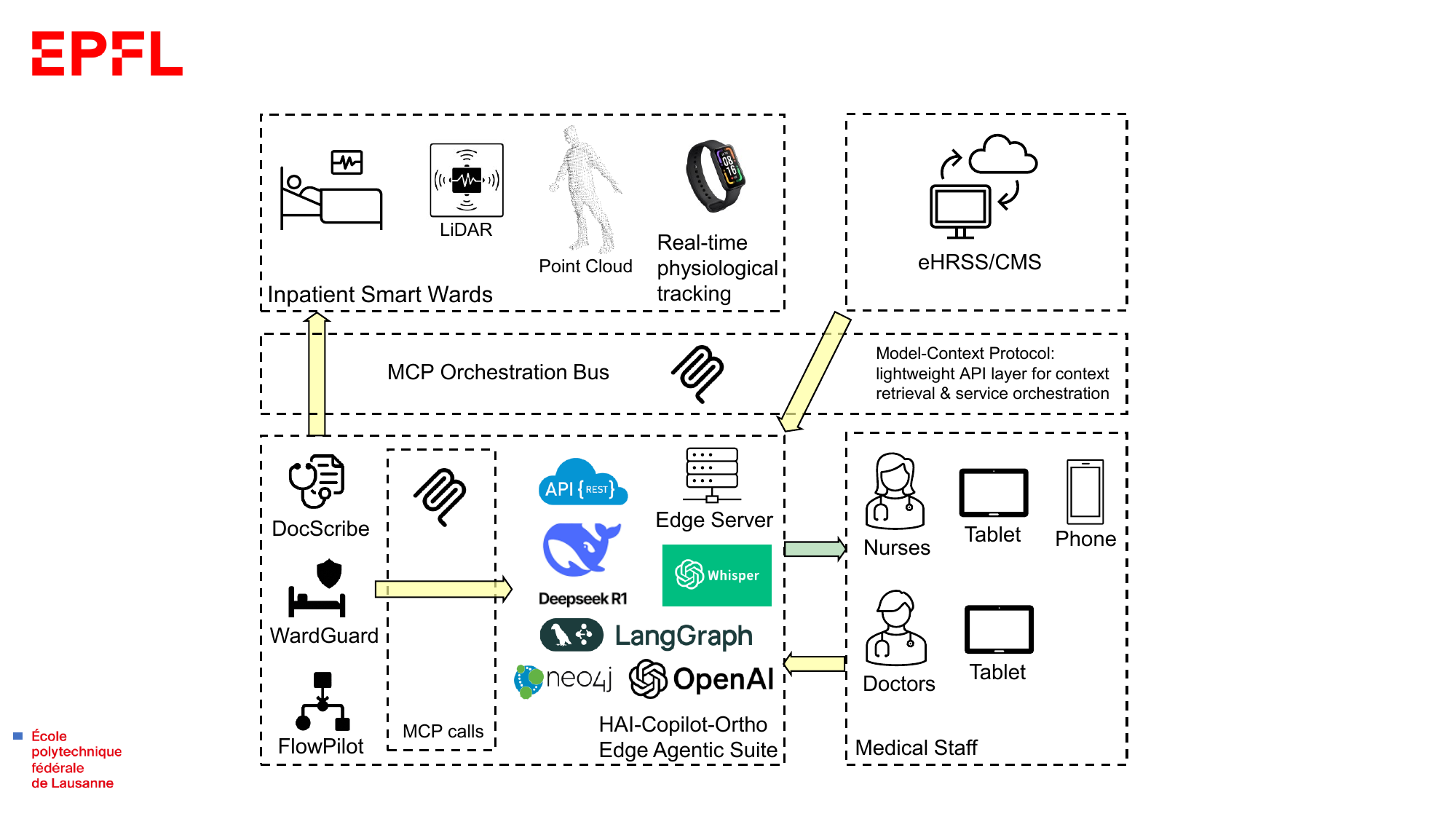}
  \caption{Architecture of the proposed \emph{HAI-Copilot-Ortho} edge agentic suite. A local MCP registry/bus exposes ward resources and tools, LiDAR/point-cloud and physiological streams, the FHIR client to CMS III/eHRSS, Whisper ASR, LLM endpoints, and the Neo4j graph store, as typed tools. The three edge agents, DocScribe (documentation), WardGuard (safety/falls), and FlowPilot (patient-flow), run on the on-ward server and invoke all sensors / external systems (CMS) / services via MCP. Nurses and doctors access the suite from ward tablets/phones. Green arrows indicate validated, routine flows; yellow connectors denote cross-boundary orchestration and policy-gated calls (e.g., LLM use).}
  \label{fig:hai-copilot-ortho-arch}
\end{figure}

\textbf{Orthopaedic inpatients context.} The target setting for the case is high-occupancy public orthopaedic wards in Hong Kong, with bilingual documentation burdens, constrained integration to CMS III/eHRSS (FHIR polling only), and a deliberate \enquote{edge-only} deployment of three agents (DocScribe, FlowPilot, WardGuard), each interacting with bedside LiDAR depth sensors and execute on an on-ward edge server, as schematized in Fig.~\ref{fig:hai-copilot-ortho-arch} (Architecture of the proposed \emph{HAI-Copilot-Ortho} edge agentic suite). Outcomes expected to improve with this architecture include documentation minutes, LOS, falls, alarm latency, NASA-TLX, and patient experience; the pilot is a cluster-RCT across four 30-bed wards, with explicit reliability, privacy and drift-audit requirements. These constraints instantiate the general challenges above as: (i) \emph{partial observability} (depth silhouettes; limited EHR fields); (ii) \emph{tight coupling} among bed flow, documentation, and fall-risk responses; (iii) \emph{nonstationarity} (seasonal load, staff rotations, evolving policies); (iv) \emph{safety + provenance} (notes must be clinician-approved; fall prompts must minimize alarm fatigue); and (v) \emph{edge-resident governance} (no central cloud orchestrator) with uptime $\geq$ 95\% and auditable changes across wards. \par

\textbf{Stage-IV collaboration requirements in this case.} A Stage-IV HAACS here must coordinate \emph{multiple human roles} (nurses, doctors), \emph{multiple edge agents} (DocScribe/FlowPilot/WardGuard), and \emph{multiple data planes} (sensors, local stores, FHIR polls) as an open complex system. Concretely, the collaboration needs: (1) a \emph{meta-level} that budgets initiative (explore–exploit) and can flip \emph{guards} over whole families of transitions (e.g., freeze a WardGuard threshold family hospital-wide after a drift alert); (2) a \emph{two-band knowledge backbone} separating provisional probes (orange) from validated assets (green) with an explicit \emph{Test + Peer Review} promotion gate; (3) \emph{boundary/collaboration transitions} that let agents escalate anomalies (HypothesisReport/DriftAlert) and route human teach-back; (4) \emph{rate-limited exploitation} with safe fallbacks (e.g., revert to standard HA bed-exit alarms and manual discharge checklists) to avoid cascading harm; and (5) a \emph{monotone event ledger} for auditable reconfiguration, aligning with implementation-science endpoints (RE-AIM, Proctor outcomes). These requirements arise directly from the edge-only, RCT-in-the-wild constraints and the need to preserve medico-legal provenance while adapting in real time. \par

Using the HE\textsuperscript{2}-Net, we specify a light-weight but complete governance shell that can be instantiated on the ward edge server shown in the study architecture. \par

On-ward we instantiate a minimal HE\textsuperscript{2}-Net control plane: a meta-orchestrator budgets initiative per agent, flips guards over capability families, and every reconfiguration is signed and logged. A two-band edge backbone provides shared knowledge: the provisional probes (e.g., draft prompts, LiDAR thresholds, discharge heuristics) and validated SOPs/prompts/thresholds with applicability, all with provenance and accessible to DocScribe/FlowPilot/WardGuard and staff GUIs. A single epistemic gate, i.e., \enquote{Test + Peer Review + teach-back}, governs knowledge promotion: DocScribe (documentation completeness + clinician teach-back), FlowPilot (advice-only runs + MDT review), WardGuard (silent calibration to acknowledged events + specificity/latency targets + nurse teach-back). Gate outcomes emit promotion/refute events that refresh the backbone and reconfigure guards, closing the explore–exploit loop. \par

\textbf{Concept-level verification.} The design is \enquote{correct by construction} with respect to our stage-IV requirements: (i) \emph{governability}: any family of actions can be paused or reconfigured via guard flips; (ii) \emph{traceable knowledge growth}: every provisional$\to$validated promotion is justified (tests + peer checks + teach-back) and reversible; (iii) \emph{safe adaptation}: drift triggers exploration without disrupting care (rate-limited exploitation flows, fallbacks); (iv) \emph{coherence}: a shared backbone plus teach-back maintains common ground across roles; and (v) \emph{experimental clarity}—policy states are first-class variables, enabling principled, auditable learning during the pilot. \par

In short, the HE\textsuperscript{2}-Net supplies the missing meta-governance, promotion gate, and common-ground backbone to coordinate DocScribe/FlowPilot/WardGuard with nurses and doctors under open, nonstationary ward conditions, satisfying the stage-IV brief (cf. Sec.~\ref{sec:pre-stages-shorts}) while remaining compatible with the edge-only, RCT design.
\section{Conclusion}
\label{sec:conclusion}

This position paper argues that moving from agentic demos to dependable human–AI collaboration requires three unifications: (i) a boundary-centric ontology that makes \emph{agenthood} precise under fluid composition (Secs.~\ref{sec:agenthood-ontology}–\ref{sec:cyber-agenthood}); (ii) a Petri net formalism that renders ownership, cross-boundary effects, concurrency, guards, and rates explicit (Secs.~\ref{sec:sys-formal-pn}–\ref{sec:three-layer-haacs}); and (iii) a collaboration pattern that governs \emph{initiative} and \emph{knowledge} together. The proposed HE\textsuperscript{2}-Net satisfies the Stage-IV brief distilled from the seven-dimension spine (defined in Sec.~\ref{sec:pre-stages-shorts}) and from the human–agent contrasts (Table~\ref{tab:ai-vs-human}): a meta-level policy budgets exploration vs reuse (exploitation), flips guards across families of transitions with auditable reconfiguration, and couples an epistemic \emph{promotion gate} to a system-wide backbone that stabilizes practice (Sec.~\ref{sec:he2-design-pn}, Sec.~\ref{sec:he2-intro}). On the knowledge side, CT+SECI with teach-back provides the governance missing in ad-hoc memory stacks, while on the problem-solving side, MEA for routine tasks and practice-guided discovery are coordinated rather than conflated (Sec.~\ref{sec:col-hai-learn}, Sec.~\ref{sec:practice-oe-ps}).

\bibliography{references} 
\appendix
\section{Dimension schema and reference tables}
\subsection{Human--AI agents collaboration evolution}

\textbf{Stage-I Man--Machine Automation} \\
\textbf{Human role:} Supervisor/goal-setter \;|\; Operator \\
\textbf{Interaction pattern (directionality $\times$ feedback):} One-way/sparse \;|\; Two-way/moderate--rich \\
\textbf{System observability/transparency:} Low \;|\; Medium \;|\; High \\
\textbf{Ethical oversight locus:} Explicit-human \;|\; Implicit \;|\; Absent \\
\textbf{Task-allocation locus (Decision, Execution):} (H,H) \;|\; (H,A) \;|\; (S,S) \;|\; (A,H) \;|\; (A,A). H = human, A = AI/automation, S = shared (joint). \\
\textbf{Collaboration cardinality:} 1H--1M \;|\; 1H--nM \;|\; nH--nM \\
\begin{table}[htbp]
\centering
\small
\setlength{\tabcolsep}{2.8pt}
\begin{adjustbox}{scale=0.88,center}
\begin{tabularx}{1.4\textwidth}{
@{}
    p{2.0cm}  
    p{1.8cm}  
    p{1.8cm}  
    p{2.8cm}  
    p{2.2cm}  
    p{1.9cm}  
    p{2.4cm}  
    p{1.6cm}  
    @{}
}
\toprule
\textbf{Work} & \textbf{Domain} & \textbf{Human role} & \textbf{\makecell[l]{Interaction\\(dir $\times$ feedback)}} & \textbf{Observability} & \textbf{\makecell[l]{Ethical\\oversight}} & \textbf{\makecell[l]{Task-allocation\\(D,E)}} & \textbf{Cardinality} \\
\toprule
\citep{wiener1988human, wiener2019cybernetics, wiener1959man} & Cybernetics \& ethics & Supervisor & Two-way / mod-rich & Medium     & Explicit-human & (H,A) & 1H--1M \\
\citep{simon1996sciences}     & Cognitive science / design theory & Supervisor & Two-way / mod-rich & Med--High  & Explicit-human & (H,A) & 1H--1M \\
\citep{xuesen1993new}     & Human factors / function allocation & Supervisor & Two-way / mod-rich & Medium     & Explicit-human & (H,A) & nH--nM \\
\citep{miller1991applications}     & Aerospace ops / process-tracking & Supervisor & Two-way / mod-rich & \textbf{High} & Explicit-human & (H,A) & nH--nM \\
\bottomrule
\end{tabularx}
\end{adjustbox}
\caption{Stage-I Man--Machine Automation: Dimension schema and representative references.}
\label{tab:app-stage-1}
\end{table}
\\
\textbf{Stage-II Flexible Autonomy \& Co-active Design} \\
\textbf{Role handoffs:} Static \;|\; Flexible \\
\textbf{Governance layer (norms \& policies):} None \;|\; Static \;|\; Dynamic \\
\textbf{Progress appraisal \& common ground:} None \;|\; Intermittent \;|\; Ongoing \\
\textbf{Initiative vs directability:} Human-led \;|\; AI-led \;|\; Balanced \\
\textbf{Primary role pattern:} Supervisory \;|\; Mediating \;|\; Cooperative \;|\; Mixed \\
\textbf{Mutual-learning (directionality):} None \;|\; H$\to$A \;|\; A$\to$H \;|\; H$\leftrightarrow$A \\

\begin{table}[htbp]
\centering
\small
\setlength{\tabcolsep}{3pt}
\renewcommand{\arraystretch}{1.05}
\setlength{\tabcolsep}{2.8pt}
\begin{adjustbox}{scale=0.88,center}
\begin{tabularx}{1.4\textwidth}{
@{}
    p{0.20\textwidth}   
    p{0.2\textwidth}  
    Y  
    p{0.16\textwidth}   
    p{0.16\textwidth}   
    p{0.16\textwidth}   
    p{0.16\textwidth}   
    Y  
}
\toprule
\textbf{Work} & \textbf{Domain} & \textbf{Role handoffs} & \textbf{Governance layer} & \textbf{Progress \& Common ground} & \textbf{Initiative vs directability} & \textbf{Primary role pattern} & \textbf{Mutual-learning} \\
\toprule
\citep{Ca_as_2022}  & AI ethics / governance                         & Flexible     & \textbf{Dynamic}  & \textbf{Ongoing}        & Balanced                  & Mixed                   & \textbf{$H\leftrightarrow A$} \\
\citep{Kolbj_rnsrud_2023}   & Organizational design / mgmt             & Flexible     & \textbf{Static}   & Intermittent            & Balanced                  & Mixed                   & $H\leftrightarrow A$ \\
\citep{Davies_2021}   & Science discovery / mathematics                & Static       & None              & None                    & \textbf{AI-led}           & Cooperative             & \textbf{$A\to H$} \\
\citep{wang2024towards}   & HCI theory / mutual learning                  & Flexible     & None              & Intermittent            & Balanced                  & Cooperative             & \textbf{$H\leftrightarrow A$} \\
\citep{fragiadakis2024evaluating}   & HCI Evaluation / metrics framework                 & \textbf{Static} & None            & Intermittent            & Mixed                     & Mixed                   & None \\
\citep{bradshaw2017human}  & human--agent interaction co-active design                         & \textbf{Flexible} & \textbf{Dynamic} & \textbf{Ongoing}     & \textbf{Balanced}         & \textbf{Cooperative /}\textbf{Mixed} & \textbf{$H\leftrightarrow A$} \\
\citep{Lai_2021}  & Healthcare HCI                  & Flexible     & Static            & Intermittent            & Human-led                 & Supervisory /Mediating   & None \\
\citep{Zhang_2024}  & Healthcare / critical care (sepsis)            & \textbf{Flexible} & Static         & \textbf{Ongoing}        & \textbf{Balanced}         & \textbf{Mediating}      & \textbf{$A\to H$} \\
\citep{Senoner_2024}  & Healthcare / XAI diagnosis                                & Static       & None              & Intermittent            & \textbf{Human-led}        & Supervisory             & \textbf{$A\to H$} \\
\citep{stammer2023learning}  & ML explainability (AI$\to$AI)                  & Static       & None              & None                    & AI-led                    & \textemdash             & \textbf{None} \\
\citep{hemmer2022factors}  & Healthcare / clinical decision-making & \textbf{Flexible} & \textbf{Dynamic} & \textbf{Ongoing}     & \textbf{Balanced}         & \textbf{Cooperative}    & \textbf{$H\leftrightarrow A$} \\
\citep{boy2024human}  & Human--AI teaming / safety-critical ops     & \textbf{Flexible} & \textbf{Dynamic} & \textbf{Ongoing}     & \textbf{Balanced}         & \textbf{Mixed}          & \textbf{$H\leftrightarrow A$} \\
\citep{boy2023epistemological}  & Systems engineering / HMS design               & Flexible     & \textbf{Dynamic}  & \textbf{Ongoing}        & Balanced                  & Mixed                   & $H\leftrightarrow A$ \\
\bottomrule
\end{tabularx}
\end{adjustbox}
\caption{Stage-II Flexible Autonomy \& Co-active Design: Dimension schema and representative systems.}
\label{tab:app-stage-2}
\end{table}

\noindent \textbf{Stage-III Agentic-AI \& Multi-Agent Collaboration} \\
\textbf{Team topology:} 1H--1A \;|\; 1H--nA \;|\; nA-only \\
\textbf{Communication:} Implicit \;|\; Explicit \;|\; Both \\
\textbf{Observability:} Full \;|\; Partial \;|\; Hybrid \\
\textbf{Coordination architecture:} Centralized-moderator \;|\; Distributed-consensus \;|\; Hybrid \\
\textbf{Common-ground / policy orchestration:} None \;|\; Static \;|\; Dynamic \\
\textbf{Human integration locus:} In-loop \;|\; On-loop \;|\; Out-of-loop \\

\begin{table}[htbp]
\centering
\small
\setlength{\tabcolsep}{2.8pt}
\begin{adjustbox}{scale=0.88,center}
\begin{tabularx}{1.4\textwidth}{
@{}
p{2.0cm} 
p{2.8cm} 
p{1.8cm} 
p{1.5cm} 
p{2.2cm} 
p{2.4cm} 
p{2.4cm} 
p{1.8cm} 
@{}
}
\toprule
\textbf{Work} & \textbf{Domain} & \textbf{Team topology} & \textbf{Comm.} & \textbf{Observability} & \textbf{Coord. arch.} & \textbf{Orchestration} & \textbf{Human integration} \\
\toprule
\citep{wu2024autogen} & Multi-agent framework (conversation programming)   & nA-only (config.)     & \textbf{Explicit}                 & Full (dialog buffer)        & \textbf{Centralized-moderator} (GroupChatManager) & Static (prompts/buffer)                         & In-loop (config.) \\
\citep{chen2024agentverse} & Multi-agent framework (sequential problem-solving) & nA-only               & \textbf{Explicit}                 & Partial                     & \textbf{Distributed-consensus} (no moderator)     & Static (round transcripts)                      & Out-of-loop \\
\citep{hong2023metagpt} & Software eng. pipeline                             & nA-only               & \textbf{Explicit}                 & Full (shared pool)          & Centralized (SOP/pipeline)                        & \textbf{Dynamic} (iterative reviews)            & On-loop \\
\citep{qian2024chatdev} & Software eng. pipeline                             & nA-only               & \textbf{Explicit}                 & Full (phase memory)         & \textbf{Centralized} (hierarchy)                  & Static (phase docs)                             & On-loop \\
\citep{zhang2024building} & Embodied multi-agent simulation (DEC-POMDP)                    & nA-only (config.)     & \textbf{Both} (msgs + actions)    & \textbf{Partial}            & \textbf{Distributed-consensus}                    & \textbf{Dynamic} (episodic/\allowbreak semantic/\allowbreak procedural mem.) & Out-of-loop (config.) \\
\citep{guo2024embodied} & Org. design / coordination optimization            & 1H--nA (pilot)        & \textbf{Explicit}                 & Partial                     & \textbf{Centralized-moderator} (leader)           & \textbf{Dynamic} (prompted re-org)              & \textbf{In-loop} (human trials) \\
\citep{tang2024medagents} & Medical QA (multi-expert)                          & nA-only               & \textbf{Explicit}                 & Full                        & \textbf{Distributed-consensus} (voting)           & \textbf{Dynamic} (iterative synthesis)          & Out-of-loop \\
\citep{kim2024mdagents} & Medical QA (moderated)                             & nA-only               & \textbf{Explicit}                 & Full                        & \textbf{Centralized-moderator}                    & \textbf{Dynamic} (dialog buffer)                & Out-of-loop \\
\citep{li2024agent} & Simulated hospital (end-to-end)                    & nA-only               & \textbf{Explicit}                 & Full & \textbf{Centralized} (pipeline)                   & \textbf{Dynamic} (experience base)              & Out-of-loop \\
\citep{sumers2024cognitive} & Cognitive architecture (single-agent)              & 1H--1A (general)      & Explicit                          & \textbf{Hybrid} (inner vs outer) & Centralized (single agent)                   & \textbf{Dynamic} (memories)                     & On-loop \\
\citep{spivack2024cognition} & Cognitive-AI (dual-layer blueprint)                & 1H--1A                & Explicit                          & \textbf{Hybrid}             & \textbf{Hybrid} (neuro-symbolic + LLM)            & \textbf{Dynamic}                                 & \textbf{In-loop} \\
\citep{strouse2021collaborating} & RL teaming (Overcooked, zero-shot)                 & \textbf{1H--1A} (test-time) & \textbf{Implicit}          & \textbf{Partial}            & \textbf{Distributed}                              & \textbf{None}                                    & \textbf{In-loop} \\
\citep{yan2023efficient} & RL coordination (single-step)                      & \textbf{1H--1A} (test-time) & \textbf{Implicit}          & \textbf{Partial}            & \textbf{Distributed}                              & \textbf{None}                                    & \textbf{In-loop} \\
\citep{feng-etal-2024-large} & Offline RL task allocation (ReAct)                 & \textbf{1H--1A}       & Explicit                          & Partial                     & \textbf{Centralized-moderator} (learned switcher) & \textbf{Dynamic} (traj buffer)                   & \textbf{In-loop} \\
\citep{puig2021watchandhelp} & Embodied two-player (Watch-\&-Help)                & \textbf{1H--1A} (or nA) & \textbf{Implicit}              & \textbf{Partial}            & \textbf{Distributed}                              & Dynamic (per-agent beliefs)                       & \textbf{In-loop} \\
\citep{liang2019implicit} & Two-player game (Hanabi)                           & \textbf{1H--1A}       & \textbf{Implicit} (minimal hints) & \textbf{Partial}            & \textbf{Distributed}                              & \textbf{Static} (conventions)                    & \textbf{In-loop} \\
\citep{zhang2024large} & GUI automation (LAM/agents)                        & \textbf{1H--1A}       & Explicit (commands)               & \textbf{Hybrid} (GUI state)  & Centralized (single agent)                        & \textbf{Dynamic} (task memory)                   & \textbf{In-loop} \\
\bottomrule
\end{tabularx}
\end{adjustbox}
\caption{Stage-III Agentic-AI \& Multi-Agent Collaboration: Dimension schema and representative systems.}
\label{tab:app-stage-3}
\end{table}
\subsection{Evolution of Conversation Theory}
\begin{table}[htbp]
\centering
\small
\setlength{\tabcolsep}{2.8pt}
\begin{adjustbox}{scale=0.88,center}
\begin{tabularx}{1.4\textwidth}{
@{}
p{0.16\textwidth} 
p{0.24\textwidth} 
p{0.2\textwidth} 
p{0.28\textwidth} 
p{0.22\textwidth} 
p{0.22\textwidth} 
}
\toprule
\textbf{Dimension} &
\textbf{CT foundations (Pask)} &
\textbf{Classroom \& tutoring operationalization} &
\textbf{THOUGHTSTICKER (mesh, non-directed)} &
\textbf{Directed entailment nets (+ CONCEPTORGANIZER)} &
\textbf{LLM-mediated multi-agent \enquote{languaging} \& consensual domains} \\
\toprule
\textbf{Key constructs} &
\enquote{\textbf{having knowledge}} as \enquote{\textbf{knowing}} and \enquote{\textbf{coming to know}}; \textbf{P-individuals \& M-individuals}. \textbf{Conversation} drives construction and coherence. \textbf{Teach-back} anchors verification.~\citep{pask1976conversation} &
\textbf{Establishing a tangible shared environment}; \textbf{multi-level exchanges} anchored to problems, examples, and scenarios; comparisons of \textbf{procedures + outcomes} guide refinement.~\citep{boyd2001reflections} &
Assisting to transform users' \textbf{implicit knowledge} into \textbf{stable and explicit knowledge} with a computer \textbf{conversational partner}; \textbf{Paskian} interaction scaffolds elicitation.~\citep{pask1976conversation} &
\textbf{Nodes are distinct precisely when their respective input/output sets differ}; \textbf{bootstrapping axiom} governs identity; \textbf{minimal ontology} organizes concepts.~\citep{heylighen2001bootstrapping} &
Within a group of intelligent agents (\textbf{human or AI}) \textbf{they learn from and teach each other}; \textbf{structural coupling} yields \textbf{emergent collective agents} in \textbf{consensual domains}.~\citep{scott2001gordon,maturana1990biological} \\
\textbf{Dialogue levels} &
\textbf{task-focused}, \textbf{explanatory}, and \textbf{meta-cognitive} levels coordinate actions and justifications.~\citep{pask1976conversation} &
\textbf{Task $\leftrightarrow$ explanatory $\leftrightarrow$ meta} with explicit comparison of \textbf{procedures + outcomes}; level shifts triggered by discrepancies.~\citep{boyd2001reflections} &
\textbf{Explanatory level} prompts (\enquote{\textbf{Why?}} and \enquote{\textbf{How do you know?}}) usable in a \textbf{Paskian environment} to surface commitments.~\citep{pask1976conversation} &
Works across \textbf{task, explanatory, meta} via explicit links and \textbf{teach-back} triggers for coherence checks.~\citep{heylighen2001bootstrapping} &
\textbf{task, explanatory, meta} with turn-taking and coordinator roles; \textbf{teach-back} checkpoints maintain alignment.~\citep{scott2001gordon,boyd2001reflections} \\
\textbf{Knowledge representation} &
\textbf{Entailment meshes} as \textbf{dynamic webs of concepts} connected by \textbf{shared procedures and outcomes}; coherence arises from interrelations.~\citep{pask1976conversation} & 
\textbf{Evolving knowledge mesh} in which \textbf{concepts that share outcomes or have overlapping procedures become linked or \enquote{entail} each other}; graph views support comparison.~\citep{boyd2001reflections} & 
\textbf{Mesh (non-directed)} of overlapping clusters; topics \textbf{defined by their network of associations}; similarity drives grouping.~\citep{pask1976conversation} &
\textbf{Directed links} with \textbf{transitive or causal chains}; \textbf{minimal ontology} (Class, Object, Property); \textbf{inheritance} clarifies hierarchy and defaults.~\citep{heylighen2001bootstrapping} &
\textbf{Languaging} as \textbf{recursive, multi-layered, interactive meaning-making} with shared artifacts in a \textbf{micro-world}; representations co-evolve through interaction.~\citep{maturana1990biological,boyd2001reflections} \\
\bottomrule
\end{tabularx}
\end{adjustbox}
\caption{Learning-as-Conversation from Entailment Meshes $\to$ Directed Entailment Nets $\to$ Consensual Domains - Part I}
\label{tab:ct-evolution-p1}
\end{table}
\begin{table}[htbp]
\setlength{\tabcolsep}{2.8pt}
\begin{adjustbox}{scale=0.88,center}
\begin{tabularx}{1.4\textwidth}{
@{}
p{0.16\textwidth} 
p{0.24\textwidth} 
p{0.2\textwidth} 
p{0.28\textwidth} 
p{0.22\textwidth} 
p{0.22\textwidth} 
}
\toprule
\textbf{Dimension} &
\textbf{CT foundations (Pask)} &
\textbf{Classroom \& tutoring operationalization} &
\textbf{THOUGHTSTICKER (mesh, non-directed)} &
\textbf{Directed entailment nets (+ CONCEPTORGANIZER)} &
\textbf{LLM-mediated multi-agent \enquote{languaging} \& consensual domains} \\
\toprule
\textbf{Refinement} or \textbf{learning operators} &
\textbf{Saturation (gathering variants)}, \textbf{bifurcation (splitting on divergence)}, \textbf{pruning (eliminating/merging redundancies)}; parallel concept execution; recursive self-reference.~\citep{pask1976conversation} &
\textbf{Mismatched outcomes} prompt \textbf{re-examination} or \textbf{branching (bifurcation)}; \textbf{redundancies winnowed (prune)}; repeated \textbf{teach-back} consolidates.~\citep{boyd2001reflections} &
\textbf{Saturating}, \textbf{pruning}, \textbf{condensing}; \textbf{merge/split} when neighbor sets are identical; user steers restructuring.~\citep{pask1976conversation} &
\textbf{Associative learning via weighted links} (\textbf{direct}, \textbf{transitive}, \textbf{symmetric}); \textbf{ambiguity resolution} by \textbf{merging or differentiating concepts}; apply \textbf{saturate--bifurcate--prune} to the net.~\citep{heylighen2001bootstrapping,pask1976conversation} &
Conversational \textbf{selectionist trial-and-error}; \textbf{co-construction} and consensus formation; \textbf{saturate--bifurcate--prune} mediated by coordination tools.~\citep{scott2001gordon,pask1976conversation} \\
\textbf{Shared context} or \textbf{validation} &
\textbf{Teach-back} verifies \enquote{what/how/why}; \textbf{shared environment} emerges through dialogue; agreement stabilizes meshes.~\citep{pask1976conversation} &
\textbf{Shared environment (micro-world)} with the anchor \enquote{\textbf{Are we talking about the same phenomenon?}}; ongoing \textbf{teach-back} for alignment.~\citep{boyd2001reflections} &
\textbf{Dialogue-driven} elicitation; \textbf{user-guided} refinement; computer partner keeps procedural trace for validation.~\citep{pask1976conversation} &
\textbf{Self-organizing, adaptive system}; \textbf{coherence} via users' \textbf{collective agreement}; continuous \textbf{ambiguity checks}.~\citep{heylighen2001bootstrapping} &
\textbf{Consensual domains}; \textbf{shared \enquote{modeling facility} / micro-world}; explicit \textbf{teach-back} gates for verification in practice.~\citep{scott2001gordon,boyd2001reflections} \\
\textbf{Known limits} or \textbf{notes} &
\textbf{Complex notations}; \textbf{insufficient treatment} of emotional, motivational, and embodied dimensions.~\citep{pask1976conversation} &
Protocols for \textbf{overlaps/interruptions} and \textbf{affective/multimodal} cues are still developing; scalability under study.~\citep{boyd2001reflections} &
\textbf{No directional or hierarchical encoding}; \textbf{heavy reliance} on user guidance for restructuring.~\citep{pask1976conversation} &
Needs \textbf{high-quality elicitation and curation}; \textbf{ontology growth} must be stewarded to avoid drift.~\citep{heylighen2001bootstrapping} &
\textbf{Interaction fidelity}, emotions, motivation, and \textbf{multi-modal cues} remain open problems; dependence on prompting and stopping criteria.~\citep{scott2001gordon,boyd2001reflections} \\
\bottomrule
\end{tabularx}
\end{adjustbox}
\caption{Learning-as-Conversation from Entailment Meshes $\to$ Directed Entailment Nets $\to$ Consensual Domains - Part II}
\label{tab:ct-evolution-p2}
\end{table}

\end{document}